\documentclass{article}

\usepackage{arxiv} 
\renewcommand{\headeright}{}
\usepackage{graphicx}
\usepackage[utf8]{inputenc}
\usepackage[T1]{fontenc}
\usepackage{url}
\usepackage{booktabs}
\usepackage{amsfonts}
\usepackage{amsmath}
\usepackage{amssymb}
\usepackage[numbers,sort&compress]{natbib}
\usepackage{nicefrac}
\usepackage{graphicx}
\usepackage{multirow}
\usepackage{enumitem}
\usepackage{array}
\usepackage{caption}
\usepackage{tabularx}
\usepackage{makecell}
\usepackage{threeparttable}
\usepackage{placeins}
\usepackage{float} 
\usepackage{longtable} 
\usepackage{booktabs} 
\usepackage{rotating}
\usepackage{makecell}   
\usepackage{multirow}
\newcolumntype{Y}{>{\RaggedRight\arraybackslash}X}
\newcolumntype{C}[1]{>{\centering\arraybackslash}p{#1}}
\newcolumntype{L}[1]{>{\raggedright\arraybackslash}p{#1}}

\newcolumntype{M}[1]{>{\RaggedRight\arraybackslash}m{#1}}
\newcolumntype{Z}[1]{>{\centering\arraybackslash}m{#1}}
\providecommand{\editormark}{\textdagger}
\newcommand{\tabremark}[1]{
\vspace{0.35em}
\begin{minipage}{0.96\linewidth}
\raggedright
\scriptsize
#1
\end{minipage}
}
\newcommand{\compactappendixtable}{%
  \scriptsize
  \setlength{\tabcolsep}{3.2pt}
  \renewcommand{\arraystretch}{0.92}
}

\makeatletter
\let\old@maketitle\@maketitle

\def\@maketitle{
  \begin{center}
    \let\footnote\thanks
    \@author
  \end{center}
}
\makeatother

\title{
PhysEditBench: A Protocol-Conditioned Benchmark for Dense Physical-Map Prediction with Image Editors
}

\author{
\textbf{Jiaxin Yang}$^{1,*}$ \quad
\textbf{Yu Hou}$^{1,*}$ \quad
\textbf{Muxin Liu}$^{2}$ \quad
\textbf{Weixuan Liu}$^{3}$ \quad
\textbf{Ze Yuan}$^{2}$ \quad \\
\textbf{Zeming Chen}$^{1}$ \quad
\textbf{Zhongrui Wang}$^{1,\dagger}$ \quad
\textbf{Xiaojuan Qi}$^{2,\dagger}$
\\[0.5em]
\normalfont
$^{1}$Southern University of Science and Technology, Shenzhen, China \\
$^{2}$The University of Hong Kong, Hong Kong, China \quad
$^{3}$East China Normal University, Shanghai, China
\\[0.4em]
\texttt{\{12532758, houy2025\}@mail.sustech.edu.cn},\quad
\texttt{mxliu@connect.hku.hk} \\
\texttt{wxliu@stu.ecnu.edu.cn},\quad
\texttt{yuanze1024@gmail.com},\quad
\texttt{chenzm2025@mail.sustech.edu.cn} \\
\texttt{wangzr@sustech.edu.cn},\quad
\texttt{xjqi@eee.hku.hk} 
\\[0.4em]
$^{*}$Equal contribution \quad
$^{\dagger}$Corresponding author
}

\date{}

\date{}

\begin{document}

\begin{center}
\rule{\linewidth}{2.5pt} \\[1em] 
\LARGE\bfseries
PhysEditBench: A Protocol-Conditioned Benchmark \\
for Dense Physical-Map Prediction with Image Editors
\\[-0.1em]  
\rule{\linewidth}{1pt}  
\end{center}

\maketitle

\begin{figure}[!htbp]
\centering
\includegraphics[width=1.0\textwidth]{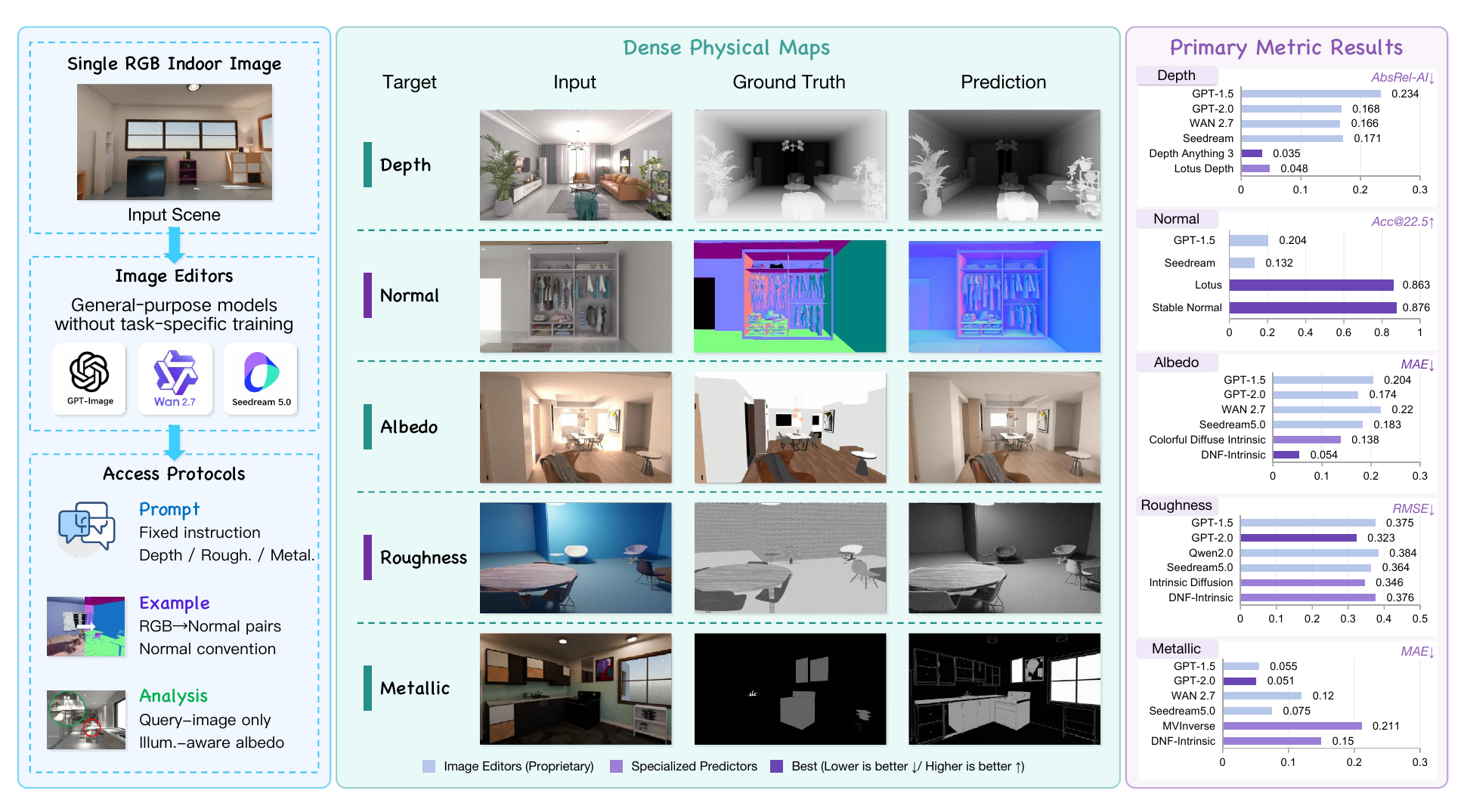}
\caption{
\textbf{The motivation and result snapshot of PhysEditBench.}
PhysEditBench evaluates whether general-purpose image editors can predict dense physical maps from a single RGB indoor image without task-specific training.
The benchmark covers depth, normal, albedo, roughness, and metallic maps under target-specific access protocols.
The figure shows representative predictions and primary-metric comparisons between proprietary image editors and specialized predictors; the full benchmark design is summarized in Figure~\ref{fig:overview}.
}
\label{fig:teaser_dense_maps}
\end{figure}

\begin{abstract}
Can general-purpose image editors predict physical maps from a single RGB image?
General-purpose image editors differ from standard task-specific dense-prediction models: they do not directly take an image and output a physical map.
Instead, they must be guided by prompts, examples, or image-based textual cues.
To this end, we introduce PhysEditBench, a novel protocol-conditioned benchmark to evaluate
and standardize image editors in dense physical-map prediction that covers five targets: depth, normal, albedo, roughness, and metallic maps.
For evaluation data, we build a target-dependent benchmark substrate.
We use OpenRooms-FF for depth, surface normal, albedo, and roughness, InteriorVerse as an additional source for depth, normal, albedo, and a new procedurally generated source for metallic maps.
We curate the data with quality checks, valid-region masks, scene-level sampling, and lighting-based stress subsets to ensure reliable and diverse evaluation.
For each target, PhysEditBench defines a fixed protocol that specifies the allowed input, expected output format, and scoring procedure.
Each score, therefore, reflects the performance of a model under a specified protocol, rather than its best possible performance under all prompts or interaction modes.
Experimental results show that specialized models remain much stronger on depth, normal, and albedo, and stronger image editors can produce more reasonable map-like outputs.
For roughness and metallic, image editors can match or outperform specialized baselines on some scalar metrics, but they still suffer from structural errors, sparsity effects, and sensitivity to lighting.
\end{abstract}

\section{Introduction}

In recent years, general-purpose image editors, such as Nano-Banana~\cite{google2025gemini_flash_image} and GPT-Image-2~\cite{openai2026gpt_image2}, have evolved into versatile image-conditioned transformation systems. Given an input image and a text instruction, they can preserve scene layout, modify appearance, and produce visually coherent edits~\cite{brooks2023instructpix2pix,zhang2023magicbrush,sheynin2024emu_edit}.
These abilities suggest that image editors may capture information about geometry, materials, lighting, object boundaries, and scene context~\cite{pbr_boost_3dgen,ir3d_bench,vision_banana}.
This raises a question: can off-the-shelf image editors predict dense physical maps, including depth, normal, albedo, roughness, and metallic maps, from a single RGB image without task-specific training?

This task differs from testing a task-trained dense predictor, which directly maps an RGB image to a predefined target map~\cite{iris_cvpr2025,dnf_intrinsic,Lotus,stablenormal,dsine,intrinsic_image_diffusion}.
For a general-purpose image editor, the requested map type, allowed evidence, output convention, and scoring rule must be specified for each target, since image editors are accessed through prompts, image-conditioned instructions, or exemplars~\cite{brooks2023instructpix2pix,zhang2023magicbrush,sheynin2024emu_edit}.
The evaluated physical maps also have different properties: depth is often ambiguous up to scale or ordering, normal maps depend on a color-coded direction convention, albedo separates intrinsic surface color from illumination, and metallic maps are sparse material-property fields~\cite{intrinsicanything,mvinverse}.
Thus, evaluating image editors for dense physical-map prediction requires target-specific experimental designs and metric choices, rather than a single prompting strategy or a single overall score.

We therefore introduce PhysEditBench, a benchmark for single-image dense physical-map prediction with image editors.
PhysEditBench covers five targets: depth, normal, albedo, roughness, and metallic maps. For each target, we define a disclosed protocol that specifies the allowed editor input and the expected output representation. This includes the task instruction, output format, target convention, allowed auxiliary information (see Appendix Table~\ref{tab:appendix_access_settings}), and any in-context examples needed to specify the target (see Appendix~\ref{app:normal_exemplar_selection}).
For evaluation, we fix the test split, scoring masks, metrics, aggregation rules, and stress-test definitions before scoring, using the same rules for all systems evaluated on the same target. Thus, each score measures performance under the specified target protocol, rather than performance under all possible prompts, exemplars, or interaction modes.

For evaluation, PhysEditBench uses target-dependent data sources because physical-map annotations differ in coverage and reliability.
OpenRooms-FF~\cite{openrooms,openrooms_ff} is used for depth, normal, albedo, and roughness, while InteriorVerse~\cite{interiorverse} complements depth, normal, and albedo.
InteriorVerse roughness and metallic are excluded from official scoring due to material-channel reliability concerns, and metallic is evaluated on a procedurally generated companion source because OpenRooms-FF does not provide metallic maps.
All retained splits, validity checks, stress subsets, and aggregation rules are fixed before scoring.
For depth, normal, and albedo, we report source-balanced macro-averages; roughness and metallic are reported on their retained single-source subsets.

This paper makes the following contributions:
\begin{itemize}[leftmargin=1.2em,itemsep=0.15em,topsep=0.25em]
    \item We introduce PhysEditBench, a benchmark for evaluating general-purpose image editors on single-image dense physical-map prediction across five targets: depth, normal, albedo, roughness, and metallic.

    \item We design target-specific access settings and diagnostic variants that separate fixed main comparisons from cue-sensitivity analyses, allowing us to examine how editor performance changes with different forms of task guidance.

    \item We provide target-wise results, uncertainty estimates, protocol diagnostics, photometric stress analyses, and qualitative examples showing that image-editor performance depends strongly on the target, metric, protocol, and image condition.
\end{itemize}

\section{Related Work}

\noindent\textbf{Image editors as structured visual transformation systems.}
Image-generation and image-editing systems were first developed for creative synthesis and instruction-guided editing, but recent systems increasingly support image-conditioned transformations, layout-preserving edits, and structured visual outputs~\cite{brooks2023instructpix2pix,zhang2023magicbrush,sheynin2024emu_edit}.
Related work has also explored PBR-aware generation, image-conditioned editing, and 3D-related scene understanding~\cite{pbr_boost_3dgen,epbr,ir3d_bench,vision_banana}.
In this paper, we use \textbf{image editors} to refer to image-generation or image-editing systems that can be queried through prompts, editing instructions, exemplars, or auxiliary textual analysis to produce structure-preserving transformations or dense-map-like outputs.
Existing evaluations mainly focus on instruction following, visual quality, editability, geometric plausibility, or downstream 3D utility~\cite{brooks2023instructpix2pix,zhang2023magicbrush,sheynin2024emu_edit,pbr_boost_3dgen,epbr,ir3d_bench,vision_banana}.
They do not measure whether off-the-shelf image editors can recover dense maps that match physical ground truth from a single RGB image.

\vspace{0.05in}\noindent\textbf{Task-trained indoor inverse rendering and dense prediction.}
Indoor inverse rendering and dense prediction have mature datasets, metrics, and specialized predictors for recovering physical scene properties from RGB images.
Datasets such as OpenRooms, OpenRooms-FF, InteriorVerse, and Hypersim provide supervision for geometry, material, and lighting-related quantities~\cite{openrooms,openrooms_ff,interiorverse,hypersim}.
Specialized methods, including IRIS, RGB$\leftrightarrow$X, Materialist, DNF-Intrinsic, Lotus, StableNormal, DSINE, and Intrinsic Image Diffusion, have advanced dense recovery of depth, normal, albedo, roughness, and related targets~\cite{iris_cvpr2025,rgbx_arxiv,materialist,dnf_intrinsic,Lotus,stablenormal,dsine,intrinsic_image_diffusion}.
These methods usually use target-specific supervision and provide direct RGB-to-map outputs.
They serve as strong reference points, but the gap between off-the-shelf image editors and dedicated dense-map predictors remains unclear under controlled, target-specific evaluation.

\vspace{0.05in}\noindent\textbf{Evaluation protocols and material reliability.}
Standard dense-prediction evaluations typically assume a direct RGB-to-map model, whereas image editors are accessed through prompts, editing instructions, exemplars, or image-derived textual cues~\cite{brooks2023instructpix2pix,zhang2023magicbrush,sheynin2024emu_edit,vision_banana}.
This interface difference makes the access setting and output convention part of the evaluation, rather than incidental implementation details.
Material targets add another complication: inverse rendering is ambiguous because geometry, materials, and illumination are coupled~\cite{intrinsicanything}, roughness and metallic supervision is less common than albedo supervision~\cite{mvinverse}, and prior studies report reliability concerns for InteriorVerse material-related channels~\cite{rgbx_arxiv,epbr}.
These factors motivate target-aware evaluation protocols and careful source selection.

\section{Benchmark Protocol}
\label{sec:benchmark_protocol}

\begin{figure}[!htbp]
\centering
\includegraphics[width=1.0\textwidth]{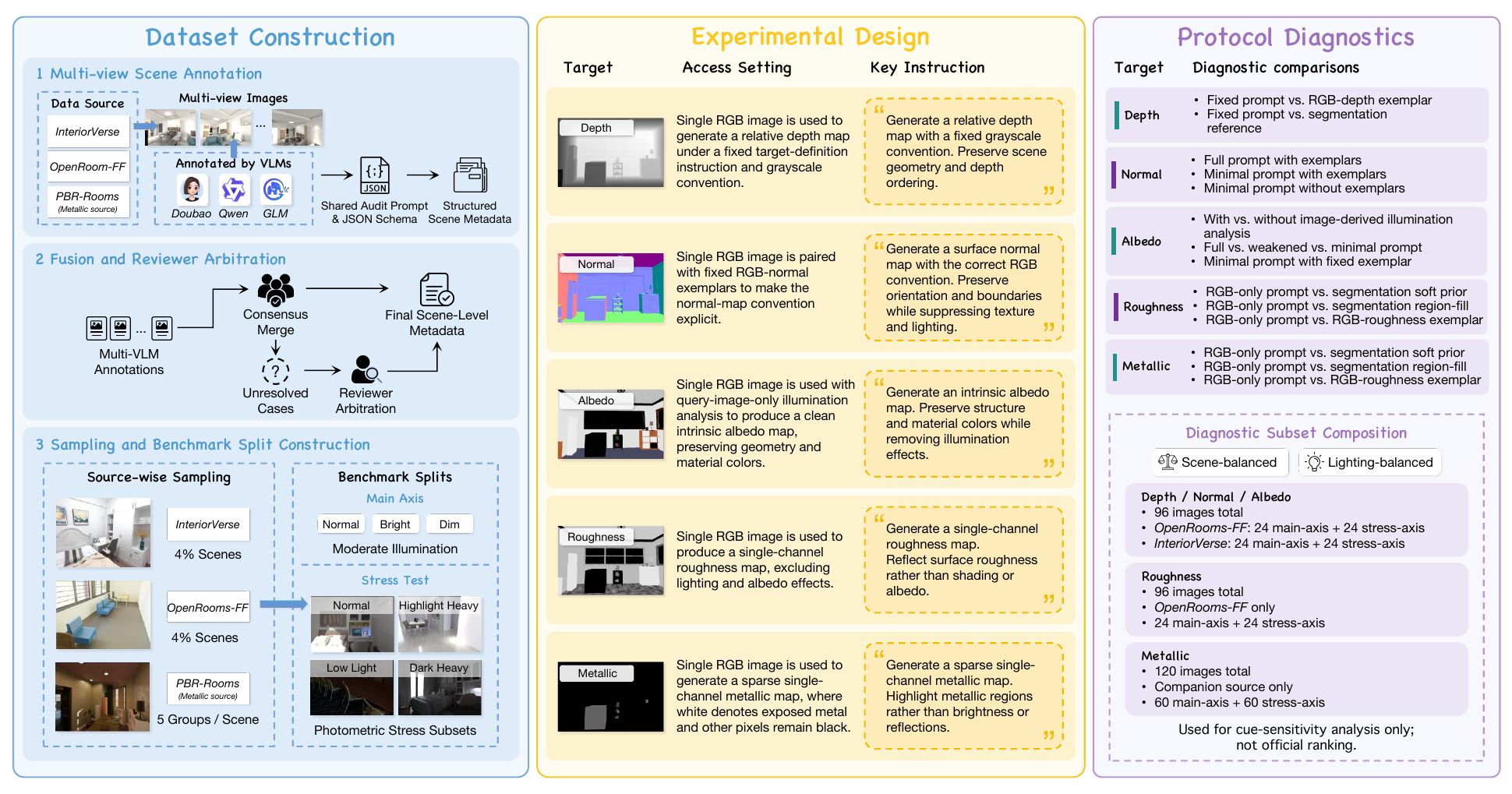}
\caption{
\textbf{Benchmark overview.}
The benchmark is organized around three components: dataset construction, target-specific experimental design, and protocol diagnostics.
Left: scene-level metadata are built from source-wise sampling, multi-view annotation, VLM-assisted auditing, and reviewer arbitration, followed by main-axis and photometric stress-axis split construction.
Middle: each target uses a fixed access setting and key instruction for eliciting depth, normal, albedo, roughness, or metallic maps from a single RGB image.
Right: diagnostic comparisons vary prompts, exemplars, illumination analysis, segmentation priors, or material exemplars on separate diagnostic subsets, and are used for cue-sensitivity analysis rather than official ranking.
}
\label{fig:overview}
\end{figure}

Figure~\ref{fig:overview} summarizes how PhysEditBench connects dataset construction, target-specific access settings, and diagnostic variants.

\noindent\textbf{Sources and target coverage.}
We use target-specific source selection rather than a single shared source pool. OpenRooms-FF is used for depth, normal, albedo, and roughness; InteriorVerse is used for depth, normal, and albedo.
InteriorVerse roughness and metallic are excluded from official scoring due to material-channel reliability concerns, consistent with prior reports~\cite{rgbx_arxiv,epbr}.
Because OpenRooms-FF does not provide metallic supervision, metallic is evaluated on a procedurally generated companion source.
Table~\ref{tab:dataset_summary} summarizes the retained sources, sample counts, primary metrics, and aggregation rules for the main-axis split. Stress-axis counts and detailed exclusion criteria are provided in Appendix~\ref{app:dataset_docs}.

\begin{table}[!htbp]
\centering
\footnotesize
\caption{Retained sources, counts, primary metrics, and aggregation rules for the main-axis split.}
\label{tab:dataset_summary}
\setlength{\tabcolsep}{5pt}
\renewcommand{\arraystretch}{1.04}
\begin{tabular}{llrrll}
\toprule
Target & Source & Scenes & Images & Primary metric & Aggregation \\
\midrule

\multirow{2}{*}{Depth}
& OpenRooms-FF & 48 & 1125 & \multirow{2}{*}{AbsRel-AI} & \multirow{2}{*}{Source-balanced} \\
& InteriorVerse & 167 & 2109 & & \\
\midrule
\addlinespace[0.15em]

\multirow{2}{*}{Normal}
& OpenRooms-FF & 48 & 1125 & \multirow{2}{*}{Acc@22.5} & \multirow{2}{*}{Source-balanced} \\
& InteriorVerse & 167 & 2109 & & \\
\midrule
\addlinespace[0.15em]

\multirow{2}{*}{Albedo}
& OpenRooms-FF & 48 & 1125 & \multirow{2}{*}{MAE} & \multirow{2}{*}{Source-balanced} \\
& InteriorVerse & 167 & 2109 & & \\
\midrule
\addlinespace[0.15em]

Roughness
& OpenRooms-FF & 48 & 1125 & RMSE & Source-specific \\
\midrule
Metallic
& Companion & 40 & 960 & MAE & Source-specific \\

\bottomrule
\end{tabular}

\tabremark{
Counts are for the official main-axis split only.
Stress-axis counts are reported in Appendix Table~\ref{tab:stress_axis_counts}.
Source-balanced scores are computed by source-specific scoring followed by macro-averaging.
InteriorVerse roughness and metallic are excluded from official scoring due to material-channel reliability concerns.
}
\end{table}

\vspace{0.1in}\noindent\textbf{Sampling design and evaluation axes.}
The source datasets are too large for exhaustive evaluation with proprietary image editors, so we construct retained evaluation subsets rather than scoring every available image. Sampling is performed at the scene level to preserve coverage across indoor scene categories and to avoid over-representing near-duplicate views from the same scene. The main-axis split emphasizes regular indoor views with relatively standard illumination and is used for the primary target-wise comparisons. The stress-axis split is constructed separately using fixed image-statistic rules to cover challenging photometric conditions such as low light, high dynamic range, highlight-heavy scenes, and dark-region-dominant scenes.
Scene metadata are used for constructing and auditing the splits, not as model inputs. All main-evaluation predictions are made from a single query RGB image. Detailed sampling rules and stress-subset definitions are provided in Appendix~\ref{app:split_construction}.

\vspace{0.05in}\noindent\textbf{Metrics and scoring.}
We use one primary metric per target, as summarized in Table~\ref{tab:dataset_summary}, and report additional metrics to expose boundary, structural, and perceptual failure modes. Depth uses affine-invariant AbsRel-AI because editor-produced depth maps may have arbitrary scale, shift, or polarity. These ambiguities are handled only at scoring time, so the depth comparison focuses on relative scene geometry rather than absolute metric depth. Normal uses Acc@22.5, albedo and metallic use MAE, and roughness uses RMSE, which penalizes larger material-parameter deviations more strongly.
Additional metrics are used to assess depth-boundary recovery, normal angular-error profiles, albedo structural/perceptual fidelity, and scalar or structural map consistency for roughness and metallic. Full evaluator definitions, scoring masks, normalization conventions, and failure-handling rules are provided in Appendix~\ref{app:extended_eval}.

\vspace{0.05in}\noindent\textbf{Image-editor access settings.} General-purpose image editors are not direct RGB-to-map predictors, so each target uses a fixed access setting before scoring. All settings include the query RGB image and a target-specific task definition, and no setting provides ground-truth maps, benchmark labels, test-set feedback, or multi-view scene evidence. Depth, roughness, and metallic use the query RGB image plus a fixed target-definition prompt. Albedo additionally uses a cached query-image-only illumination analysis, shared unchanged across proprietary albedo systems. Normal uses three fixed RGB--normal exemplar pairs to communicate the normal-map convention; the main normal score averages the three exemplar-conditioned runs, with per-exemplar results reported in Appendix~\ref{app:normal_exemplar_diagnostics}.
The full access-setting table and generation details are provided in Appendix~\ref{app:elicitation_protocols}.

\section{Main Evaluation Results}
\label{sec:main_comparison_results}
Evaluation metrics for all targets are summarized in Appendix Table~\ref{tab:metric_summary}, with full definitions, scoring masks, post-processing rules, and failure-handling details included in Appendix~\ref{app:extended_eval}.
In the main tables, we use one primary metric for each target: AbsRel-AI for depth, Acc@22.5 for normal, MAE for albedo and metallic, and RMSE for roughness.

\begin{figure}[!htbp]
\centering
\includegraphics[width=1.0\textwidth]{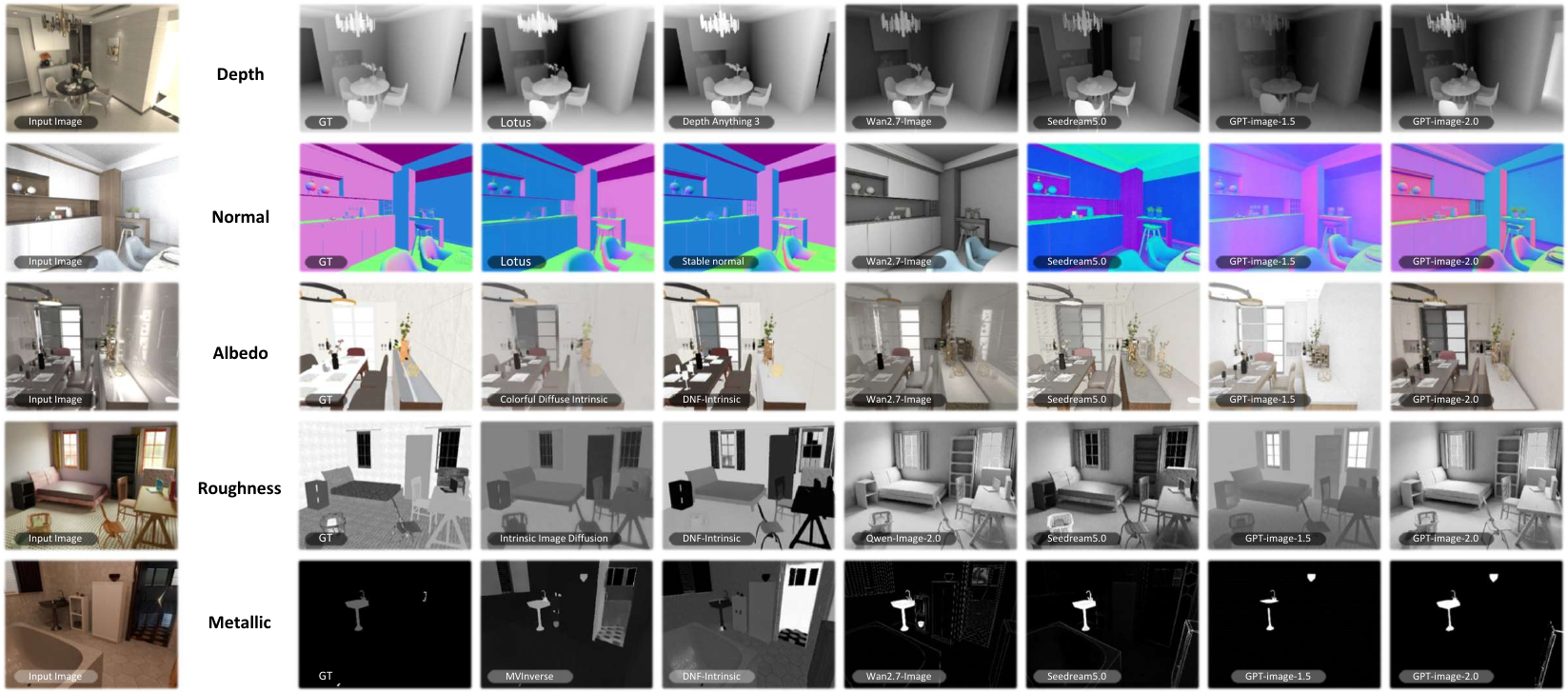}
\caption{
\textbf{Representative qualitative examples across the five evaluated targets.}
Each row shows the input RGB image, ground-truth map, and predictions from representative specialized predictors and proprietary image editors for one target.
The examples complement the quantitative results by illustrating differences in spatial structure, map fidelity, and target-specific failure patterns across depth, normal, albedo, roughness, and metallic prediction.
}
\label{fig:examples}
\end{figure}

\subsection{Target-wise Main-Axis Results}
\label{sec:main_axis_results}

We report main results separately for each target under the fixed access settings defined in Section~\ref{sec:benchmark_protocol}. Depth, normal, and albedo are source-balanced over OpenRooms-FF and InteriorVerse; roughness and metallic are evaluated on their retained single-source subsets. We compare proprietary image editors with specialized RGB-to-map predictors. In each table, the horizontal rule separates proprietary image editors from specialized predictors, and boldface marks the best rounded point estimate for each metric. Figure~\ref{fig:examples} provides representative qualitative examples to accompany the quantitative comparisons.

\begin{table}[!htbp]
\centering
\caption{Source-balanced relative-depth results under the affine-invariant single-image evaluation protocol.}
\label{tab:depth_mainaxis_combined_all}
\footnotesize
\setlength{\tabcolsep}{4.5pt}
\renewcommand{\arraystretch}{0.95}
\begin{tabular}{lcccccc}
\toprule
Method 
& AbsRel-AI$\downarrow$ 
& RMSE-AI$\downarrow$ 
& MAE-AI$\downarrow$ 
& $\delta_1$-AI$\uparrow$ 
& $\delta_2$-AI$\uparrow$ 
& Boundary F1$\uparrow$ \\
\midrule
GPT-Image-1.5~\cite{openai2025gpt_image15}
& 0.234 & 0.505 & 0.388 & 0.760 & 0.932 & 0.027 \\
GPT-Image-2.0~\cite{openai2026gpt_image2}
& 0.168 & 0.370 & 0.288 & 0.838 & 0.952 & 0.272 \\
WAN2.7-Image~\cite{wan27_image}
& 0.166 & 0.359 & 0.278 & 0.862 & 0.962 & 0.182 \\
Doubao Seedream 5.0~\cite{seedream5_lite}
& 0.171 & 0.458 & 0.352 & 0.796 & 0.951 & 0.136 \\
\midrule
Depth Anything 3~\cite{depth_anything3}
& \textbf{0.035} & \textbf{0.117} & \textbf{0.062} & \textbf{0.981} & \textbf{0.994} & 0.240 \\
Lotus~\cite{Lotus}
& 0.048 & 0.143 & 0.091 & 0.978 & \textbf{0.994} & \textbf{0.318} \\
\bottomrule
\end{tabular}
\end{table}

For \textbf{depth}, Table~\ref{tab:depth_mainaxis_combined_all} shows affine-invariant relative-depth results.
Scale, shift, and polarity are handled only at scoring time, so scores reflect relative scene geometry rather than metric depth.
Specialized predictors still dominate the scalar relative-depth metrics: \textbf{Depth Anything 3} gives the best AbsRel-AI, RMSE-AI, MAE-AI, and threshold accuracies, while \textbf{Lotus} gives the best Boundary F1.
Among proprietary editors, \textbf{WAN2.7-Image} gives the best AbsRel-AI point estimate, whereas \textbf{GPT-Image-2.0} gives the best Boundary F1 and substantially improves over \textbf{GPT-Image-1.5}.
This split indicates that scalar relative-depth accuracy and local discontinuity recovery favor different methods.
Additional ordinal, boundary, and bootstrap diagnostics are reported in Appendix~\ref{app:depth_alignment_diagnostics} and Appendix Table~\ref{tab:depth_absrel_bootstrap_ci}.

\begin{table}[!htbp]
\centering
\caption{Source-balanced normal estimation results under the fixed three-exemplar normal setting.}
\label{tab:normal_mainaxis_combined_all}
\footnotesize
\setlength{\tabcolsep}{5.5pt}
\renewcommand{\arraystretch}{0.95}
\begin{tabular}{lccccc}
\toprule
Method & Mean$\downarrow$ & Median$\downarrow$ & Acc@11.25$\uparrow$ & Acc@22.5$\uparrow$ & Acc@30$\uparrow$ \\
\midrule
GPT-Image-1.5~\cite{openai2025gpt_image15}
& 44.803 & 42.517 & 0.057 & 0.204 & 0.325 \\
Doubao Seedream 5.0~\cite{seedream5_lite}
& 56.243 & 50.157 & 0.035 & 0.132 & 0.227 \\
\midrule
Lotus~\cite{Lotus}
& 11.679 & \textbf{7.305} & 0.770 & 0.863 & 0.888 \\
StableNormal~\cite{stablenormal}
& \textbf{11.216} & 7.699 & \textbf{0.777} & \textbf{0.876} & \textbf{0.900} \\
\bottomrule
\end{tabular}
\end{table}

For \textbf{normal}, Table~\ref{tab:normal_mainaxis_combined_all} shows source-balanced results under the fixed three-exemplar setting.
Editor scores are averaged over the three exemplar-conditioned runs, with per-exemplar results reported in Appendix~\ref{app:normal_exemplar_diagnostics}.
\textbf{StableNormal} and \textbf{Lotus} substantially outperform proprietary editors, suggesting that exemplar pairs help stabilize the normal-map convention but do not close the dense directional-recovery gap. Scene-cluster bootstrap intervals for Acc@22.5 preserve this separation; full intervals are reported in Appendix Table~\ref{tab:normal_acc225_bootstrap_ci}. \textbf{GPT-Image-2.0} is excluded from the official normal aggregate because no complete retained-split run satisfying the full three-exemplar protocol was completed before the manifest was frozen. \textbf{WAN2.7-Image} is treated as a representation failure rather than a scored entry, because its probes produced stylized or normal-map-like renderings instead of spatially aligned RGB-encoded normal fields; examples are shown in Appendix~\ref{app:normal_failure_cases}, Figure~\ref{fig:wan_normal_failure_cases}.

\begin{table}[!htbp]
\centering
\caption{Source-balanced albedo results under the disclosed auxiliary-analysis-assisted, illumination-aware structure-preserving albedo elicitation protocol.}
\label{tab:albedo_mainaxis_combined_all}
\footnotesize
\setlength{\tabcolsep}{6pt}
\renewcommand{\arraystretch}{0.95}
\begin{tabular}{lcccc}
\toprule
Method & MAE$\downarrow$ & PSNR$\uparrow$ & SSIM$\uparrow$ & LPIPS$\downarrow$ \\
\midrule
GPT-Image-1.5~\cite{openai2025gpt_image15}
& 0.204 & 11.983 & 0.521 & 0.443 \\
GPT-Image-2.0~\cite{openai2026gpt_image2}
& 0.174 & 13.519 & 0.554 & 0.413 \\
WAN2.7-Image~\cite{wan27_image}
& 0.220 & 11.867 & 0.583 & 0.412 \\
Doubao Seedream 5.0~\cite{seedream5_lite}
& 0.183 & 13.126 & 0.606 & 0.351 \\
\midrule
Colorful Diffuse Intrinsic~\cite{colorful_diffuse_intrinsic}
& 0.138 & 16.137 & 0.744 & 0.337 \\
DNF-Intrinsic~\cite{dnf_intrinsic}
& \textbf{0.094} & \textbf{19.217} & \textbf{0.758} & \textbf{0.175} \\
\bottomrule
\end{tabular}
\end{table}

For \textbf{albedo}, Table~\ref{tab:albedo_mainaxis_combined_all} reports results under the auxiliary-analysis-assisted setting.
The auxiliary analysis is generated only from the query RGB image and reused unchanged across proprietary systems; it is not a benchmark label, ground-truth signal, model-output feedback, or test-set metric cue. On the source-balanced main-axis aggregate, \textbf{DNF-Intrinsic} is strongest across MAE, PSNR, SSIM, and LPIPS. Among proprietary editors, \textbf{GPT-Image-2.0} gives the best MAE and PSNR, while \textbf{Doubao Seedream 5.0} gives the best SSIM and LPIPS. This split shows that scalar distortion and structural/perceptual fidelity can favor different editors.
Scene-cluster bootstrap intervals preserve the MAE ordering; full intervals and protocol-sensitivity variants are reported in Appendix Table~\ref{tab:albedo_mae_bootstrap_ci} and Appendix~\ref{app:diagnostic_results}.

\begin{table}[!htbp]
\centering
\caption{Controlled RGB-to-roughness results on the OpenRooms-FF main-axis subset.}
\label{tab:roughness_mainaxis_openrooms}
\footnotesize
\setlength{\tabcolsep}{6pt}
\renewcommand{\arraystretch}{0.95}
\begin{tabular}{lcccc}
\toprule
Method & RMSE$\downarrow$ & MAE$\downarrow$ & SSIM$\uparrow$ & PSNR$\uparrow$ \\
\midrule
GPT-Image-1.5~\cite{openai2025gpt_image15}
& 0.375 & 0.321 & 0.399 & 9.103 \\
GPT-Image-2.0~\cite{openai2026gpt_image2}
& \textbf{0.323} & \textbf{0.285} & 0.444 & \textbf{10.317} \\
Qwen-Image-2.0~\cite{qwen_image2}
& 0.384 & 0.324 & 0.397 & 8.593 \\
Doubao Seedream 5.0~\cite{seedream5_lite}
& 0.364 & 0.310 & 0.348 & 9.167 \\
\midrule
Intrinsic Image Diffusion~\cite{intrinsic_image_diffusion}
& 0.346 & 0.308 & \textbf{0.513} & 9.814 \\
DNF-Intrinsic~\cite{dnf_intrinsic}
& 0.376 & 0.324 & 0.478 & 9.090 \\
\bottomrule
\end{tabular}
\end{table}

For \textbf{roughness}, Table~\ref{tab:roughness_mainaxis_openrooms} reports controlled RGB-to-roughness results on OpenRooms-FF.
\textbf{GPT-Image-2.0} gives the best RMSE point estimate, whereas \textbf{Intrinsic Image Diffusion} gives substantially higher SSIM.
This scalar--structural split indicates that lower roughness error does not necessarily preserve material-region structure.
Scene-cluster paired-bootstrap intervals preserve this split; full intervals are reported in Appendix Table~\ref{tab:roughness_mainaxis_ci}.

\begin{table}[!htbp]
\centering
\caption{Controlled RGB-to-metallic continuous-map results on the companion main-axis subset.}
\label{tab:metallic_mainaxis_companion}
\footnotesize
\setlength{\tabcolsep}{7pt}
\renewcommand{\arraystretch}{0.95}
\begin{tabular}{lcccc}
\toprule
Method & MAE$\downarrow$ & PSNR$\uparrow$ & SSIM$\uparrow$ & LPIPS$\downarrow$ \\
\midrule
GPT-Image-1.5~\cite{openai2025gpt_image15}
& 0.055 & 17.423 & 0.871 & 0.181 \\
GPT-Image-2.0~\cite{openai2026gpt_image2}
& \textbf{0.051} & \textbf{18.242} & \textbf{0.933} & \textbf{0.137} \\
WAN2.7-Image~\cite{wan27_image}
& 0.120 & 13.203 & 0.482 & 0.508 \\
Doubao Seedream 5.0~\cite{seedream5_lite}
& 0.073 & 16.478 & 0.516 & 0.365 \\
\midrule
MVInverse~\cite{mvinverse}
& 0.211 & 11.189 & 0.119 & 0.518 \\
DNF-Intrinsic~\cite{dnf_intrinsic}
& 0.150 & 13.807 & 0.236 & 0.499 \\
\bottomrule
\end{tabular}
\end{table}

\FloatBarrier

For \textbf{metallic}, Table~\ref{tab:metallic_mainaxis_companion} reports continuous-map results on the companion source. \textbf{GPT-Image-2.0} gives the best MAE, PSNR, SSIM, and LPIPS point estimates.
However, scene-cluster bootstrap intervals for metallic MAE overlap among the top proprietary editors, so this ranking should be read as a point-estimate ordering rather than a statistically separated top-model claim. Because metallic maps are sparse and background-dominated, these scores measure masked continuous-map fidelity rather than explicit metallic-region detection.

Overall, specialized predictors retain clear advantages on the primary metrics for depth, normal, and albedo. Roughness and metallic differ: proprietary editors achieve the best primary scalar scores under the controlled material-map settings. At the same time, additional metrics show that scalar error, structural consistency, and sparse-material behavior can favor different methods.

\vspace{0.05in}\noindent\textbf{Source-wise consistency.}
For depth, normal, and albedo, the main tables report source-balanced averages over OpenRooms-FF and InteriorVerse; Appendix~\ref{app:source_specific_results} reports the two sources separately. These source-specific results preserve the main trend: specialized predictors remain stronger than proprietary editors on both sources, although the gap varies by target and source.
Roughness and metallic each use one retained official source, so their main tables are already source-specific.

\subsection{Photometric Stress Analysis}
\label{sec:stress_results}

Photometric stress tests examine whether main-axis trends remain stable under challenging illumination. Before scoring, we define stress subsets from image-level statistics, covering low-light, high-dynamic-range, highlight-heavy, and dark-region-dominant images. These subsets are fixed independently of model errors or metric gaps. Each slice uses the same target-specific metrics, masks, and aggregation rules as the corresponding main-axis evaluation.

The stress results are diagnostic rather than a second leaderboard.
For targets evaluated on both OpenRooms-FF and InteriorVerse, we report source-balanced stress scores only when both sources have sufficient slice support; otherwise, the slice is reported source-specifically. Table~\ref{tab:stress_summary_main} summarizes representative per-target behavior, with complete slice-wise results in Appendix~\ref{app:stress_results}.

\begin{table}[!htbp]
\centering
\scriptsize
\caption{Compact summary of photometric stress-subset behavior.
Rows use pre-defined stress slices and the same metrics and aggregation rules as the corresponding main-axis evaluation.
Complete slice-wise results are reported in Appendix~\ref{app:stress_results}.}
\label{tab:stress_summary_main}
\setlength{\tabcolsep}{4pt}
\renewcommand{\arraystretch}{1.15}

\begin{tabularx}{\linewidth}{@{}L{2.25cm} X@{}}
\toprule
Target / slice & Key observation under stress \\
\midrule

\textbf{Depth}\newline
HDR, OR-FF+IV
&
On the HDR slice, \textbf{Depth Anything 3} gives the best scalar relative-depth scores
\((0.036/0.124/0.069\) for AbsRel-AI/RMSE-AI/MAE-AI), while \textbf{Lotus Depth} gives the best Boundary F1 \((0.341)\).
Thus, scalar relative-depth accuracy and local discontinuity recovery remain distinct stress behaviors. \\

\midrule

\textbf{Normal}\newline
HDR, OR-FF+IV
&
The geometry gap persists under HDR stress:
\textbf{StableNormal} achieves the best Acc@22.5 \((0.885)\), followed by \textbf{Lotus} \((0.872)\), while \textbf{GPT-Image-1.5} remains much lower \((0.205)\).
Exemplar conditioning does not remove the dense directional-recovery gap. \\

\midrule

\textbf{Albedo}\newline
Low-light, OR-FF
&
Low-light albedo shows a source-specific ranking change:
\textbf{Colorful Diffuse Intrinsic} outperforms \textbf{DNF-Intrinsic} on MAE/PSNR/SSIM
\((0.124/16.516/0.660\) vs.\ \(0.189/13.582/0.519)\), despite \textbf{DNF-Intrinsic} leading the source-balanced main-axis table.
This supports reporting source-specific stress slices. \\

\midrule

\textbf{Roughness}\newline
HDR, OR-FF
&
HDR roughness preserves the scalar--structural split. \textbf{Intrinsic Image Diffusion} gives the best RMSE and SSIM \((0.286/0.517)\), while the main-axis scalar winner \textbf{GPT-Image-2.0} is close in RMSE but lower in SSIM \((0.289/0.449)\).
Lower scalar error does not necessarily preserve material-region structure. \\

\midrule

\textbf{Metallic}\newline
mix\_temperature, Comp.
&
On the companion mixed-temperature slice, \textbf{GPT-Image-2.0} gives the best continuous-map MAE/PSNR/SSIM/LPIPS
\((0.035/20.785/0.954/0.112)\).
Because metallic pixels are sparse, this measures continuous-map fidelity over the evaluated mask rather than calibrated metallic-region detection. \\

\bottomrule
\end{tabularx}
\vspace{-0.3em}
\end{table}

Overall, stress results affect not only score magnitudes but also metric-wise rankings. They preserve the large geometry gap for depth and normal, expose source- or metric-specific changes for albedo and roughness, and require sparsity-aware interpretation for metallic.
Complete slice-wise tables are provided in Appendix~\ref{app:stress_results}.

\subsection{Access-Cue Diagnostics}
\label{sec:diag_results}

The main results use one fixed access setting per target.
We also run controlled diagnostic variants to isolate how prompt strength, exemplars, image-derived analysis, and spatial priors affect editor outputs.
These variants are diagnostic rather than used to choose the official setting; full tables are reported in Appendix~\ref{app:diagnostic_results}.

For \textbf{depth}, adding a fixed RGB--depth exemplar improves scalar and ordinal scores on the diagnostic subset but reduces Boundary F1.
A segmentation reference does not consistently improve over the controlled RGB-to-depth prompt.
Thus, extra cues shift the recovered depth structure rather than uniformly improving all depth diagnostics.

For \textbf{normal}, removing the fixed RGB--normal exemplars increases Mean Angular from 57.00 to 71.34 and reduces Acc@22.5 from 0.1181 to 0.0672. The exemplars therefore mainly stabilize the normal-map convention, rather than providing query-specific supervision.

For \textbf{albedo}, query-image-derived illumination analysis slightly improves MAE, SSIM, and LPIPS relative to the same structure-preserving prompt without analysis. Weaker prompts can improve some scalar scores while relaxing the illumination-aware constraint, so these diagnostics test the role of explicit lighting cues rather than selecting a universally best prompt.

For \textbf{roughness}, segmentation-assisted region filling improves RMSE and SSIM for Doubao Seedream 5.0 on the diagnostic subset, suggesting that coarse spatial priors help material-map structure.
Because this changes direct RGB-to-map prediction into a spatial-prior-assisted pipeline, it remains separate from the controlled roughness comparison.

For \textbf{metallic}, prior-assisted variants for Doubao Seedream 5.0 on the 120-image companion diagnostic subset show metric-dependent behavior. Soft segmentation improves MAE and SSIM relative to the controlled RGB-to-metallic setting, while segmentation-based region filling gives the best RMSE, LPIPS, and PSNR. The fixed RGB--metallic exemplar does not improve over the controlled setting. As with roughness, these variants add spatial or exemplar cues beyond the controlled RGB-only setting and are therefore reported only as diagnostics.

Overall, access-cue diagnostics show that allowed cues can materially change editor outputs, especially for material-map targets. The main tables therefore keep one fixed target-specific access setting per target, while cue variants are reported separately as diagnostic evidence.

\section{Limitations}

PhysEditBench uses fixed access settings to make systems comparable, but this excludes many possible prompts, exemplar choices, and multi-turn interactions.
The diagnostic variants test several common cues, but they are not an exhaustive prompt search.
Each score should therefore be read as performance under the evaluated protocol, not as the maximum capability of the model.

The evaluated model set is limited to completed and scoreable retained-split runs.
GPT-Image-2.0 is included for depth, albedo, roughness, and metallic, but the full three-exemplar normal run was not completed before the manifest was frozen and is reported only as diagnostic.

PhysEditBench is limited to synthetic indoor sources and a procedurally generated companion source for metallic evaluation.
The results therefore measure controlled benchmark performance rather than downstream utility or full real-world generalization.

\section{Reproducibility and Release}

We release materials to support auditing and reproduction of the benchmark evaluation procedure.
The review-time package includes evaluation code, aggregation scripts, prompt templates, access-setting specifications, exemplar identifiers, and related evaluation metadata.
For the companion source used in metallic evaluation, we provide source metadata, channel specifications, retained-subset statistics, and quality-control records.

Some retained image lists, sampled data, companion-source files, and proprietary image-editor outputs may not be redistributable during review.
For these cases, we provide audit records that allow the reported scores and sampling design to be checked against the frozen evaluation setup.
These records include prompts, model versions, output metadata, file hashes when available, post-processing records, evaluator logs, failure logs, and split-level summary statistics.

All review-time links are anonymized.
Upon acceptance, we will release the finalized benchmark documentation, evaluation code, prompt and access-setting files, valid-region definitions, stress-subset definitions, companion-source metadata, split statistics, and aggregated scoring artifacts.

\section{Conclusion}

We introduced \textbf{PhysEditBench}, a benchmark for evaluating whether general-purpose image editors can predict dense physical maps from a single indoor RGB image without task-specific fine-tuning.
PhysEditBench covers five targets: depth, normal, albedo, roughness, and metallic.
It evaluates editors under fixed target-specific access settings, with per-target data sources, valid masks, metrics, aggregation rules, and photometric stress subsets.

Our results show that image editors can often produce outputs that look like physical maps, but visual plausibility does not guarantee physical accuracy.
Specialized predictors remain stronger on the primary metrics for depth, normal, and albedo.
Among proprietary editors, \textbf{GPT-Image-2.0} improves over \textbf{GPT-Image-1.5} in the completed depth, albedo, roughness, and metallic evaluations, showing that stronger image editors can produce more faithful map predictions.
For roughness and metallic, proprietary editors can match or exceed specialized baselines on some primary scalar metrics.
However, auxiliary metrics, protocol diagnostics, and stress subsets reveal limitations in structure preservation, sparse material prediction, and robustness under challenging illumination.

These findings support reporting results separately for each target under disclosed access settings, rather than reducing all systems to a single overall ranking.
PhysEditBench highlights where current image editors are becoming competitive, where dedicated predictors still dominate, and how editor performance changes with the target, metric, protocol, and image condition.

\bibliographystyle{plainnat}
\bibliography{reference}

@inproceedings{openrooms,
  title     = {OpenRooms: An Open Framework for Photorealistic Indoor Scene Datasets},
  author    = {Li, Zhengqin and Yu, Ting-Wei and Sang, Shen and Wang, Sarah and Song, Meng and Liu, Yuhan and Yeh, Yu-Ying and Zhu, Rui and Gundavarapu, Nitesh Bharadwaj and Shi, Jia and Bi, Sai and Yu, Hong-Xing and Xu, Zexiang and Sunkavalli, Kalyan and Ha\v{s}an, Milo\v{s} and Ramamoorthi, Ravi and Chandraker, Manmohan},
  booktitle = {Proceedings of the IEEE/CVF Conference on Computer Vision and Pattern Recognition (CVPR)},
  pages     = {7190--7199},
  year      = {2021},
  doi       = {10.1109/CVPR46437.2021.00711}
}

@inproceedings{openrooms_ff,
  title     = {MAIR: Multi-view Attention Inverse Rendering with 3D Spatially-Varying Lighting Estimation},
  author    = {Choi, JunYong and Lee, SeokYeong and Park, Haesol and Jung, Seung-Won and Kim, Ig-Jae and Cho, Junghyun},
  booktitle = {Proceedings of the IEEE/CVF Conference on Computer Vision and Pattern Recognition (CVPR)},
  year      = {2023},
  note      = {Introduces the OpenRooms Forward Facing (OpenRooms FF) dataset}
}

@inproceedings{interiorverse,
  title     = {Learning-Based Inverse Rendering of Complex Indoor Scenes with Differentiable Monte Carlo Raytracing},
  author    = {Zhu, Jingsen and Luan, Fujun and Huo, Yuchi and Lin, Zihao and Zhong, Zhihua and Xi, Dianbing and Wang, Rui and Bao, Hujun and Zheng, Jiaxiang and Tang, Rui},
  booktitle = {SIGGRAPH Asia 2022 Conference Papers},
  year      = {2022},
  doi       = {10.1145/3550469.3555407}
}

@inproceedings{hypersim,
  title     = {Hypersim: A Photorealistic Synthetic Dataset for Holistic Indoor Scene Understanding},
  author    = {Roberts, Mike and Ramapuram, Jason and Ranjan, Anurag and Kumar, Atulit and Bautista, Miguel Angel and Paczan, Nathan and Webb, Russ and Susskind, Joshua M.},
  booktitle = {Proceedings of the IEEE/CVF International Conference on Computer Vision (ICCV)},
  year      = {2021}
}

@inproceedings{iris_cvpr2025,
  title     = {IRIS: Inverse Rendering of Indoor Scenes from Low Dynamic Range Images},
  author    = {Lin, Chih-Hao and Huang, Jia-Bin and Li, Zhengqin and Dong, Zhao and Richardt, Christian and Li, Tuotuo and Zollh\"ofer, Michael and Kopf, Johannes and Wang, Shenlong and Kim, Changil},
  booktitle = {Proceedings of the IEEE/CVF Conference on Computer Vision and Pattern Recognition (CVPR)},
  pages     = {465--474},
  year      = {2025},
  doi       = {10.1109/CVPR52734.2025.00052}
}

@inproceedings{rgbx_arxiv,
  title     = {RGB$\leftrightarrow$X: Image Decomposition and Synthesis Using Material- and Lighting-Aware Diffusion Models},
  author    = {Zeng, Zheng and Deschaintre, Valentin and Georgiev, Iliyan and Hold-Geoffroy, Yannick and Hu, Yiwei and Luan, Fujun and Yan, Ling-Qi and Ha\v{s}an, Milo\v{s}},
  booktitle = {ACM SIGGRAPH 2024 Conference Papers},
  year      = {2024},
  doi       = {10.1145/3641519.3657445}
}

@misc{materialist,
  title        = {Materialist: Physically Based Editing Using Single-Image Inverse Rendering},
  author       = {Wang, Lezhong and Tran, Duc Minh and Cui, Ruiqi and TG, Thomson and Dahl, Anders Bjorholm and Bigdeli, Siavash Arjomand and Frisvad, Jeppe Revall and Chandraker, Manmohan},
  year         = {2025},
  eprint       = {2501.03717},
  archivePrefix= {arXiv},
  primaryClass = {cs.CV},
  url          = {https://arxiv.org/abs/2501.03717}
}

@inproceedings{dnf_intrinsic,
  title     = {DNF-Intrinsic: Deterministic Noise-Free Diffusion for Indoor Inverse Rendering},
  author    = {Zheng, Rongjia and Zhang, Qing and Long, Chengjiang and Zheng, Wei-Shi},
  booktitle = {Proceedings of the IEEE/CVF International Conference on Computer Vision (ICCV)},
  year      = {2025}
}

@inproceedings{Lotus,
  title     = {Lotus: Diffusion-based Visual Foundation Model for High-quality Dense Prediction},
  author    = {He, Jing and Li, Haodong and Yin, Wei and Liang, Yixun and Li, Leheng and Zhou, Kaiqiang and Zhang, Hongbo and Liu, Bingbing and Chen, Ying-Cong},
  booktitle = {International Conference on Learning Representations (ICLR)},
  year      = {2025}
}

@article{stablenormal,
  title     = {StableNormal: Reducing Diffusion Variance for Stable and Sharp Normal},
  author    = {Ye, Chongjie and Qiu, Lingteng and Gu, Xiaodong and Zuo, Qi and Wu, Yushuang and Dong, Zilong and Bo, Liefeng and Xiu, Yuliang and Han, Xiaoguang},
  journal   = {ACM Transactions on Graphics (TOG)},
  volume    = {43},
  number    = {6},
  pages     = {1--18},
  year      = {2024},
  doi       = {10.1145/3687971}
}

@inproceedings{dsine,
  title     = {Rethinking Inductive Biases for Surface Normal Estimation},
  author    = {Bae, Gwangbin and Davison, Andrew J.},
  booktitle = {Proceedings of the IEEE/CVF Conference on Computer Vision and Pattern Recognition (CVPR)},
  year      = {2024}
}

@inproceedings{intrinsic_image_diffusion,
  title     = {Intrinsic Image Diffusion for Indoor Single-view Material Estimation},
  author    = {Kocsis, Peter and Sitzmann, Vincent and Nie\ss ner, Matthias},
  booktitle = {Proceedings of the IEEE/CVF Conference on Computer Vision and Pattern Recognition (CVPR)},
  year      = {2024}
}

@misc{pbr_boost_3dgen,
  title        = {Boosting 3D Object Generation through PBR Materials},
  author       = {Wang, Yao and others},
  year         = {2024},
  eprint       = {2411.16080},
  archivePrefix= {arXiv},
  primaryClass = {cs.CV},
  url          = {https://arxiv.org/abs/2411.16080}
}

@inproceedings{ir3d_bench,
  title     = {IR3D-Bench: Evaluating Vision-Language Model Scene Understanding as Agentic Inverse Rendering},
  author    = {Liu, Hengyu and Li, Chenxin and Li, Zhengxin and Wu, Yipeng and Li, Wuyang and Yang, Zhiqin and Zhang, Zhenyuan and Lin, Yunlong and Han, Sirui and Feng, Brandon Y.},
  booktitle = {Advances in Neural Information Processing Systems, Track on Datasets and Benchmarks},
  year      = {2025}
}

@article{epbr,
  author        = {Yu Guo and Zhiqiang Lao and Xiyun Song and Yubin Zhou and Zongfang Lin and Heather Yu},
  title         = {ePBR: Extended PBR Materials in Image Synthesis},
  journal       = {arXiv preprint arXiv:2504.17062},
  year          = {2025},
  eprint        = {2504.17062},
  archivePrefix = {arXiv},
  primaryClass  = {cs.GR}
}

@misc{vision_banana,
  title        = {Image Generators are Generalist Vision Learners},
  author       = {Valentin Gabeur and Shangbang Long and Songyou Peng and Paul Voigtlaender and Shuyang Sun and Yanan Bao and Karen Truong and Zhicheng Wang and Wenlei Zhou and Jonathan T. Barron and Kyle Genova and Nithish Kannen and Sherry Ben and Yandong Li and Mandy Guo and Suhas Yogin and Yiming Gu and Huizhong Chen and Oliver Wang and Saining Xie and Howard Zhou and Kaiming He and Thomas Funkhouser and Jean-Baptiste Alayrac and Radu Soricut},
  year         = {2026},
  eprint       = {2604.20329},
  archivePrefix= {arXiv},
  primaryClass = {cs.CV},
  note         = {Accepted by CVPR 2026}
}

@inproceedings{intrinsicanything,
  title     = {IntrinsicAnything: Learning Diffusion Priors for Inverse Rendering Under Unknown Illumination},
  author    = {Chen, Xi and Peng, Sida and Yang, Dongchen and Liu, Yuan and Pan, Bowen and Lv, Chengfei and Zhou, Xiaowei},
  booktitle = {European Conference on Computer Vision},
  year      = {2024}
}

@article{mvinverse,
  title   = {{MVInverse}: Feed-forward Multi-view Inverse Rendering in Seconds},
  author  = {Wu, Xiangzuo and Ren, Chengwei and Zhou, Jun and Li, Xiu and Liu, Yuan},
  journal = {arXiv preprint arXiv:2512.21003},
  year    = {2025}
}

@inproceedings{infinigen,
  title     = {Infinite Photorealistic Worlds using Procedural Generation},
  author    = {Raistrick, Alexander and Lipson, Lahav and Ma, Zeyu and Mei, Lingjie and Wang, Mingzhe and Zuo, Yiming and Kayan, Karhan and Wen, Hongyu and Han, Beining and Wang, Yihan and Newell, Alejandro and Law, Hei and Goyal, Ankit and Yang, Kaiyu and Deng, Jia},
  booktitle = {Proceedings of the IEEE/CVF Conference on Computer Vision and Pattern Recognition (CVPR)},
  pages     = {12630--12641},
  year      = {2023}
}

@inproceedings{infinigen_indoors,
  title     = {Infinigen Indoors: Photorealistic Indoor Scenes using Procedural Generation},
  author    = {Raistrick, Alexander and Mei, Lingjie and Kayan, Karhan and Yan, David and Zuo, Yiming and Han, Beining and Wen, Hongyu and Parakh, Meenal and Alexandropoulos, Stamatis and Lipson, Lahav and Ma, Zeyu and Deng, Jia},
  booktitle = {Proceedings of the IEEE/CVF Conference on Computer Vision and Pattern Recognition (CVPR)},
  pages     = {21783--21794},
  year      = {2024}
}

@misc{google2025gemini_flash_image,
  title        = {Introducing Gemini 2.5 Flash Image, Google's state-of-the-art image generation and editing model},
  author       = {{Google}},
  year         = {2025},
  howpublished = {\url
  {https://developers.googleblog.com/en/introducing-gemini-2-5-flash-image/}},
  note         = {Accessed: 2026-05-06}
}

@misc{openai2026gpt_image2,
  title        = {GPT Image 2 Model},
  author       = {{OpenAI}},
  year         = {2026},
  howpublished = {\url{https://developers.openai.com/api/docs/models/gpt-image-2}},
  note         = {Accessed: 2026-05-06}
}

@inproceedings{brooks2023instructpix2pix,
  title     = {InstructPix2Pix: Learning to Follow Image Editing Instructions},
  author    = {Brooks, Tim and Holynski, Aleksander and Efros, Alexei A.},
  booktitle = {Proceedings of the IEEE/CVF Conference on Computer Vision and Pattern Recognition},
  year      = {2023}
}

@inproceedings{zhang2023magicbrush,
  title     = {MagicBrush: A Manually Annotated Dataset for Instruction-Guided Image Editing},
  author    = {Zhang, Kai and Mo, Lingbo and Chen, Wenhu and Sun, Huan and Su, Yu},
  booktitle = {Advances in Neural Information Processing Systems},
  year      = {2023}
}

@inproceedings{sheynin2024emu_edit,
  title     = {Emu Edit: Precise Image Editing via Recognition and Generation Tasks},
  author    = {Sheynin, Shelly and Polyak, Adam and Singer, Uriel and Kirstain, Yuval and Zohar, Amit and Ashual, Oron and Parikh, Devi and Taigman, Yaniv},
  booktitle = {Proceedings of the IEEE/CVF Conference on Computer Vision and Pattern Recognition},
  year      = {2024}
}

@misc{openai2025gpt_image15,
  title        = {GPT Image 1.5 Model},
  author       = {{OpenAI}},
  year         = {2025},
  howpublished = {\url{https://developers.openai.com/api/docs/models/gpt-image-1.5}},
  note         = {Accessed: 2026-05-06}
}

@misc{wan27_image,
  title         = {Wan-Image: Pushing the Boundaries of Generative Visual Intelligence},
  author        = {Mao, Chaojie and Xie, Chen-Wei and Zhong, Chongyang and Deng, Haoyou and Zhao, Jiaxing and Xiao, Jie and Xing, Jinbo and Zhang, Jingfeng and Zhou, Jingren and Zhang, Jingyi and others},
  year          = {2026},
  eprint        = {2604.19858},
  archivePrefix = {arXiv},
  primaryClass  = {cs.CV}
}

@misc{seedream5_lite,
  title        = {Seedream 5.0 Lite},
  author       = {{ByteDance Seed}},
  year         = {2026},
  howpublished = {\url{https://seed.bytedance.com/en/seedream5_0_lite}},
  note         = {Accessed: 2026-05-06}
}

@misc{qwen_image2,
  title        = {Qwen-Image-2.0: Professional Infographics, Exquisite Text Rendering, and Unified Generation + Editing},
  author       = {{Qwen Team}},
  year         = {2026},
  howpublished = {\url{https://qwen.ai/blog?id=qwen-image-2.0}},
  note         = {Accessed: 2026-05-06}
}

@misc{depth_anything3,
  title        = {Depth Anything 3: Recovering the Visual Space from Any Views},
  author       = {Lin, Haotong and Chen, Sili and Liew, Junhao and Chen, Donny Y. and Li, Zhenyu and Shi, Guang and Feng, Jiashi and Kang, Bingyi},
  year         = {2025},
  eprint       = {2511.10647},
  archivePrefix= {arXiv},
  primaryClass = {cs.CV},
  url          = {https://arxiv.org/abs/2511.10647}
}

@article{colorful_diffuse_intrinsic,
  title     = {Colorful Diffuse Intrinsic Image Decomposition in the Wild},
  author    = {Careaga, Chris and Aksoy, Ya\u{g}{\i}z},
  journal   = {ACM Transactions on Graphics},
  volume    = {43},
  number    = {6},
  year      = {2024},
  doi       = {10.1145/3687984}
}

\clearpage
\appendix

\section{Dataset Documentation and Source Reliability}
\label{app:dataset_docs}

This appendix documents the datasets used by the benchmark, their target coverage, source-specific supervision constraints, and target-dependent source-reliability decisions. 
The main paper summarizes the source design; here we provide source-level statistics and the rationale for retaining or excluding each source for each target.

\subsection{Public Source Datasets}
\label{app:source_stats}

\subsubsection{OpenRooms and OpenRooms-FF}

OpenRooms is a synthetic indoor inverse-rendering dataset constructed from ScanNet-based CAD scenes. It assigns randomized materials to object parts in the scene and renders HDR indoor images with detailed material and lighting annotations. In addition, OpenRooms provides multiple discrete rendering subsets for the same indoor scenes, including \texttt{main\_xml}, \texttt{main\_xml1}, \texttt{mainDiffMat\_xml}, \texttt{mainDiffMat\_xml1}, \texttt{mainDiffLight\_xml}, and \texttt{mainDiffLight\_xml1}. Within each paired configuration, \texttt{main} and \texttt{mainDiffMat} preserve the lighting while changing materials, whereas \texttt{main} and \texttt{mainDiffLight} preserve the materials while changing lighting. Therefore, OpenRooms provides discrete material and illumination variations rather than continuous parametric control.

OpenRooms-FF is a multi-view extension built on top of OpenRooms. Starting from selected valid images in OpenRooms, the dataset constructs a local $3 \times 3$ view set for each reference image by translating the camera to eight neighboring directions while keeping a consistent viewing direction. As a result, OpenRooms-FF provides richer local multi-view observations together with corresponding material and lighting annotations.

\subsubsection{InteriorVerse}

InteriorVerse is a large-scale indoor inverse-rendering dataset with scenes designed by professional artists. Compared with OpenRooms-style synthetic scenes, InteriorVerse contains richer scene details and more diverse indoor objects. The dataset includes a large number of HDR images and provides synthetic ground-truth material and geometry annotations. Its scene diversity and annotation coverage make it a valuable complementary source for benchmark construction.

\begin{table}[!htbp]
\centering
\small
\caption{Basic statistics of the public source datasets used in this work.}
\label{tab:source_dataset_overview}
\begin{tabular}{lccc}
\toprule
Dataset & Scene count & Image / view count & Resolution \\
\midrule
OpenRooms-FF & 1248 & $9 \times 23618 = 212562$ & $640 \times 480$ \\
InteriorVerse & 4176 & 52769 & $640 \times 480$ \\
\bottomrule
\end{tabular}
\end{table}

\subsection{Target Coverage and Benchmark Roles}
\label{app:dataset_target_coverage}

Although all three sources contain material-related information, their target coverage and annotation reliability differ substantially. InteriorVerse provides relatively comprehensive material annotations, but its roughness and metallic maps are excluded from official scoring due to material-channel reliability concerns. OpenRooms-FF provides supervision for several core inverse-rendering targets, but does not include metallic maps. Therefore, different datasets play different roles depending on the prediction target.

\begin{table}[!htbp]
\centering
\small
\caption{Dataset-by-target coverage and benchmark roles.}
\label{tab:dataset_target_coverage}
\begin{tabular}{lccccc l}
\toprule
Source & Depth & Normal & Albedo & Roughness & Metallic & Benchmark role \\
\midrule
OpenRooms-FF & Yes & Yes & Yes & Yes & No & Primary source for D/N/A/R \\
InteriorVerse & Yes & Yes & Yes & Excluded & Excluded & Supplementary source for D/N/A \\
Companion$^\dagger$ & -- & -- & -- & -- & Yes & Official metallic source \\
\bottomrule
\end{tabular}

\tabremark{
$^\dagger$ The companion source exports additional aligned inverse-rendering channels for quality control and diagnostics, but in this benchmark it is used only as the official source for metallic evaluation.
}
\end{table}

\subsection{Companion Dataset for Metallic Evaluation}
\label{app:companion_dataset}

The companion source is released as \textbf{PBR-Rooms}, a synthetic indoor PBR benchmark constructed for controlled inverse-rendering evaluation.
It is introduced in this paper because OpenRooms-FF does not provide metallic supervision and InteriorVerse metallic maps are excluded from official scoring due to the reliability concerns discussed in Appendix~\ref{app:iv_material_reliability}.
Although PBR-Rooms provides multiple aligned inverse-rendering channels, its official role in this benchmark is deliberately restricted to metallic evaluation.
This design fills the metallic-supervision gap without changing the public-source evaluation protocol for depth, normal, albedo, and roughness.

\subsubsection{Motivation and Benchmark Role}

Existing public indoor inverse-rendering datasets do not fully satisfy the requirements of benchmark-grade metallic evaluation.
Some sources do not provide pixel-aligned metallic maps, while others contain material channels whose values are visually inconsistent with the intended physical material parameters.
In addition, many indoor scenes contain sparse visible metallic regions, making metallic evaluation easily dominated by non-metallic background pixels.
These limitations are problematic for dense metallic prediction, because a model can obtain deceptively low scalar error by predicting mostly non-metallic values.

PBR-Rooms is therefore designed as a controlled metallic-evaluation source.
It complements the public sources rather than replacing them: OpenRooms-FF and InteriorVerse remain the sources for the main geometry and albedo targets, OpenRooms-FF remains the official roughness source, and PBR-Rooms supplies reliable metallic supervision for the metallic target.
The released segmentation maps are auxiliary annotations for inspection and diagnostic variants, but segmentation is not treated as a primary benchmark target in this paper.

\subsubsection{Release Composition and Retained Benchmark Subset}
\label{app:pbr_rooms_release}

The full PBR-Rooms release contains 1,000 procedurally generated indoor scenes across five room categories: bathroom, bedroom, dining room, kitchen, and living room.
Each category contains 200 valid scenes and 7,800 rendered images, giving 39,000 RGB images at \(1280\times720\) resolution.
The release includes 24,000 viewpoint-variation images and 15,000 lighting-variation images.
Each RGB image is paired with pixel-aligned ground-truth channels: metric depth, surface normal, albedo, roughness, metallic, and object segmentation.

The paper uses a retained metallic evaluation subset rather than the full release.
Scenes are sampled independently within each room category with a fixed seed.
For the main-axis split, we retain 4\% of valid scenes per category, corresponding to 8 scenes per category and 40 scenes in total.
For the stress-axis split, we retain 5 additional non-overlapping scenes per category, corresponding to 25 scenes in total.
The resulting metallic evaluation subset contains 960 main-axis images and 450 stress-axis images.

\begin{table}[!htbp]
\centering
\small
\caption{PBR-Rooms release statistics and retained benchmark subset used in this paper.}
\label{tab:companion_dataset_stats}

\resizebox{0.9\linewidth}{!}{
\begin{tabular}{@{}l l@{}} 
\toprule
\textbf{Item} & \textbf{Value} \\
\midrule
\multicolumn{2}{@{}l}{\textit{\textbf{Full Release Statistics}}} \\
\addlinespace
Total Scenes & 1,000 indoor scenes (200 per category: Bathroom, bedroom, dining room, kitchen, living room) \\
Total Images & 39,000 rendered RGB images (1280×720) \\
Variation Types & Viewpoint-variation: 24,000; Lighting-variation: 15,000 \\
Released Channels & RGB, depth, surface normal, albedo, roughness, metallic, object segmentation \\
\midrule
\multicolumn{2}{@{}l}{\textit{\textbf{Benchmark Subset Used}}} \\
\addlinespace
Official Role & Metallic evaluation only \\
Main-axis Subset & 40 scenes, 960 images (40 scenes × 24 camera views) \\
Stress-axis Subset & 25 scenes, 450 images (25 scenes × 3 cameras × 6 lighting conditions) \\
License & CC BY-NC 4.0 \\
\bottomrule
\end{tabular}
} 

\vspace{0.5em}
\footnotesize
\textit{Note: The full PBR-Rooms release is multi-channel, but this paper uses it only as the official metallic source.}
\end{table}

\subsubsection{Procedural Scene Generation}

PBR-Rooms is generated with an Infinigen-based indoor scene generation and rendering pipeline~\cite{infinigen,infinigen_indoors}.
The pipeline procedurally creates indoor scene geometry, object layouts, material assignments, lighting configurations, and camera views.
Compared with manually curated asset collections, this procedural construction enables controlled variation over room category, camera viewpoint, and illumination while retaining pixel-aligned PBR ground truth.

The generated scenes cover five room categories: bathroom, bedroom, dining room, kitchen, and living room.
Each category contains 200 valid scenes.
The generation process is configured to increase the presence of visible metallic objects and metallic components while preserving physically and semantically plausible indoor layouts.
This makes the source useful for evaluating metallic prediction, where reliable and non-trivial metallic supervision is often limited in existing indoor inverse-rendering sources.

\subsubsection{Rendered Ground-Truth Channels}

For each rendered RGB image, PBR-Rooms exports pixel-aligned inverse-rendering ground-truth channels.
Depth is represented as metric depth in meters.
Surface normals are represented in camera space and are evaluated under the released normal convention.
Albedo is exported as intrinsic base color or diffuse reflectance.
Roughness and metallic are exported as PBR material parameters in \([0,1]\).
Object segmentation is released for inspection, quality control, and auxiliary diagnostics, but is not used as a primary benchmark target in this paper.

Official quantitative evaluation should use numeric ground-truth files, such as \texttt{.exr} or \texttt{.npy} files when available.
PNG files are used for RGB model input, visualization, and quick inspection.
The near-white depth channel, when present, is a visualization of depth and is not used as the official depth ground truth.
For metallic evaluation, predictions are interpreted as bounded continuous metallic maps rather than semantic metallic-object detections.

\begin{table}[!htbp]
\centering
\small
\caption{Released PBR-Rooms channels and their benchmark interpretation.}
\label{tab:pbr_rooms_channel_specs}
\setlength{\tabcolsep}{4.5pt}
\renewcommand{\arraystretch}{1.08}
\begin{tabularx}{\linewidth}{lX}
\toprule
Channel & Interpretation \\
\midrule
RGB & Rendered indoor image used as model input. \\
Depth & Metric depth in meters. \\
Surface normal & Camera-space surface normal under the released evaluation convention. \\
Albedo & Intrinsic base color / diffuse reflectance. \\
Roughness & PBR roughness parameter in \([0,1]\), where lower values denote smoother surfaces. \\
Metallic & PBR metallic parameter in \([0,1]\), where 0 denotes dielectric material and 1 denotes metallic material. \\
Object segmentation & Auxiliary annotation for inspection and diagnostics; not a primary benchmark target in this paper. \\
\bottomrule
\end{tabularx}
\end{table}

\subsubsection{Viewpoint and Lighting Variation}

PBR-Rooms includes both viewpoint variation and lighting variation.
For each retained main-axis scene, the benchmark uses 24 camera-view images rendered under the base lighting condition, yielding \(40\times24=960\) main-axis metallic evaluation images.
For stress-axis evaluation, the retained stress scenes are rendered under six lighting conditions: normal light, low light, HDR, dark-heavy, highlight-heavy, and mixed-temperature lighting.
The metallic stress-axis subset therefore contains \(25\) scenes, \(3\) cameras per scene, and \(6\) lighting conditions, yielding \(25\times3\times6=450\) stress-axis images.

The stress-axis lighting conditions are used for condition-oriented robustness analysis rather than for constructing a second leaderboard.
They allow the metallic evaluation to test whether continuous-map fidelity remains stable under challenging illumination conditions while keeping the scene source and evaluator fixed.

\subsubsection{Metallic Coverage Statistics}

PBR-Rooms is constructed to provide non-trivial metallic supervision while retaining the sparsity characteristic of indoor metallic materials.
Metallic values remain sparse overall, but visible metallic regions occur across multiple room categories.
Over evaluable pixels, the mean metallic value is 0.0229, the fraction of pixels with metallic value greater than 0.05 is 4.94\%, and the fraction greater than 0.50 is 0.83\%.
At the image level, 12.44\% of images contain more than 1\% high-metallic pixels under the \(\texttt{metallic}>0.5\) criterion, and 4.04\% contain more than 5\%.
At the scene level, 33.57\% of scenes contain at least one view with more than 1\% high-metallic pixels, and 15.58\% contain at least one view with more than 5\%.

\begin{table}[!htbp]
\centering
\small
\caption{Metallic coverage statistics for PBR-Rooms over evaluable pixels.}
\label{tab:pbr_rooms_metallic_statistics}
\setlength{\tabcolsep}{6pt}
\renewcommand{\arraystretch}{1.1}

\begin{tabular}{l c}
\toprule
\textbf{Statistic} & \textbf{Value} \\
\midrule
\multicolumn{2}{l}{\textit{\textbf{Pixel-Level Statistics}}} \\
Mean metallic value & 0.0229 \\
Pixel ratio (Metallic $> 0.05$) & 4.94\% \\
Pixel ratio (Metallic $> 0.10$) & 4.81\% \\
Pixel ratio (Metallic $> 0.50$) & 0.83\% \\
\midrule
\multicolumn{2}{l}{\textit{\textbf{Image-Level Coverage}}} \\
Images with coverage $> 0.1\%$ & 41.77\% \\
Images with coverage $> 1\%$   & 12.44\% \\
Images with coverage $> 5\%$   & 4.04\% \\
Images with coverage $> 10\%$  & 2.02\% \\
\midrule
\multicolumn{2}{l}{\textit{\textbf{Scene-Level Coverage}}} \\
Scenes (any view) with coverage $> 1\%$  & 33.57\% \\
Scenes (any view) with coverage $> 5\%$  & 15.58\% \\
\bottomrule
\end{tabular}

\tabremark{
Image-level coverage is computed as the fraction of pixels with metallic $> 0.5$.
}
\end{table}

\subsubsection{Quality Control}

We apply several quality-control checks before using PBR-Rooms for metallic evaluation.
First, invalid or failed renders are excluded from benchmark sampling.
Second, channel-completeness checks ensure that each retained RGB image has aligned metallic supervision and the required auxiliary metadata.
Third, the metallic channel is verified as a material-parameter export rather than a rendering-intensity, RGB-appearance, or highlight-derived signal.
Fourth, evaluable masks are used to exclude invalid geometry, background, and non-evaluable pixels from metric computation.
Finally, scene category, rendering condition, camera identifier, and channel paths are stored in released manifests so that every evaluation sample is auditable.

These checks are intended to ensure that metallic metrics reward agreement with the PBR material parameter rather than agreement with source-side artifacts, image brightness, specular highlights, or semantic object identity.

\subsubsection{Intended Use and Limitations}

PBR-Rooms is intended for controlled synthetic evaluation of inverse-rendering models and image-generation systems.
It should not be used as sole evidence for real-world deployment performance because it does not capture real camera noise, lens artifacts, motion blur, exposure changes, compression artifacts, or the full diversity of real indoor environments.
The source does not intentionally contain real people, real faces, real private homes, real addresses, biometric information, or personal user data.
Because the dataset intentionally provides reliable and non-trivial metallic supervision, its metallic distribution should not be interpreted as an estimate of real-world indoor metallic-material frequency.
Out-of-scope uses include person identification, surveillance, biometric inference, privacy-sensitive indoor reconstruction, real-estate profiling, autonomous driving, and outdoor scene understanding.

The companion source is used only for the official metallic target in this paper.
Its additional ground-truth channels are retained for quality control, inspection, and possible diagnostic analysis, but they are not mixed with the public-source evaluation of depth, normal, albedo, or roughness.

\subsection{InteriorVerse Material-Channel Reliability Screening}
\label{app:iv_material_reliability}

We exclude InteriorVerse roughness and metallic maps from official scoring because these released channels are not sufficiently reliable as benchmark-grade ground truth under our target definitions. This is a target-specific source-eligibility decision, not a dataset-wide rejection of InteriorVerse.

The decision is supported by two sources of evidence. First, prior work has reported reliability concerns for InteriorVerse material-related channels: RGB$\leftrightarrow$X notes that the roughness and metallicity values in InteriorVerse are often dubious and therefore does not use them for that dataset, while ePBR reports that metallicity and related mask/transmission channels in InteriorVerse are not reliable for their image-composition setting and require manual correction for rendering consistency~\cite{rgbx_arxiv,epbr}. Second, our source-screening examples show recurring failure patterns that are directly relevant to roughness and metallic scoring.

For roughness, the released maps can appear overly piecewise-constant, weakly differentiated across visibly distinct materials, coarsely aligned with object or material boundaries, or contaminated by appearance- and shading-like patterns. For metallic, the released channel can preserve visible RGB texture, screen or artwork content, specular highlights, emissive or over-exposed regions, and other appearance-dependent structures. These patterns are problematic for benchmark scoring because roughness and metallic are bounded material parameters, and official metrics would reward agreement with source-side artifacts rather than physically meaningful material recovery.

We therefore conservatively exclude InteriorVerse roughness and metallic from official scoring, use OpenRooms-FF for roughness evaluation, and use the companion source for metallic evaluation. This exclusion does not affect the use of InteriorVerse for depth, normal, and albedo, whose annotations are retained under the corresponding target definitions. Representative qualitative examples are provided in Appendix~\ref{app:iv_roughness_reliability_examples} and Appendix~\ref{app:iv_metallic_reliability_examples}.

\subsection{Source Reliability and Exclusion Rationale}
\label{app:source_reliability}

The benchmark uses target-dependent source selection rather than forcing all targets into a single source pool. 
OpenRooms-FF is retained for depth, normal, albedo, and roughness, but not metallic because metallic ground truth is unavailable. 
InteriorVerse is retained for depth, normal, and albedo, but its roughness and metallic channels are excluded from official scoring due to the qualitative reliability concerns described in Appendix~\ref{app:iv_material_reliability}. 
The companion dataset is therefore introduced only for metallic evaluation.

\section{Benchmark Split, Metadata, and Stress-Subset Construction}
\label{app:split_construction}

This appendix documents scene-level metadata construction, reviewer arbitration, dataset-specific stratification, retained-view rules, quota allocation, and programmatic stress-subset definitions. 
These procedures define how benchmark samples, audit metadata, and programmatic stress subsets are constructed; quantitative results are reported separately in Appendix~\ref{app:additional_results}.

\subsection{Scene-Level Metadata Schema}
\label{app:metadata_schema}

To reduce circularity between benchmark metadata and the evaluated multimodal systems, we distinguish between programmatic photometric fields and semantic annotation fields. 
Programmatic fields are used whenever stress subsets can be defined directly from image statistics, while semantic fields are used more conservatively for split construction, distribution auditing, and quality control. 
In particular, \texttt{scene\_category} and \texttt{illumination\_level} are used to balance and audit the benchmark split, rather than to define the main leaderboard axes.

Table~\ref{tab:metadata_schema} summarizes the scene-level metadata fields and their roles in the benchmark. 
Scene category and illumination level are used for sampling control and distribution auditing. 
Photometric challenge fields are used for condition-oriented stress subsets, while plausibility and coherence fields are used only for filtering, auditing, and post-hoc failure analysis.

\begin{table}[!htbp]
\centering
\caption{Scene-level metadata fields and benchmark roles.}
\label{tab:metadata_schema}
\small
\setlength{\tabcolsep}{5pt}
\renewcommand{\arraystretch}{1.25} 
\begin{tabular}{@{} l p{4.5cm} p{2.8cm} p{3.5cm} @{}}
\toprule
\textbf{Field group} & \textbf{Fields} & \textbf{Source} & \textbf{Benchmark role} \\
\midrule

\textbf{Split auditing} &
\texttt{scene\_category} \newline \texttt{illumination\_level} &
Base annotators / programmatic &
Sampling control and distribution auditing \\
\midrule

\textbf{Challenge subsets} &
\texttt{brightness\_level} \newline \texttt{dynamic\_range\_level} \newline \texttt{highlight\_strength} \newline \texttt{dark\_region\_ratio\_level} &
Programmatic &
Lighting stress tests \\
\midrule

\textbf{Quality control} &
\texttt{layout\_plausibility} \newline \texttt{functional\_coherence} \newline \texttt{object\_arrangement\_coherence} \newline \texttt{lighting\_naturalness} \newline \texttt{overall\_plausibility} \newline \texttt{confidence} \newline \texttt{issues} &
Base annotators / reviewer arbitration &
Filtering, auditing, and post-hoc failure analysis \\

\bottomrule
\end{tabular}
\end{table}

We do not use plausibility or coherence fields as official sampling axes because their distributions are highly skewed in both source datasets. Similarly, we do not use \texttt{lighting\_challenge\_type} as a sampling axis, since its raw counts are not stable scene-level unique labels in the current annotation pipeline. Instead, lighting stress subsets are defined from programmatic photometric fields, which are more reproducible and directly tied to image statistics.

\subsection{Metadata Annotation Pipeline}
\label{app:metadata_pipeline}

\subsubsection{Base Annotator Prompt and Model-Specific Instantiation}
\label{app:base_annotator_prompts}

The three base annotators---Doubao, GLM, and Qwen---use a shared scene-audit prompt with the same task definition, field schema, and decision rules. In all three scripts, the model is instructed to act as a strict, conservative, evidence-based indoor scene auditor and to produce a scene-level judgment from multiple representative views of the same indoor scene. The prompt explicitly states that the goal is not to judge whether an image looks photographic or rendered, but to assess whether the scene is reasonable, self-consistent, and usable in terms of spatial layout, functional organization, object arrangement, and lighting logic.

\paragraph{Input organization.}
Each base annotator receives a set of images belonging to the same scene as a single multimodal request. The user message first provides a short textual instruction asking the model to analyze the multi-view images of the same scene and return only valid JSON, and then appends the representative view images one by one as ordered visual inputs. The audit result is therefore produced at the scene level rather than at the single-image level.

\paragraph{Conservative shared decision policy.}
The prompt suppresses overly aggressive or realism-biased judgments. Rendered appearance, repeated textures, regular materials, unified furniture style, sparse clutter, or a generally clean/simple room should not be treated as default evidence of implausibility. Partially visible objects are allowed unless multi-view evidence indicates unreasonable position, support, scale, or use logic. The model instead focuses on structural failures such as object interpenetration, floating objects, implausible placement, orientation conflicts, blocked pathways, inconsistent functional zoning, or contradictory lighting and shadow evidence. When evidence is insufficient, the model outputs \texttt{uncertain} and lowers the confidence score.

\paragraph{Output schema and JSON constraint.}
All three base annotators output a strict JSON object with an identical schema. The schema includes \texttt{scene\_category}, plausibility and coherence fields, lighting-related fields, \texttt{overall\_plausibility}, \texttt{confidence}, \texttt{issues}, and \texttt{reason\_short}. The prompt forbids markdown, code fences, comments, or extra explanation outside the JSON object, and all categorical labels must be chosen from predefined enumerations.

\paragraph{Programmatic photometric fields.}
The prompt explicitly distinguishes programmatic photometric fields from semantic audit fields. Fields such as \texttt{brightness\_level}, \texttt{illumination\_level}, \texttt{dynamic\_range\_level}, \texttt{highlight\_strength}, and \texttt{dark\_region\_ratio\_level} are derived programmatically rather than being directly annotated by base annotators. Although base annotators may provisionally label these fields as \texttt{uncertain}, automated scripts subsequently compute and assign the final values based on image statistics.

\paragraph{Model-specific instantiation.}
The prompt body is shared across annotators, but the implementations differ at the API layer. The Doubao script uses an Ark-compatible endpoint, the GLM script uses a BigModel-compatible endpoint, and the Qwen script uses a DashScope-compatible endpoint. Apart from endpoint and model-name differences, the audit logic, multimodal message construction, JSON extraction strategy, and scene-level prompt design are aligned.

\subsubsection{Reviewer Arbitration}
\label{app:reviewer_arbitration}

After obtaining scene-level annotations from the three base annotators, we first construct an intermediate merged record. For semantic categorical fields, if at least two annotators agree on the same label, that label is taken as the provisional merged result by majority vote. If all three annotators output different labels for the same field, the field is treated as unresolved and is passed to a stronger reviewer VLM for final arbitration. Photometric fields are generated by programmatic image-statistics computation rather than by subjective VLM judgment.

The reviewer primarily arbitrates semantic and quality-related fields, including scene\_category, layout\_plausibility, functional\_coherence, object\_arrangement\_coherence, lighting\_naturalness, and overall\_plausibility, along with meta-fields like issues, reason\_short, and confidence. In contrast, programmatic photometric fields are treated as \textbf{strong priors}; they are preserved by default unless the reviewer identifies clear visual evidence that contradicts the computed labels.

\subsubsection{Field-Wise Fusion Rules}
\label{app:fusion_rules}

The final scene-level metadata are produced in two stages: intermediate merging over the outputs of the three base annotators, followed by reviewer arbitration for unresolved cases. Semantic categorical fields are resolved by majority vote when at least two annotators agree; otherwise they are passed to the reviewer. Free-text fields such as issues and reason\_short are retained as supporting evidence and exposed to the reviewer rather than directly concatenated. Programmatic photometric fields are derived from image statistics and used as the default final values.

After reviewer arbitration, the pipeline validates the reviewed JSON against the full target schema. Missing fields are filled with schema-consistent defaults, and the final annotation is stored together with auxiliary logging information such as disagreement fields, computed statistics, parse errors, and reviewer outputs.

\subsection{Dataset-Specific Strata}
\label{app:dataset_specific_strata}

We use dataset-native stratification rather than forcing all source datasets into a single global scene taxonomy. This design preserves the natural structure of each dataset during sampling, while allowing a lightweight harmonized reporting layer for cross-dataset summaries.

\paragraph{OpenRooms-FF.}
For OpenRooms-FF, we use the following native scene-category bins:
\[
\{\textit{bedroom}, \textit{living\_room}, \textit{office}, \textit{bathroom},
\textit{conference\_room}, \textit{dining\_room}, \textit{mixed\_open\_plan},
\textit{kitchen}, \textit{other}\}.
\]
The illumination labels are merged into three bins:
\[
\textit{dim}=\{\texttt{very\_low},\texttt{low}\},\quad
\textit{normal}=\{\texttt{medium}\},\quad
\textit{bright}=\{\texttt{high},\texttt{very\_high}\}.
\]
Each OpenRooms-FF sampling stratum is defined as a scene-category$\times$illumination cell.

\paragraph{InteriorVerse.}
For InteriorVerse, we use the following native scene-category bins:
\[
\{\textit{bedroom}, \textit{living\_room}, \textit{mixed\_open\_plan},
\textit{dining\_room}, \textit{kitchen}, \textit{office}, \textit{other}\}.
\]
Because InteriorVerse contains very few low-light scenes, its main benchmark strata use only two illumination bins:
\[
\textit{normal}=\{\texttt{medium}\},\quad
\textit{bright}=\{\texttt{high},\texttt{very\_high}\}.
\]
Dim InteriorVerse scenes are excluded from the main aggregation and retained only for diagnostic analysis when applicable.

\subsection{Source-Specific Retained-View Rules}
\label{app:retained_view_rules}

All official predictions are made image by image, without cross-view visual input. For OpenRooms-FF, each selected source pose forms a local \(3\times3\) forward-facing view group. Because the nine views are spatially close and highly redundant, the main evaluation uses only the central view of each local group. The remaining views are retained for supplementary diagnostics. For other benchmark sources, retained evaluation images follow the corresponding source-specific construction protocol.

\subsection{Quota Allocation Details}
\label{app:quota_allocation}

We use a minimum-allocation plus square-root reweighting rule for dataset-native scene-level stratified sampling. Let \(c_h\) denote the number of valid scene groups in stratum \(h\), \(N\) the total sampling budget for the dataset, and \(n_{\min}\) the minimum quota assigned to each non-empty stratum. After assigning the minimum quota to every non-empty stratum, the remaining budget \(R\) is distributed as
\[
n_h=\min\!\left(c_h,\; n_{\min}+\mathrm{round}\!\left(
R\cdot\frac{\sqrt{c_h}}{\sum_j \sqrt{c_j}}
\right)\right).
\]
Here, \(R\) is the remaining budget after minimum allocations. In our benchmark construction, we use \(n_{\min}=8\) for OpenRooms-FF and \(n_{\min}=6\) for InteriorVerse. The square-root term is used for diagnostic coverage rather than population estimation.

\subsection{Sampling Stability}
\label{app:sampling_stability}

All benchmark splits are generated with fixed random seeds and released as explicit manifest files. Official results are computed only from these released manifests, so repeated evaluation does not depend on stochastic resampling.

\subsection{Programmatic Photometric Stress Fields}
\label{app:photometric_fields}

This section provides the definitions of the programmatic photometric fields used for official stress slicing. Unlike semantic labels, these fields are derived from image statistics and used to construct reproducible challenge subsets for benchmark analysis.

\subsubsection{Challenge-Subset Definitions}
\label{app:challenge_subset_defs}

In addition to the main-axis target-wise summaries, we define four programmatic lighting stress subsets for condition-oriented robustness analysis. These subsets are used for diagnostic analysis and are reported separately from the main benchmark aggregation.
\paragraph{Low-light.}
Samples are included if $\texttt{illumination\_level} \in \{\texttt{very\_low}, \texttt{low}\}$. Alternatively, under the robustness definition, a sample is included if $\texttt{brightness\_level} = \texttt{low}$ and $\texttt{dark\_region\_ratio\_level} = \texttt{high}$.

\paragraph{HDR.}
Samples are selected where $\texttt{dynamic\_range\_level} = \texttt{high}$.

\paragraph{Highlight-heavy.}
Samples are selected where $\texttt{highlight\_strength} = \texttt{high}$.

\paragraph{Dark-region-dominant.}
Samples are selected where $\texttt{dark\_region\_ratio\_level} = \texttt{high}$.

\subsubsection{Image-Statistic Definitions and Thresholds}
\label{app:photometric_thresholds}

The photometric stress fields are computed programmatically from RGB image statistics rather than from subjective VLM judgments. 
For official stress-subset evaluation, each retained evaluation image is assigned stress labels from the exact RGB image used for prediction and scoring. 
Multi-view images used by the metadata annotation pipeline are not used to assign official photometric stress labels.

For shared stress slices that appear in both OpenRooms-FF and InteriorVerse, we first compute source-specific slice scores using retained evaluation images whose image-level stress label matches the slice.  

When both sources contain sufficient retained samples for the slice, we report the source-balanced macro-average of the source-specific slice scores. 

When a stress slice is insufficiently represented in one source, we report it as a source-specific stress diagnostic rather than image-pooling it with the other source. 

The \textit{low\_light} slice is therefore reported only for OpenRooms-FF because InteriorVerse provides insufficient low-light coverage under our retained split.

Before computing image statistics, each retained RGB image is converted to RGB and resized, if necessary, so that its maximum side length does not exceed 512 pixels. 
All pixel values are normalized to \([0,1]\).

For each pixel with sRGB values \((R_s,G_s,B_s)\), we compute an sRGB luma value
\[
Y_s = 0.2126 R_s + 0.7152 G_s + 0.0722 B_s .
\]
We also convert each sRGB channel to linear intensity:
\[
c_\ell =
\begin{cases}
c_s / 12.92, & c_s \le 0.04045,\\
\left(\frac{c_s+0.055}{1.055}\right)^{2.4}, & c_s > 0.04045,
\end{cases}
\]
and compute linear luminance
\[
Y_\ell = 0.2126 R_\ell + 0.7152 G_\ell + 0.0722 B_\ell .
\]
The mean sRGB luma and mean linear luminance are
\[
\mu_s=\frac{1}{|\Omega|}\sum_{p\in\Omega}Y_s(p),
\qquad
\mu_\ell=\frac{1}{|\Omega|}\sum_{p\in\Omega}Y_\ell(p),
\]
where \(\Omega\) denotes all pixels used for the photometric computation. 
We compute the 5th and 95th percentiles of linear luminance, denoted by \(P_5(Y_\ell)\) and \(P_{95}(Y_\ell)\), using a 1024-bin histogram. 
The dynamic range statistic is measured in exposure stops:
\[
D_{\mathrm{stop}}
=
\log_2
\frac{P_{95}(Y_\ell)+\epsilon}{P_5(Y_\ell)+\epsilon},
\qquad
\epsilon=10^{-6}.
\]
We define the highlight ratio and dark-region ratio as
\[
r_{\mathrm{hi}}
=
\frac{1}{|\Omega|}
\sum_{p\in\Omega}
\mathbf{1}\left[Y_\ell(p)\ge 0.85\right],
\]
\[
r_{\mathrm{dark}}
=
\frac{1}{|\Omega|}
\sum_{p\in\Omega}
\mathbf{1}\left[Y_\ell(p)\le 0.10\right].
\]

The continuous statistics are discretized using the fixed thresholds summarized in Table~\ref{tab:photometric_thresholds}. 
For brightness, the thresholds are derived from 18\% gray with a one-stop margin after conversion from linear intensity to sRGB luma, yielding approximately 0.332 and 0.634. 
For illumination, we use exposure stops relative to 18\% gray:
\[
E=\log_2 \frac{\mu_\ell+\epsilon}{0.18}.
\]
Dynamic range is bucketed from \(D_{\mathrm{stop}}\), while highlight and dark-region categories are bucketed from the pixel ratios \(r_{\mathrm{hi}}\) and \(r_{\mathrm{dark}}\), respectively.

\begin{table*}[!htbp]
\centering
\small
\caption{Programmatic photometric fields and discretization thresholds.}
\label{tab:photometric_thresholds}
\setlength{\tabcolsep}{6pt}
\renewcommand{\arraystretch}{1.15}
\begin{tabularx}{\textwidth}{@{}L{3.8cm} L{3cm} X@{}}
\toprule
Field & Statistic & Discretization rule \\
\midrule

\texttt{brightness\_level}
& mean sRGB luma \(\mu_s\)
& \(\texttt{low}: \mu_s<0.332\);
  \(\texttt{high}: \mu_s>0.634\);
  \(\texttt{medium}\): otherwise. \\

\addlinespace[0.2em]

\texttt{illumination\_level}
& exposure \(E\)
& \(\texttt{very\_low}: E\le-2\);
  \(\texttt{low}: -2<E\le-1\);
  \(\texttt{medium}: -1<E\le1\);
  \(\texttt{high}: 1<E\le2\);
  \(\texttt{very\_high}: E>2\). \\

\addlinespace[0.2em]

\texttt{dynamic\_range\_level}
& \(D_{\mathrm{stop}}\)
& \(\texttt{low}: D_{\mathrm{stop}}<2\);
  \(\texttt{medium}: 2\le D_{\mathrm{stop}}<4\);
  \(\texttt{high}: D_{\mathrm{stop}}\ge4\). \\

\addlinespace[0.2em]

\texttt{highlight\_strength}
& highlight ratio \(r_{\mathrm{hi}}\)
& \(\texttt{low}: r_{\mathrm{hi}}<0.01\);
  \(\texttt{medium}: 0.01\le r_{\mathrm{hi}}<0.05\);
  \(\texttt{high}: r_{\mathrm{hi}}\ge0.05\). \\

\addlinespace[0.2em]

\texttt{dark\_region\_ratio\_level}
& dark ratio \(r_{\mathrm{dark}}\)
& \(\texttt{low}: r_{\mathrm{dark}}<0.10\);
  \(\texttt{medium}: 0.10\le r_{\mathrm{dark}}<0.30\);
  \(\texttt{high}: r_{\mathrm{dark}}\ge0.30\). \\

\bottomrule
\end{tabularx}
\end{table*}

\begin{table}[!htbp]
\centering
\small
\caption{Stress-axis evaluation counts by target and source.}
\label{tab:stress_axis_counts}
\setlength{\tabcolsep}{3.6pt}
\renewcommand{\arraystretch}{1.12}

\begin{tabular}{lcccccccc}
\toprule
\multirow{2}{*}{Count}
& \multicolumn{2}{c}{Depth}
& \multicolumn{2}{c}{Normal}
& \multicolumn{2}{c}{Albedo}
& \multicolumn{1}{c}{Roughness}
& \multicolumn{1}{c}{Metallic} \\
\cmidrule(lr){2-3}
\cmidrule(lr){4-5}
\cmidrule(lr){6-7}
\cmidrule(lr){8-8}
\cmidrule(lr){9-9}
& OR-FF & IV
& OR-FF & IV
& OR-FF & IV
& OR-FF
& Comp. \\
\midrule
Scenes
& 180 & 105
& 180 & 105
& 180 & 105
& 180
& 25 \\
Images
& 709 & 1251
& 709 & 1251
& 709 & 1251
& 709
& 450 \\
\bottomrule
\end{tabular}

\tabremark{
OR-FF denotes OpenRooms-FF, IV denotes InteriorVerse, and Comp. denotes the companion metallic source.
Counts correspond to retained scenes and images used for stress-axis diagnostics.
InteriorVerse roughness and metallic are excluded from stress-axis scoring for the same source-reliability reasons as in the main evaluation.
}
\end{table}

\FloatBarrier

\section{Evaluation Metrics and Failure Handling}
\label{app:extended_eval}

This appendix specifies target-specific output-validity checks, post-processing rules, failure handling, metrics, masking rules, normalization conventions, and source-balanced aggregation. 
It defines how a generated prediction is converted into benchmark scores after the access protocol has produced an output.

\subsection{Common Evaluator Rules}
\label{app:common_eval_rules}

All official predictions are matched to the released benchmark manifests and evaluated image by image.
Predictions are resized to the corresponding ground-truth resolution when necessary, and all metrics are computed only over target-valid pixels.
Invalid, unreadable, missing, or unmatched prediction files are handled according to the target-specific failure rules and are not silently replaced.
For targets evaluated on both OpenRooms-FF and InteriorVerse, we first compute source-specific scores and then report source-balanced macro-averages.
For targets with a single official retained source, results are reported on that source-specific main-axis or stress subset.

\subsection{Output-Validity and Post-processing Summary}
\label{app:output_validity_checks}

Before metric computation, each prediction is checked for target-specific output validity and then converted into the representation expected by the corresponding evaluator.
These checks determine whether a generated output instantiates the required scoreable target map, while post-processing defines the deterministic conversions applied before scoring.
They are separate from metric scoring: an output must first be readable, matched to a retained benchmark sample, spatially aligned or resizable to the ground-truth resolution, and convertible to the target representation.
Outputs that fail these checks are documented according to the failure-handling rules in Appendix~\ref{app:failure_handling} rather than silently assigned artificial metric values.

\begin{table*}[!htbp]
\centering
\footnotesize
\caption{Target-specific output-validity checks and deterministic post-processing before metric computation.}
\label{tab:output_validity_checks}
\setlength{\tabcolsep}{4pt}
\renewcommand{\arraystretch}{1.12}
\begin{tabularx}{\textwidth}{@{}L{1.45cm} X X@{}}
\toprule
Target & Required scoreable output representation & Deterministic post-processing before scoring \\
\midrule

Depth
& Readable output convertible to a single-channel depth-like map spatially aligned with the query image.
The output must encode a dense relative-depth-like field rather than a semantic map, text overlay, stylized rendering, or unrelated grayscale image.
& Convert to a single-channel continuous map and resize to the ground-truth resolution.
Metrics are computed over valid finite ground-truth depth pixels.
Prediction normalization, polarity selection, and per-image affine fitting are applied only at scoring time; these operations do not make a non-depth output valid. \\

\midrule

Normal
& Dense RGB-encoded normal-map representation aligned with the query image.
Stylized renderings, normal-map-like colorizations, or irregular RGB patterns that do not represent coherent surface orientations are treated as non-scoreable representation failures.
& Resize to the ground-truth resolution.
Decode RGB values as \(\mathbf{n}=2\mathbf{rgb}-1\), normalize to unit vectors, and remove invalid pixels with zero-length, near-zero-length, NaN, or Inf normals before angular-error computation. \\

\midrule

Albedo
& Spatially aligned RGB reflectance-like map for the query image.
The output should preserve scene layout and visible material regions rather than produce a relit image, stylized rendering, segmentation map, or scene reconstruction.
& Resize to the ground-truth resolution and convert to RGB values in \([0,1]\).
Metrics are computed within the valid albedo region.
For structural or perceptual diagnostics, invalid pixels are handled by the target evaluator as described in Appendix~\ref{app:albedo_eval}. \\

\midrule

Roughness
& Spatially aligned single-channel bounded roughness map.
Values are interpreted as material roughness in \([0,1]\), not as RGB brightness, shading, shadows, edges, or a grayscale rendering of the input image.
& Resize to the ground-truth resolution, convert to a single-channel map, and clip or normalize to \([0,1]\) according to the roughness evaluator.
Metrics are computed within the released material-valid region. \\

\midrule

Metallic
& Spatially aligned single-channel bounded continuous metallic map.
Values are interpreted as metallic material parameters in \([0,1]\), not as edge strength, reflectance intensity, highlights, or a grayscale rendering.
& Resize to the ground-truth resolution, convert to a single-channel continuous map, and clip to \([0,1]\).
No per-image min--max normalization is applied.
Metrics are computed within \texttt{MetallicEvalMask}. \\

\midrule

All targets
& The prediction must be readable, matched to a retained benchmark image, and associated with the correct target protocol.
& Predictions are evaluated image by image, resized when needed, and scored only within target-specific valid regions or masks.
Missing, unreadable, unmatched, or non-scoreable outputs are logged under the failure-handling rules rather than replaced silently. \\

\bottomrule
\end{tabularx}

\tabremark{
This table is a compact entry point for output-validity and post-processing rules.
Target-specific metric definitions, masks, normalization conventions, and failure handling are provided in the following subsections.
}
\end{table*}
\subsection{Depth Affine-Invariant Evaluation}
\label{app:depth_alignment_diagnostics}

Depth is evaluated primarily as affine-invariant relative-geometry recovery rather than calibrated metric-depth prediction.
This protocol is used because image editors and some dense-depth systems may output relative-depth-like maps with different scale, shift, or polarity conventions.

For each prediction, we first convert the output to a single-channel continuous depth-like map and resize it to the ground-truth resolution.
Metrics are computed within valid ground-truth pixels.
The prediction is normalized over valid pixels, and its polarity is selected by comparing Spearman rank correlation with the ground-truth depth under the two possible polarity conventions.
After polarity selection, scalar depth errors are computed after least-squares affine alignment to the metric ground truth over valid pixels.
The aligned prediction is
\[
\hat d_{\mathrm{AI}} = a\hat d + b,
\]
where \(a\) and \(b\) are fitted independently for each evaluated image over valid pixels.

The primary depth metric is AbsRel-AI.
We additionally report RMSE-AI, MAE-AI, \(\delta_1\)-AI, \(\delta_2\)-AI, and Boundary F1.
Boundary F1 is an inverse-depth boundary diagnostic that evaluates local depth-discontinuity recovery rather than global scalar alignment.
Because affine alignment and polarity selection use ground truth only at scoring time, these numbers should not be interpreted as calibrated metric-depth performance.
They measure whether the output preserves relative scene geometry after convention-level scale, shift, and polarity ambiguities are removed.
This choice is intentionally favorable to systems that output only visually plausible or relative-depth-like maps: it removes global convention errors before scoring and therefore tests whether the remaining spatial ordering and depth structure are physically meaningful.

We additionally report ordinal-agreement and boundary diagnostics to make the role of affine alignment explicit.
Spearman's \(\rho\) and Kendall's \(\tau\) measure relative depth-order agreement before affine fitting, while Boundary F1 evaluates local depth-discontinuity recovery after affine alignment.
These diagnostics are not used for official ranking, but help separate global scalar alignment from ordinal and boundary-structure fidelity.

\begin{table}[!htbp]
  \centering
  \compactappendixtable
  \caption{Depth access-setting diagnostics on a 96-image subset. Spearman's $\rho$ and Kendall's $\tau$ are computed before affine fitting, while AbsRel-AI/RMSE-AI/MAE-AI are computed after polarity selection and affine alignment.}
  \label{tab:depth_protocol_diagnostics_subset}

  \begin{tabular}{@{} >{\raggedright\arraybackslash}p{4.9cm} *{6}{c} @{}} 
    \toprule
    Setting 
    & AbsRel-AI$\downarrow$ 
    & RMSE-AI$\downarrow$ 
    & MAE-AI$\downarrow$ 
    & Spearman $\rho\uparrow$ 
    & Kendall $\tau\uparrow$ 
    & B-F1$\uparrow$ \\
    \midrule
    A0: RGB + prompt
    & 0.187 & 0.498 & 0.384 & 0.792 & 0.647 & \textbf{0.142} \\
    A1: RGB + prompt + fixed RGB--depth exemplar
    & \textbf{0.141} & \textbf{0.432} & \textbf{0.327} & \textbf{0.839} & \textbf{0.700} & 0.082 \\
    A2: RGB + prompt + segmentation reference
    & 0.178 & 0.499 & 0.384 & 0.771 & 0.620 & 0.125 \\
    \bottomrule
  \end{tabular}

  \tabremark{
    AI denotes affine-invariant scoring after prediction min--max normalization, scoring-time polarity selection, and least-squares affine fitting.
    Spearman's $\rho$ and Kendall's $\tau$ are computed on the selected-polarity relative-depth prediction before affine fitting.
    B-F1 denotes the DepthPro-style inverse-depth Boundary F1 diagnostic.
    A0 is the controlled RGB-to-depth prompt setting.
    A1 adds one fixed RGB--depth exemplar pair, and A2 provides a segmentation map as a visual reference.
    These variants change the access setting and are reported as protocol diagnostics rather than as official ranking results.
    Boldface marks within-diagnostic numerical optima only and does not indicate official benchmark ranking.
  }
\end{table}

\subsection{Normal Evaluation Overview}
\label{app:normal_eval_overview}

Normal prediction is evaluated with angular-error metrics over valid pixels.
The primary metric is Acc@22.5, the fraction of valid pixels whose angular error is at most \(22.5^\circ\).
We additionally report Mean Angular error, Median Angular error, Acc@11.25, and Acc@30.
For source-balanced normal results, OpenRooms-FF and InteriorVerse are scored separately and then macro-averaged across sources.
For proprietary generative systems under the normal protocol, scores are first computed for each fixed exemplar-conditioned run; the reported Avg row is the arithmetic mean over the three pre-specified exemplar runs.

\subsection{Albedo Evaluation Overview}
\label{app:albedo_eval_overview}

Albedo is evaluated as an intrinsic-appearance map under the fixed illumination-aware structure-preserving access setting.
MAE is the primary scalar distortion metric.
PSNR, SSIM, and LPIPS are reported as complementary diagnostics for structural and perceptual consistency.
All metrics are computed over valid evaluation regions after resizing predictions to the ground-truth resolution.
For source-balanced albedo results, OpenRooms-FF and InteriorVerse are scored separately and then macro-averaged across sources.

\subsection{Target-Specific Failure Handling}
\label{app:failure_handling}

Depth metrics are computed only on finite and positive ground-truth depth pixels. Normal maps are decoded into unit vectors, and pixels with zero-length, near-zero-length, NaN, or Inf normals are excluded before angular-error computation. Albedo, roughness, and metallic predictions are clipped to their valid target ranges before scoring. Material-map metrics are computed only on released material-valid regions.

\subsection{Metric Summary}
\label{app:metric_summary}

Table~\ref{tab:metric_summary} summarizes the primary metric and diagnostic metrics used for each target. 
The benchmark uses target-specific metrics because depth, normal, albedo, roughness, and metallic represent different physical quantities and have different validity and aggregation requirements.

\begin{table}[!htbp]
\centering
\small
\caption{Summary of primary and diagnostic metrics for each target.}
\label{tab:metric_summary}
\resizebox{0.9\linewidth}{!}{
\begin{tabular}{l l l}
\toprule
\textbf{Target} & \textbf{Primary Metric} & \textbf{Diagnostic Metrics} \\
\midrule
Depth 
& AbsRel-AI 
& RMSE-AI; MAE-AI; $\delta_1$-AI; $\delta_2$-AI; Spearman $\rho$; Kendall $\tau$; Boundary F1 \\
\addlinespace 
Normal 
& Acc@22.5 
& Mean angular error; Median angular error; Acc@11.25; Acc@30 \\
\addlinespace
Albedo 
& MAE 
& PSNR; SSIM; LPIPS \\
\addlinespace
Roughness 
& RMSE 
& MAE; PSNR; SSIM; auxiliary LPIPS in appendix diagnostics \\
\addlinespace
Metallic 
& MAE 
& PSNR; SSIM; LPIPS \\
\bottomrule
\end{tabular}
}

\vspace{0.5em}
\begin{minipage}{0.9\linewidth}
\footnotesize
\textit{Note:} AI denotes affine-invariant scoring for depth. For normal, Acc@22.5 is the fraction of valid pixels whose angular error is at most $22.5^\circ$. For roughness, RMSE is the primary metric as larger deviations are disruptive for material appearance.
\end{minipage}
\end{table}

We use affine-invariant AbsRel (AbsRel-AI) as the primary metric for depth, Acc@22.5 as the primary directional-accuracy metric for normal, RMSE as the primary scalar error metric for roughness, and MAE as the primary scalar distortion metric for albedo and metallic.
PSNR, SSIM, and LPIPS are reported as diagnostic metrics where appropriate to characterize distortion, structural consistency, and perceptual map-space similarity.
No thresholded metallic-region metric is used for official metallic ranking because the official metallic protocol evaluates metallic as a continuous bounded material map rather than as a calibrated binary detection task.

\subsection{Normal Evaluation}
\label{app:normal_eval}

Predicted normal maps are resized to the ground-truth spatial resolution. RGB normal maps are decoded as \(\mathbf{n}=2\mathbf{rgb}-1\) and then normalized to unit vectors. Invalid pixels are removed using the intersection of the dataset-provided valid mask and legal-normal masks for both prediction and ground truth. Pixels with near-zero vector norm or non-finite values are excluded.

Angular error is defined as
\[
\theta = \arccos(\mathrm{clip}(\hat{\mathbf n}\cdot \mathbf n, -1, 1)) \cdot \frac{180}{\pi}.
\]
Acc@22.5 is reported as the ratio of valid pixels with \(\theta < 22.5^\circ\). 
All Acc@ thresholds are reported as ratios in \([0,1]\). We additionally report mean angular error, median angular error, angular RMSE, Acc@11.25, and Acc@30. The evaluator writes per-image results to \texttt{normal\_metrics.csv} and aggregate summaries to \texttt{normal\_summary.json}.

\subsection{Albedo Evaluation}
\label{app:albedo_eval}

For albedo evaluation, ground-truth albedo maps, predicted albedo maps, and valid masks are matched by relative path and filename stem. Dataset-specific suffixes are stripped before matching. Predicted maps are resized to the ground-truth spatial resolution before scoring.

Images are converted to RGB and normalized to \([0,1]\). Unless otherwise specified, evaluation is performed in normalized RGB space rather than linearized RGB. Valid masks are thresholded and used to restrict metric computation to valid pixels. We report masked MAE as the main scalar distortion axis for albedo, computed as the mean absolute RGB error over valid pixels. Because albedo evaluation targets structure-preserving reflectance recovery rather than scalar distortion minimization alone, MAE is interpreted together with SSIM and LPIPS. PSNR is computed from masked MSE. SSIM is computed on the valid-mask bounding box after filling invalid prediction pixels with ground-truth values. LPIPS is computed on the same masked crop when enabled.

\subsection{Roughness Evaluation}
\label{app:roughness_eval}

Roughness is evaluated on OpenRooms-FF under the controlled RGB-to-map track.
Ground-truth roughness maps, predicted roughness maps, and optional material-valid masks are matched by relative path and filename stem after dataset-specific suffix stripping.
Predicted maps are resized to the ground-truth spatial resolution before scoring.
All roughness maps are converted to single-channel grayscale and clipped or normalized to the released target range \([0,1]\).

Evaluation is restricted to the released material-valid region.
When a valid mask is provided, it is resized to the ground-truth resolution if necessary, thresholded, and used to restrict all metric computation to valid pixels.
When no valid mask is provided, all pixels are treated as valid.
Invalid, non-finite, or unmatched prediction files are excluded according to the evaluator's failure-handling rules.

RMSE is the primary roughness scalar error metric:
\[
\mathrm{RMSE}
=
\sqrt{
\frac{1}{|\Omega|}
\sum_{p\in\Omega}
(\hat r_p-r_p)^2
},
\]
where \(r_p\) and \(\hat r_p\) denote the ground-truth and predicted roughness values at valid pixel \(p\), and \(\Omega\) is the valid evaluation region.

MAE is reported as an auxiliary scalar diagnostic:
\[
\mathrm{MAE}
=
\frac{1}{|\Omega|}
\sum_{p\in\Omega}
|\hat r_p-r_p|.
\]
PSNR is reported as a supplementary distortion metric:
\[
\mathrm{PSNR}
=
10\log_{10}\frac{1}{\mathrm{MSE}}.
\]
SSIM is a required structural diagnostic for material-region consistency.
LPIPS is reported only as an auxiliary perceptual indicator because roughness is a single-channel material map; the single-channel roughness map is replicated to three channels before LPIPS computation, and LPIPS is not used for official ranking.

For SSIM and LPIPS, we compute the valid-mask bounding box and fill invalid prediction pixels inside the crop with the corresponding ground-truth values before scoring.
This prevents invalid regions outside the material-valid mask from affecting structural or perceptual diagnostics.
Because roughness is a bounded material map, RMSE and MAE are interpreted together with SSIM rather than as a standalone complete quality measure.

\subsection{Metallic Evaluation}
\label{app:metallic_evaluation}

For metallic evaluation, predictions are resized to the ground-truth spatial resolution, converted to floating-point maps, clipped to \([0,1]\), and evaluated within \texttt{MetallicEvalMask}.
The mask is derived from valid rendered geometry and material-valid regions, so that background and invalid pixels are excluded from metric computation.
No per-image min--max normalization is applied, because metallic is a bounded physical material parameter and rescaling each prediction would change the meaning of the predicted values.

Following dense material-map evaluation practice, the official metallic comparison treats metallic as a continuous bounded map rather than as a thresholded binary detection task.
MAE is used as the primary scalar distortion metric.
PSNR, SSIM, and LPIPS are reported as complementary full-map diagnostics for distortion, structural consistency, and perceptual map-space similarity.

Because metallic regions are sparse, continuous full-map metrics can favor conservative near-background predictions.
We therefore interpret metallic results as continuous-map fidelity under the released evaluation mask, not as calibrated binary metallic-region detection.
No thresholded metallic-region metric is used for official ranking.
Thresholded metallic-region scores are not reported because they introduce threshold-calibration choices that are not part of the official continuous-map evaluation protocol.

\subsection{Source-Balanced Aggregation}
\label{app:source_balanced_aggregation}

Sampling is performed in the native label space of each source dataset. We do not force OpenRooms-FF and InteriorVerse into a single global scene taxonomy during sampling, because their scene-category distributions and label granularities differ substantially. This avoids introducing unnecessary mapping noise into benchmark construction.

For cross-dataset summaries, we use a lightweight harmonized reporting layer. Harmonization is applied only at reporting time and only for shared high-level categories, such as \textit{bedroom}, \textit{living\_room}, \textit{mixed\_open\_plan}, \textit{dining\_room}, \textit{kitchen}, \textit{office}, and \textit{other}. Dataset-native strata remain the basis for sampling and official within-dataset leaderboards. This design follows the principle of native sampling with harmonized reporting.

\section{Access Settings and Generation Pipelines}
\label{app:elicitation_protocols}

This appendix documents the official access settings, target-specific elicitation protocols, generation pipelines, diagnostic access variants, and run-level logging used for proprietary generative systems. 
The access-setting definitions specify what evidence is provided to a system before prediction; the evaluator in Appendix~\ref{app:extended_eval} specifies how the resulting prediction is scored.

\subsection{Official Access Settings}
\label{app:official_access_settings}

All official proprietary-editor runs follow the same single-image evidence rule. The query RGB image is the only scene evidence provided to the model. Additional information, when used, specifies the requested target definition, map convention, or allowed query-image-derived cue. No official setting provides ground-truth maps, benchmark labels, test-set feedback, multi-view scene evidence, or model-output feedback.

\begin{table}[!htbp]
\centering
\small
\caption{Official access settings used in the main comparison.}
\label{tab:appendix_access_settings}

\resizebox{0.9\linewidth}{!}{
\begin{tabular}{p{2.2cm} p{4.2cm} p{9cm}} 
\toprule
\textbf{Target} & \textbf{Official setting} & \textbf{Official auxiliary cue} \\
\midrule

Depth
& Controlled RGB-to-relative-depth
& Query RGB image and a fixed relative-depth prompt specifying the near-white convention (white=near, black=far). No masks, analysis text, paired examples, or cross-view inputs are used in the official main comparison. Affine alignment and polarity selection are scoring-time operations, not model inputs. \\
\midrule

Normal
& Convention-stabilized elicitation
& Pre-specified three RGB--normal exemplar pairs for communicating the normal-map convention. \\
\midrule

Albedo
& Illumination-aware structure-preserving elicitation
& Query-image-only illumination-confound analysis derived from the same RGB image. No masks, GT maps, benchmark labels, external references, or multi-view fields. \\
\midrule

Roughness
& Controlled RGB-to-map
& None beyond the query RGB image and the fixed roughness prompt. \\
\midrule

Metallic
& Controlled RGB-to-map
& None beyond the query RGB image and the fixed metallic prompt. \\

\bottomrule
\end{tabular}
}

\vspace{0.5em}

\begin{minipage}{0.9\linewidth}
\raggedright
\scriptsize
All official settings include the query RGB image and a target-specific fixed task definition, and share the same single-image evidence constraint, benchmark split, evaluator, legality checks, failure-handling rules, source-balanced aggregation rule, and stress-subset reporting protocol. Target-specific auxiliary cues are allowed only when listed in the table. For depth, roughness, and metallic, no auxiliary cue beyond the query RGB image and fixed target-definition prompt is used in the main comparison. For depth, affine alignment and polarity selection are applied only by the evaluator after prediction and are not model inputs.
\end{minipage}

\end{table}

\subsection{Prompt Templates and Target Instructions}
\label{app:prompt_templates}

The released code package contains the exact prompt-construction logic used for the main proprietary image-editor experiments.
Here we summarize the target-level instruction constraints rather than reproducing the full prompts verbatim.
The prompts are fixed before scoring and are used to define the requested output representation, not to provide ground-truth maps, benchmark labels, metric feedback, or multi-view evidence.

The exact prompts are implemented inside the released generation scripts rather than reconstructed from this summary.
For example, the depth prompt fixes the near-white relative-depth convention; the albedo scripts construct a prompt from a fixed intrinsic-albedo instruction plus cached query-image-derived lighting notes; the normal scripts use fixed RGB--normal exemplar pairs to communicate the normal-map convention; and the roughness and metallic scripts define controlled RGB-to-map material-map prompts.

\begin{table*}[t]
\centering
\footnotesize
\caption{Compact summary of target-level prompt requirements used in the main image-editor experiments.}
\label{tab:prompt_template_summary}
\setlength{\tabcolsep}{4.5pt}
\renewcommand{\arraystretch}{1.12}
\begin{tabularx}{\textwidth}{@{}L{1.7cm} L{2.7cm} X@{}}
\toprule
Target & Official editor input & Compact target instruction summary \\
\midrule

Depth
& Query RGB + fixed prompt
& Generate a dense single-image relative-depth map aligned with the query RGB image.
The output is a single-channel grayscale depth-buffer-like map.
The official prompt uses a fixed near-white convention, where white denotes surfaces closer to the camera and black denotes farther surfaces.
The prompt emphasizes perspective, occlusion ordering, contact/support relations, room layout, object placement, and relative size as geometric cues.
It forbids cropping, shifting, recomposition, added content, text, labels, overlays, color, stylization, and visually pleasing grayscale rendering unrelated to depth. \\

\midrule

Normal
& Query RGB + fixed RGB--normal exemplar pair + fixed prompt
& Generate a dense view-space surface normal map for the query RGB image.
The exemplar normal map is used only to communicate the RGB normal-map convention, including palette, direction-family mapping, saturation range, boundary style, and large-region consistency.
The prompt requires RGB values to encode surface orientation only, not material color, lighting, texture, shading, highlights, reflections, or appearance patterns.
It asks the model to preserve object boundaries, broad geometry, major silhouettes, planar regions, large-scale curvature, folds, panel recesses, and real shape discontinuities, while suppressing texture-like detail and illumination residue.
The output is only the final normal map for the query image. \\

\midrule

Albedo
& Query RGB + cached query-image-derived analysis + fixed prompt
& Convert the query RGB image into a clean intrinsic albedo map.
The prompt requires preserving scene layout, visible objects, object boundaries, material regions, intrinsic surface color, and material-region color boundaries.
It removes illumination-dependent appearance, including cast shadows, attached shading, highlights, reflections, interreflections, exposure variation, ambient occlusion, and illumination color cast.
The cached analysis provides query-image-derived lighting and albedo hints, such as likely highlight, shadow, or exposure-confounded regions.
The prompt forbids object insertion, deletion, stylization, redesign, scene completion, semantic reinterpretation, topology change, spatial rearrangement, or hallucinated local structure. \\

\midrule

Roughness
& Query RGB + fixed prompt
& Generate a spatially aligned single-channel grayscale roughness map.
Black denotes very smooth or polished surfaces, while white denotes rough and diffuse surfaces.
The prompt asks the model to infer roughness from material appearance, highlight sharpness, reflection behavior, coating cues, and visible surface micro-structure, rather than RGB brightness alone.
It forbids copying direct illumination, cast shadows, self-shadowing, ambient occlusion, reflections, bright highlights, low-frequency lighting gradients, albedo patterns, printed textures, or semantic identity into the roughness map.
The prompt encourages spatially stable material maps while allowing local variation only when supported by visible material evidence. \\

\midrule

Metallic
& Query RGB + fixed prompt
& Generate a spatially aligned single-channel grayscale metallic map.
Black denotes dielectric or non-metal surfaces, and white denotes clearly exposed metallic material.
The prompt treats the target as a sparse material map: most indoor pixels should remain black unless visible evidence clearly indicates exposed metal.
It forbids relit grayscale renderings, edge maps, reflectance sketches, contour tracing, gray outlines around non-metal objects, or using low gray values to preserve scene structure.
It also forbids inferring metal from brightness, smoothness, reflections, shadows, specular highlights, mirror-like appearance, glass, ceramic, wood, plastic, painted surfaces, or glossy non-metal materials alone.
When evidence is ambiguous, the prompt instructs the model to choose conservative non-metal predictions. \\

\bottomrule
\end{tabularx}

\tabremark{
This table summarizes the target-level prompt constraints used in the main experiments.
The exact prompt strings and provider-specific request wrappers are released in the experiment scripts.
Diagnostic variants such as RGB--target exemplars for depth or roughness, segmentation-assisted roughness, and metallic text-prior or exemplar variants are documented separately and are not mixed into the official main comparison unless explicitly stated.
}
\end{table*}

\subsection{Controlled RGB-to-Map Protocol}
\label{app:controlled_rgb_to_map_protocol}
\label{app:controlled_protocol}

In the controlled track, each system receives only the query RGB image as visual input. No paired exemplars, segmentation masks, region-filling priors, external reference images, cross-view inputs, or auxiliary conditioning derived from other images are allowed. Proprietary multimodal or generative systems use fixed minimal prompts, while specialized inverse-rendering predictors are evaluated in their standard single-image inference mode.

\subsection{Albedo Elicitation Protocol}
\label{app:albedo_protocol}

For albedo, the disclosed target-specific elicitation track uses a structure-preserving single-image transformation setting. All allowed conditioning must be derived exclusively from the query RGB image. No segmentation masks, ground-truth annotations, benchmark labels, multi-view inputs, or auxiliary information derived from other images are used. Query-image-only auxiliary analysis, when used, is treated as a cached representation of single-image conditioning and not as external information. The model is required to remove illumination-dependent appearance while preserving intrinsic material textures.

\subsection{Normal Exemplar-Stabilized Protocol}
\label{app:normal_protocol}

Normal-map prediction requires a stable RGB encoding convention: colors must represent surface orientation rather than material appearance, shading, or semantic regions. In pilot runs, prompt-only instructions were often insufficient to communicate this convention consistently across proprietary image editors. The official normal protocol therefore uses fixed RGB--normal exemplar pairs as convention cues.

For normal estimation, the disclosed target-specific elicitation track uses a fixed set of three shared RGB--normal exemplar pairs as a convention-stabilization mechanism. Each query RGB image is processed independently under each exemplar pair. The three exemplar pairs are fixed before benchmark evaluation and are shared across all benchmark images and all proprietary systems.
They communicate the intended RGB-to-normal mapping convention, including color encoding, orientation-family mapping, saturation range, and image-wide consistency, and are not treated as query-specific supervision.

Using three fixed exemplar pairs reduces dependence on a single scene layout, orientation distribution, boundary style, or exemplar-specific normal-map appearance prior while keeping the access setting fixed before evaluation.
Because exemplars may still affect output smoothness, boundary style, saturation behavior, or other dense-map appearance priors, results obtained under this setting are interpreted as protocol-conditioned proprietary-system results rather than pure model-intrinsic normal-prediction scores.

\subsection{Albedo Generation Pipeline}
\label{app:albedo_generation_pipeline}

This subsection documents the proprietary-system implementation used in the disclosed structure-preserving albedo elicitation protocol. We evaluate albedo generation with GPT-Image-1.5, WAN2.7-Image, and Doubao Seedream 5.0. In all cases, generation is performed on benchmark-retained target images independently, and final scoring is image-wise under the albedo evaluation protocol.

\paragraph{Common task definition.}
All systems preserve scene layout, visible objects, object boundaries, and intrinsic material texture, while removing illumination-dependent appearance such as cast shadows, attached shading, highlights, reflections, interreflections, exposure variation, ambient occlusion, and illumination color cast. Prompts discourage object insertion, deletion, stylization, geometry changes, or content completion.

\paragraph{Shared auxiliary-analysis generation.}
For the disclosed albedo protocol, we generate a single auxiliary-analysis record for each retained query RGB image using Doubao Seed 2.0 Pro.
The record is generated from the same query image only, cached as a per-image JSON file aligned to the query image path, and reused unchanged across all proprietary albedo systems.
No ground-truth albedo map, target label, evaluation metric, model output, multi-view observation, or scene-level metadata field is used to generate or revise this record.

The JSON record may contain fields such as scene\_summary, albedo\_material\_notes, albedo\_lighting\_notes, and albedo\_prompt\_suffix.
These fields are used to construct the final albedo-generation prompt under the fixed structure-preserving task definition.
View-dependent or multi-view fields, when present in auxiliary code paths, are not used in the reported albedo runs.

\paragraph{System-specific implementation.}
GPT-Image-1.5, WAN2.7-Image, and Doubao Seedream 5.0 are all evaluated with the same cached query-image-only auxiliary-analysis records.
The systems differ only in their generation backends and request formats, while the auxiliary analysis supplied for a given query image is identical across systems.
GPT-Image-1.5 uses an image-generation/editing endpoint and records client configuration and output paths in the manifest.
WAN2.7-Image uses the released query RGB image as the edit input and records request metadata in the manifest.
Doubao Seedream 5.0 uses independent-image mode and processes each query image separately with the same shared albedo prompt construction.
Optional multi-view analysis code paths are not used in the reported albedo results.

\paragraph{Evaluation handling.}
All generated albedo maps are resized to the benchmark target resolution and evaluated with the official albedo protocol described in Appendix~\ref{app:albedo_eval}. We report masked MAE as the main scalar distortion axis, together with PSNR, SSIM, and LPIPS. For albedo, SSIM and LPIPS are treated as required diagnostics rather than optional secondary indicators, because the benchmark targets structure-preserving reflectance recovery rather than scalar distortion minimization alone.

\subsection{Normal Generation Pipeline}
\label{app:normal_generation_pipeline}

In the disclosed multi-exemplar convention-stabilized normal elicitation protocol, proprietary systems are evaluated under a fixed set of three paired-exemplar settings. 
Each query RGB image is processed independently under each exemplar pair. 
Each exemplar pair consists of one indoor RGB image and its corresponding normal map. 
The three exemplar pairs are fixed before benchmark evaluation and are shared across all proprietary systems and all benchmark images. 
Pilot experiments showed that prompt-only instructions were often insufficient to communicate the intended normal-map convention and RGB-to-normal transformation behavior.

\paragraph{Evaluated proprietary systems.}
We evaluate normal-map generation for GPT-Image-1.5 and Doubao Seedream 5.0 under the fixed three-exemplar convention-stabilized protocol.
WAN2.7-Image was also probed under the evaluated normal access setting, but its outputs did not instantiate a scoreable dense normal-map representation.
Specifically, the probes produced stylized scene renderings, normal-map-like colorizations, or irregular RGB channel patterns rather than spatially aligned RGB-encoded surface-orientation fields.
We therefore document WAN2.7-Image as a normal-representation failure diagnostic and exclude it from angular-error aggregation.
For GPT-Image-1.5, the pipeline uses an image-generation/editing endpoint and normalizes the input to a fixed editing canvas before post-processing the generated output back to the target image region.
For Doubao Seedream 5.0, the pipeline uses the Volcengine Ark image-generation endpoint and processes each retained benchmark image independently under the same ordered exemplar-conditioned input protocol.
All pipelines record the input mode, model identifier, prompt version, exemplar identifier, input order, output path, and request metadata in generation manifests.

\paragraph{Protocol finalization.}
Normal-map generation is more sensitive to prompt design than albedo generation because the output must satisfy both geometric consistency and a stable image-wide normal-encoding convention. 
Pilot experiments were used to finalize the exemplar set, target convention, prompt template, and validity checks. 
No benchmark-retained test results were used for selecting or modifying the exemplar set. 
Once finalized, the same prompt template and the same three exemplar pairs were applied unchanged to all benchmark images for each reported proprietary system. 
Alternative exemplar pairs are not selected post hoc, and no single best exemplar is used to redefine the main normal protocol.

\paragraph{GPT-Image-1.5 pipeline.}
GPT-Image-1.5 uses an image-editing interface. For each exemplar-conditioned run, the query RGB image, the corresponding exemplar RGB image, and the corresponding exemplar normal map are provided as visual inputs. To reduce aspect-ratio and canvas drift, inputs are placed on a fixed editing canvas with edge-replicated padding. After generation, the query-region crop is extracted and resized to the benchmark resolution before evaluation.

\paragraph{Doubao Seedream 5.0 pipeline.}
Doubao Seedream 5.0 uses the Ark image-generation interface. For each exemplar-conditioned run, the model receives the corresponding exemplar RGB image, the corresponding exemplar normal map, and the query RGB image as ordered visual inputs.

\paragraph{Prompt constraints.}
The prompt defines the task as generation of a dense view-space surface normal map, not a relit image, edge map, semantic segmentation map, or stylized rendering. It instructs the model to use color only for surface orientation, suppress material color and texture, remove illumination residue, and preserve visible geometry and true shape discontinuities.

\paragraph{Evaluation handling.}
All generated normal maps are resized to the benchmark target resolution and evaluated with the official normal-map protocol described in Appendix~\ref{app:normal_eval}. RGB values are mapped from \([0,1]\) to \([-1,1]\), normalized to unit length, and evaluated using angular-error metrics. Acc@22.5 is reported as the primary directional-accuracy metric for normal estimation.

\subsection{Normal Exemplar-Pair Selection}
\label{app:normal_exemplar_selection}

The three RGB--normal exemplar pairs used in the official normal protocol are selected before benchmark scoring and are shared across all proprietary systems and all benchmark images. 
They are used only to communicate the normal-map encoding convention and the expected dense-map transformation behavior. 
They are not query-specific references, are not selected from benchmark-retained test images, and are not changed based on per-model performance.

The exemplar set is intentionally heterogeneous. 
Rather than using three visually similar examples, we select three indoor pairs with different scene layouts, surface-orientation distributions, boundary complexity, and object-scale composition. 
This reduces dependence on a single scene type or visual style and makes the reported normal results less sensitive to one arbitrary demonstration. 
The final reported proprietary-system score is the arithmetic average over all three exemplar-conditioned runs, while the per-exemplar rows are shown to expose residual sensitivity.

\begin{figure}[!tbp]
\centering
\setlength{\tabcolsep}{2pt}
\renewcommand{\arraystretch}{1.00}

\begin{tabular}{c c c}
\textbf{Exemplar} & \textbf{RGB } & \textbf{Normal map} \\

\textbf{E1} &
\includegraphics[width=0.22\textwidth]{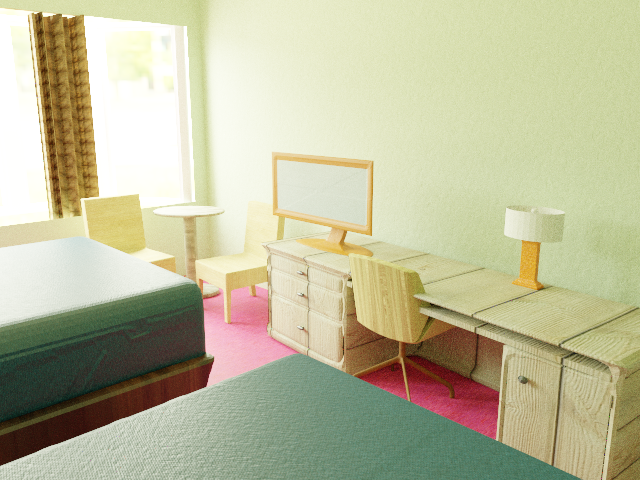} &
\includegraphics[width=0.22\textwidth]{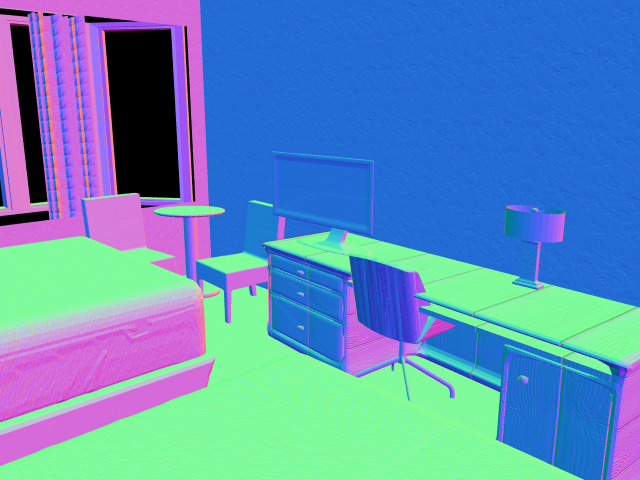} \\

\textbf{E2} &
\includegraphics[width=0.22\textwidth]{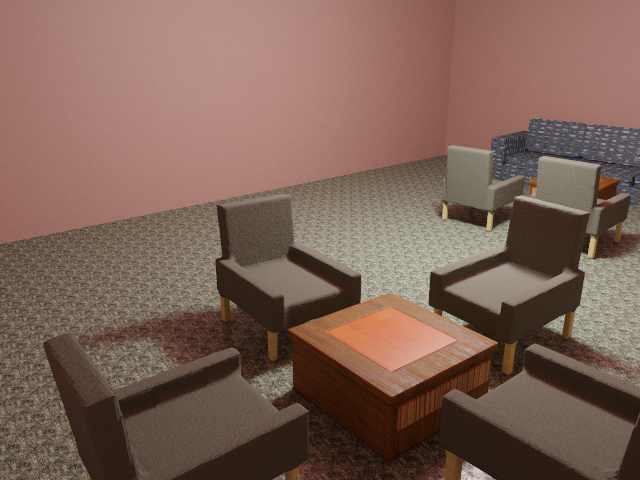} &
\includegraphics[width=0.22\textwidth]{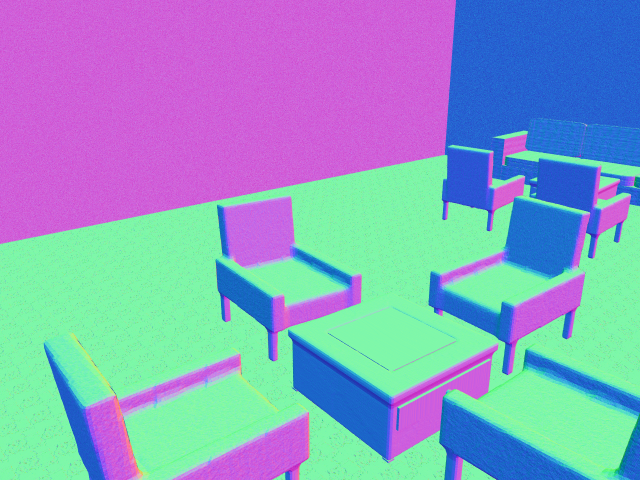} \\

\textbf{E3} &
\includegraphics[width=0.22\textwidth]{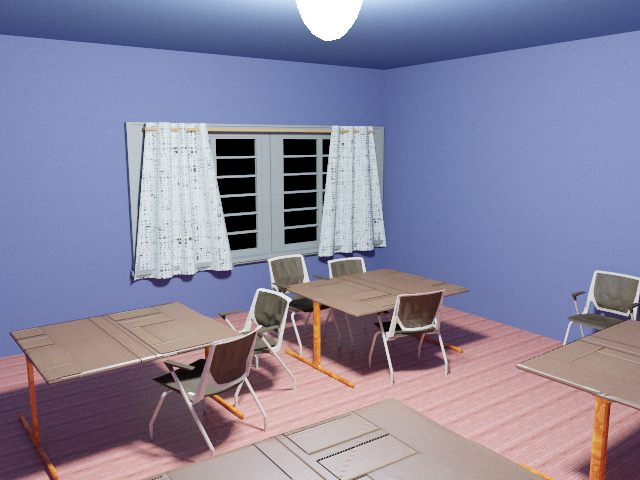} &
\includegraphics[width=0.22\textwidth]{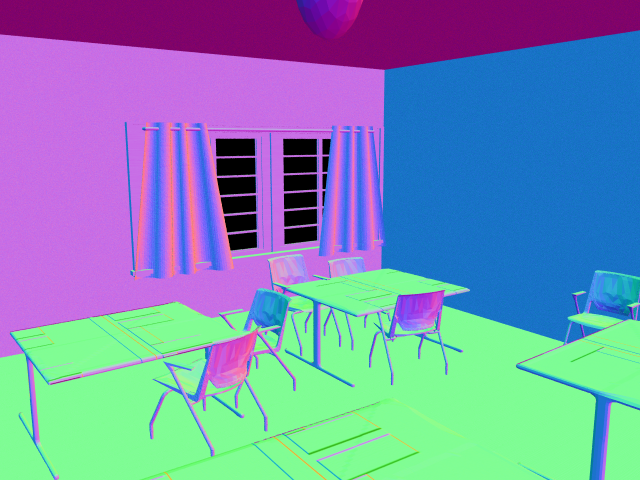} \\
\end{tabular}

\caption{The three fixed RGB--normal exemplar pairs used in the main normal-evaluation protocol. 
Each row shows one exemplar pair, with the RGB image on the left and the corresponding normal map on the right. 
These exemplar pairs are shared across all proprietary systems and benchmark images and are used only to communicate the normal-map convention.}
\label{fig:normal_exemplar_pairs}
\end{figure}

\subsection{Roughness Direct-Prediction Protocol}
\label{app:roughness_generation_pipeline}

This subsection documents the proprietary-system implementation used for roughness prediction. Roughness is evaluated primarily under the controlled direct RGB-to-map setting. Each benchmark image is processed independently, and the output is expected to be a single-channel grayscale roughness map spatially aligned with the query RGB image. Black denotes very smooth or polished surfaces, while white denotes rough and diffuse surfaces.

\paragraph{Controlled RGB-only setting.}
The official roughness comparison uses the controlled direct RGB-to-map track. In this setting, the model receives only the query RGB image and is asked to generate the corresponding roughness map. No segmentation mask, region-filling prior, paired RGB--roughness exemplar, ground-truth annotation, benchmark label, multi-view input, or auxiliary information derived from other images is provided. This setting is used for the main roughness comparison reported in Table~\ref{tab:roughness_mainaxis_openrooms}.

\paragraph{Prompt constraints.}
The roughness prompt enforces a material-parameter interpretation rather than a grayscale rendering of the input image. The model is instructed to preserve the exact scene layout and object presence, to output only one grayscale roughness map, and to avoid copying RGB brightness, cast shadows, ambient occlusion, reflections, highlights, or low-frequency illumination gradients into the roughness output. The prompt also emphasizes that roughness should be inferred from material appearance, highlight sharpness, reflection behavior, coating cues, and visible surface micro-structure, rather than from brightness alone.

\paragraph{Evaluated proprietary systems.}
We evaluate roughness generation for GPT-Image-1.5, WAN2.7-Image, and Doubao Seedream 5.0 using target-specific API wrappers. 
For GPT-Image-1.5, the pipeline uses an image-generation/editing endpoint and normalizes the input to a fixed editing canvas before post-processing the generated output back to the target image region. 
For WAN2.7-Image, the pipeline uses the image-conditioned editing endpoint and processes each retained benchmark image independently under the controlled RGB-only access setting. 
For Doubao Seedream 5.0, the pipeline uses the Volcengine Ark image-generation endpoint and processes each retained benchmark image independently. 
All pipelines record the input mode, model identifier, prompt version, output path, and request metadata in generation manifests.

\paragraph{Diagnostic prior variants.}
In addition to the controlled RGB-only setting, we evaluate diagnostic variants that alter the access assumptions. The \textit{RGB + segmentation} variant provides an externally computed segmentation prior as a soft spatial cue. This prior is not treated as ground-truth material segmentation and is used only to test whether coarse spatial grouping helps roughness estimation. The \textit{RGB + exemplar} variant provides a fixed RGB--roughness reference pair in addition to the target RGB image. These variants are diagnostic only: they are not used for the official controlled-track roughness ranking because they introduce auxiliary spatial or reference information beyond the query RGB image.

\paragraph{Output handling.}
Generated outputs are saved as image files and evaluated by the roughness evaluator in Appendix~\ref{app:roughness_eval}. Predictions are resized to the ground-truth resolution, converted to grayscale, normalized to \([0,1]\), and scored on the material-valid region. The main reported roughness metrics are RMSE, MAE, SSIM, and PSNR, with RMSE used as the primary scalar ranking metric and LPIPS retained only as an auxiliary perceptual diagnostic.

\subsection{Metallic Direct-Prediction Protocol}
\label{app:metallic_generation_pipeline}

For proprietary generative systems, official metallic prediction is evaluated under a controlled RGB-only access setting. 
Each model receives the target RGB image and a fixed task instruction that defines the metallic-map convention. 
No per-image text prior, segmentation mask, region-level material prior, paired reference image, or external example is used in the official metallic comparison. 
The output is required to be a spatially aligned single-channel grayscale metallic map.

\paragraph{Official access mode.}
All official metallic runs use \texttt{rgb\_only}. 
Although the generation scripts support multiple input modes for diagnostic purposes, the official benchmark setting uses only the target RGB image. 
The \texttt{rgb\_plus\_prompt} mode, which adds a per-image or built-in text prior, and the \texttt{rgb\_plus\_example} mode, which adds a fixed RGB--metallic reference pair, are excluded from the main metallic comparison and are reported only as diagnostic access-setting variants when used.

\paragraph{Metallic-map convention.}
The fixed task instruction defines metallic as a sparse bounded material map rather than a grayscale reconstruction of scene appearance. 
Black denotes dielectric or non-metal surfaces, while white denotes clearly exposed metallic material. 
Because most indoor pixels are non-metallic, the instruction emphasizes conservative near-black predictions under ambiguity. 
It also discourages common invalid outputs such as relit grayscale images, edge maps, contour maps, reflectance sketches, or low-gray structure-preserving renderings. 
Bright highlights, mirror-like appearance, glossy reflections, direct lighting, shadows, and smoothness alone are not treated as sufficient evidence of metal.

\paragraph{Diagnostic access modes.}
The diagnostic \texttt{rgb\_plus\_prompt} setting adds a text prior to the same target RGB image. 
For Doubao Seedream 5.0 and GPT-Image-1.5, this mode reads a per-image prompt from a prompt directory matched by filename. 
For WAN2.7-Image, the final script uses a built-in soft prior rather than reading per-image prompt files. 
The diagnostic \texttt{rgb\_plus\_example} setting provides a fixed reference RGB image and a fixed reference metallic map in addition to the target RGB image. 
This reference is used only as output-style guidance and is not part of the official metallic access setting. 
These diagnostic modes are useful for analyzing whether textual or paired-reference priors help proprietary systems avoid metallic-map failure modes, but they instantiate different evidence conditions and are therefore not mixed with the official RGB-only results.

\paragraph{System-specific execution.}
For \textbf{Doubao Seedream 5.0}, we use the final \texttt{gate\_black} metallic prompt variant with the \texttt{v3\_visualprior\_noboundary} preset. 
Official runs are invoked with \texttt{--input\_mode rgb\_only}; the script also retains \texttt{rgb\_plus\_prompt} and \texttt{rgb\_plus\_example} for diagnostics. 
The generated image is converted to single-channel grayscale PNG before evaluation, and the output metadata record the model identifier, input mode, prompt preset, prompt text, output-size policy, and reference paths when applicable.

For \textbf{GPT-Image-1.5}, official runs also use \texttt{--input\_mode rgb\_only}. 
The implementation uses an image-editing call and applies letterbox preprocessing to match the model canvas. 
After generation, the output is cropped back to the original image region and saved as a single-channel grayscale PNG. 
The metadata record the model identifier, input mode, prompt preset, quality setting, seed, requested canvas size, and crop-to-original output policy.

For \textbf{WAN2.7-Image}, official runs use \texttt{--input\_mode rgb\_only} with the image-conditioned editing endpoint. 
The script retains a built-in soft-prior \texttt{rgb\_plus\_prompt} mode and an \texttt{rgb\_plus\_example} mode for diagnostics only. 
The output is saved as a grayscale PNG, and the metadata record the input mode, prompt version, requested output size, seed, request identifier, returned size, and output URL.

\paragraph{Output handling and evaluation.}
For all proprietary metallic generators, predictions are written to a dedicated \texttt{metallic/} directory, while run-level and per-image metadata are written under \texttt{meta/}. 
Each generated map is named by the RGB stem with the suffix \texttt{\_metallic.png}. 
Before evaluation, metallic predictions are converted to single-channel grayscale, resized to the ground-truth resolution if needed, clipped to the valid target range, and scored only on the released material-valid region according to Appendix~\ref{app:metallic_evaluation}. 
The official metallic comparison therefore remains a controlled RGB-only direct-prediction setting; prompt-prior and exemplar-assisted variants are diagnostic-only.

\subsection{Diagnostic Access Variants}
\label{app:ablation_protocols}

Diagnostic experiments analyze the role of auxiliary conditioning, convention stabilization, and automatically derived priors.  These experiments are not used to define the official access setting for any target and are not mixed with the main comparison results unless explicitly stated otherwise.

\paragraph{Albedo auxiliary-conditioning ablation.}
For albedo, we compare the disclosed structure-preserving albedo elicitation protocol against reduced variants with weaker prompting or without query-image-only auxiliary analysis. The purpose is to determine whether structured conditioning derived only from the same query image improves illumination removal and material preservation, while keeping the task within the same single-image evidence constraint.

\paragraph{Normal exemplar ablation.}
For normal estimation, we compare the prompt-only setting against the fixed shared exemplar setting under otherwise matched conditions. This ablation is used to assess whether paired exemplars stabilize the normal-map convention, reduce invalid outputs, and improve metric performance.

\paragraph{Prior-assisted roughness / metallic variants.}
For roughness and metallic, we evaluate several prior-assisted variants, including segmentation-soft prompting, segmentation-based region filling, and paired-reference variants when applicable. These settings alter access assumptions relative to the controlled RGB-to-map material protocol and are therefore treated only as diagnostics.

\FloatBarrier
\subsection{Proprietary Model Versioning}
\label{app:proprietary_versioning}

\begin{table}[!htbp]
\centering
\scriptsize
\caption{Versioning information and target-wise run status for proprietary-system evaluations.}
\label{tab:proprietary_versioning}
\setlength{\tabcolsep}{2.6pt}
\renewcommand{\arraystretch}{1.08}

\begin{tabularx}{\linewidth}{@{}L{1.75cm} L{3.15cm} Z{0.55cm} Z{0.55cm} Z{0.55cm} Z{0.60cm} Z{0.55cm} X@{}}
\toprule
System
& API / backend
& D
& N
& A
& R
& M
& Notes \\
\midrule

GPT-Image-1.5
&
\texttt{gpt-image-1.5}\newline
Azure OpenAI / OpenAI-compatible
& \textsc{C}
& \textsc{C}
& \textsc{C}
& \textsc{C}
& \textsc{C}
& Complete official retained-split runs for all evaluated targets. \\

GPT-Image-2.0
&
\texttt{gpt-image-2.0}\newline
Azure OpenAI / OpenAI-compatible
& \textsc{C}
& \textsc{Diag}
& \textsc{C}
& \textsc{C}
& \textsc{C}
& Complete official retained-split runs are available for depth, albedo, roughness, and metallic.
No complete retained-split run satisfying the full three-exemplar normal protocol is present in the frozen manifest.
Available normal probes are documented as diagnostic-only and are not mixed into the official normal aggregate. \\

Doubao Seedream 5.0
&
\texttt{doubao-seedream-5-0-260128}\newline
Volcengine Ark
& \textsc{C}
& \textsc{C}
& \textsc{C}
& \textsc{C}
& \textsc{C}
& Complete official retained-split runs for all evaluated targets. \\

WAN2.7-Image
&
\texttt{wan2.7-image}\newline
DashScope
& \textsc{C}
& \textsc{Fail}
& \textsc{C}
& --
& \textsc{C}
& Complete official runs are available for depth, albedo, and metallic.
For normal, outputs failed to instantiate the dense RGB-encoded normal-map representation and are documented as representation failures rather than converted into angular scores. \\

Qwen-Image-2.0
&
\texttt{Qwen-Image-2.0}\newline
DashScope
& --
& --
& --
& \textsc{C}
& --
& Complete official retained-split run for roughness only. \\

Doubao Seed 2.0 Pro
&
\texttt{doubao-seed-2-0-pro-260215}\newline
Volcengine Ark
& --
& --
& \textsc{Aux}
& --
& --
& Used only to generate shared query-image-only auxiliary-analysis records for the disclosed albedo protocol; not evaluated as an image editor. \\

\bottomrule
\end{tabularx}

\tabremark{
D/N/A/R/M denote depth, normal, albedo, roughness, and metallic, respectively.
\textsc{C} denotes a complete official retained-split run included in the corresponding main-axis aggregation when scoreable.
\textsc{Diag} denotes an incomplete diagnostic-only probe.
\textsc{Fail} denotes a representation-convention failure that is documented but not assigned artificial metric scores.
\textsc{Aux} denotes auxiliary use for protocol construction rather than image-editor evaluation.
``--'' denotes that no complete official retained-split run was included under the corresponding target protocol.
Incomplete probes, diagnostic-only runs, unavailable complete runs, and representation-convention failures are documented here rather than silently omitted.
}
\end{table}

\FloatBarrier

\subsection{Manifest and Failure Logging}
\label{app:manifest_logging}

All proprietary generation scripts maintain explicit metadata records for reproducibility. For each run, the pipeline writes a manifest file and a skip log under the output metadata directory. Completed generations store image name, relative path, model identifier, output path, prompt text, and protocol-specific settings. For multi-exemplar normal generation, records include the exemplar-pair identifier, one-shot mode, exemplar paths, input order, and interface-specific fields when available. Skipped samples record relative path, processing stage, error code, error type, message, part index, and timestamp.

\section{Additional Quantitative Results}
\label{app:additional_results}

This appendix reports additional quantitative results that complement the compact main-paper summaries. 
These tables do not introduce new access assumptions unless explicitly marked as diagnostic settings. 
Source-specific results, main-axis uncertainty analyses, stress-subset results, and diagnostic / ablation results are grouped here so that definitions and protocols remain separate from empirical tables.

\subsection{Source-Specific Main-Axis Results}
\label{app:source_specific_results}

\subsubsection{Depth Source-Specific Results}
\label{app:depth_source_results}

\begin{table}[!htbp]
\centering
\caption{Relative-depth results on the OpenRooms-FF main-axis subset under the affine-invariant single-image evaluation protocol.}
\label{tab:depth_mainaxis_openroomsff}
\footnotesize
\setlength{\tabcolsep}{4.5pt}
\renewcommand{\arraystretch}{0.95}
\begin{tabular}{lcccccc}
\toprule
Method
& AbsRel-AI$\downarrow$
& RMSE-AI$\downarrow$
& MAE-AI$\downarrow$
& $\delta_1$-AI$\uparrow$
& $\delta_2$-AI$\uparrow$
& Boundary F1$\uparrow$ \\
\midrule
GPT-Image-2.0~\cite{openai2026gpt_image2}
& 0.102 & 0.340 & 0.268 & 0.901 & 0.978 & 0.315 \\
GPT-Image-1.5~\cite{openai2025gpt_image15}
& 0.134 & 0.486 & 0.373 & 0.834 & 0.965 & 0.038 \\
WAN2.7-Image~\cite{wan27_image}
& 0.097 & 0.346 & 0.271 & 0.915 & 0.981 & 0.213 \\
Doubao Seedream 5.0~\cite{seedream5_lite}
& 0.121 & 0.447 & 0.344 & 0.866 & 0.977 & 0.166 \\
\midrule
Depth Anything 3~\cite{depth_anything3}
& \textbf{0.012} & \textbf{0.081} & \textbf{0.033} & \textbf{0.996} & \textbf{0.999} & 0.304 \\
Lotus Depth~\cite{Lotus}
& 0.031 & 0.121 & 0.078 & 0.993 & 0.999 & \textbf{0.384} \\
\bottomrule
\end{tabular}

\tabremark{
Scores are per-image averages over the 1,125-image OpenRooms-FF main-axis subset.
Lower AbsRel-AI, RMSE-AI, and MAE-AI are better; higher $\delta_1$-AI, $\delta_2$-AI, and Boundary F1 are better.
Boldface marks the best result in each metric column.
}
\end{table}

\begin{table}[!htbp]
\centering
\caption{Relative-depth results on the InteriorVerse main-axis subset under the affine-invariant single-image evaluation protocol.}
\label{tab:depth_mainaxis_interiorverse}
\footnotesize
\setlength{\tabcolsep}{4.5pt}
\renewcommand{\arraystretch}{0.95}
\begin{tabular}{lcccccc}
\toprule
Method
& AbsRel-AI$\downarrow$
& RMSE-AI$\downarrow$
& MAE-AI$\downarrow$
& $\delta_1$-AI$\uparrow$
& $\delta_2$-AI$\uparrow$
& Boundary F1$\uparrow$ \\
\midrule
GPT-Image-2.0~\cite{openai2026gpt_image2}
& 0.234 & 0.399 & 0.307 & 0.776 & 0.925 & 0.228 \\
GPT-Image-1.5~\cite{openai2025gpt_image15}
& 0.334 & 0.525 & 0.403 & 0.686 & 0.899 & 0.017 \\
WAN2.7-Image~\cite{wan27_image}
& 0.234 & 0.372 & 0.286 & 0.808 & 0.942 & 0.150 \\
Doubao Seedream 5.0~\cite{seedream5_lite}
& 0.222 & 0.469 & 0.359 & 0.727 & 0.924 & 0.106 \\
\midrule
Depth Anything 3~\cite{depth_anything3}
& \textbf{0.058} & \textbf{0.154} & \textbf{0.092} & \textbf{0.966} & \textbf{0.989} & 0.176 \\
Lotus Depth~\cite{Lotus}
& 0.065 & 0.165 & 0.104 & 0.964 & 0.988 & \textbf{0.252} \\
\bottomrule
\end{tabular}

\tabremark{
Scores are per-image averages over the 2,109-image InteriorVerse main-axis subset.
Lower AbsRel-AI, RMSE-AI, and MAE-AI are better; higher $\delta_1$-AI, $\delta_2$-AI, and Boundary F1 are better.
Boldface marks the best result in each metric column.
}
\end{table}

\begin{figure}[!htbp]
\centering
\includegraphics[width=1.0\textwidth]{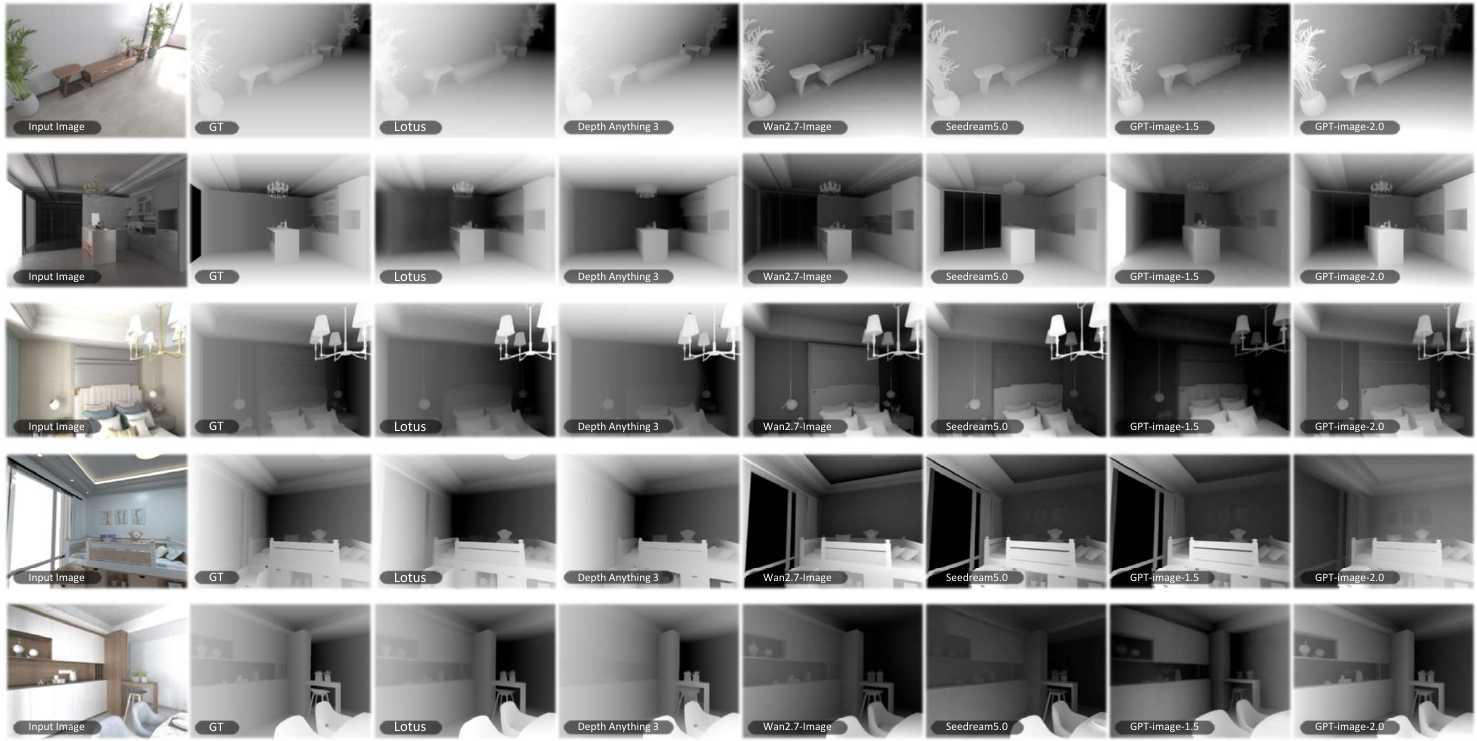}
\caption{Representative qualitative examples for depth. Each row shows the input RGB
image, ground-truth map, and predictions from representative specialized predictors and proprietary image editors for
one target.}
\label{fig:depth_example}
\end{figure}

\paragraph{Depth source-specific interpretation.}
Tables~\ref{tab:depth_mainaxis_openroomsff} and~\ref{tab:depth_mainaxis_interiorverse} report the two source-specific components of the source-balanced depth table.
On both OpenRooms-FF and InteriorVerse, specialized depth predictors remain strongest on the scalar relative-depth metrics.
\textbf{Depth Anything 3} gives the best AbsRel-AI, RMSE-AI, MAE-AI, and threshold accuracies on both sources, while \textbf{Lotus Depth} gives the best Boundary F1.
Among proprietary editors, \textbf{WAN2.7-Image} gives the best AbsRel-AI and threshold accuracies on OpenRooms-FF, while \textbf{GPT-Image-2.0} gives the best RMSE-AI, MAE-AI, and Boundary F1.
On InteriorVerse, \textbf{Doubao Seedream 5.0} gives the best proprietary AbsRel-AI, while \textbf{GPT-Image-2.0} gives the best proprietary Boundary F1.
\textbf{GPT-Image-2.0} substantially improves over \textbf{GPT-Image-1.5}, especially in Boundary F1, but remains behind specialized predictors on the primary scalar metric.
These results support the source-balanced conclusion that scalar relative-depth accuracy and local discontinuity recovery capture different aspects of depth prediction. Figure~\ref{fig:depth_example} provides representative qualitative depth comparisons that complement the two source-specific tables.

\subsubsection{Normal Source-Specific Results}
\label{app:normal_source_results}

\begin{table*}[!htbp]
\centering
\caption{Surface normal estimation results on OpenRooms-FF.}
\label{tab:normal_mainaxis_openrooms}
\compactappendixtable
\begin{tabular}{llcccccc}
\toprule
Method & Exemplar & Mean$\downarrow$ & Median$\downarrow$ & RMSE$\downarrow$ & Acc@11.25$\uparrow$ & Acc@22.5$\uparrow$ & Acc@30$\uparrow$ \\
\midrule
\multirow{4}{*}{\makecell[c]{GPT-Image-1.5}}
& E1  & 41.607 & 38.711 & 46.342 & 0.044 & 0.198 & 0.343 \\
& E2  & 39.082 & 35.578 & 44.373 & 0.071 & 0.257 & 0.408 \\
& E3  & 40.496 & 37.013 & 45.809 & 0.067 & 0.237 & 0.378 \\
& Avg & 40.395 & 37.100 & 45.508 & 0.060 & 0.231 & 0.376 \\
\midrule
\multirow{4}{*}{\makecell[c]{Doubao\\Seedream 5.0}}
& E1  & 46.427 & 39.429 & 53.654 & 0.058 & 0.195 & 0.317 \\
& E2  & 46.886 & 40.098 & 53.719 & 0.045 & 0.176 & 0.305 \\
& E3  & 47.834 & 41.661 & 54.862 & 0.043 & 0.166 & 0.282 \\
& Avg & 47.049 & 40.396 & 54.079 & 0.049 & 0.179 & 0.301 \\
\midrule
Lotus
& -- & \textbf{8.697} & \textbf{5.187} & \textbf{14.223} & \textbf{0.815} & 0.923 & 0.948 \\
StableNormal
& -- & 9.583 & 6.808 & 14.393 & 0.784 & \textbf{0.926} & \textbf{0.953} \\
\bottomrule
\end{tabular}

\tabremark{
Scores are per-image averages within the source.
For proprietary systems, E1--E3 denote the three fixed RGB--normal exemplar-conditioned runs, and Avg denotes their arithmetic mean.
Specialized predictors do not use exemplars.
Acc@22.5 is the primary metric.
Boldface marks the best value for each metric.
}
\end{table*}

\begin{table*}[!htbp]
\centering
\caption{Surface normal estimation results on InteriorVerse.}
\label{tab:normal_mainaxis_interiorverse}
\compactappendixtable
\begin{tabular}{llcccccc}
\toprule
Method & Exemplar & Mean Ang.$\downarrow$ & Median Ang.$\downarrow$ & RMSE$\downarrow$ & Acc@11.25$\uparrow$ & Acc@22.5$\uparrow$ & Acc@30$\uparrow$ \\
\midrule
\multirow{4}{*}{\makecell[c]{GPT-Image-1.5}}
& E1  & 49.793 & 48.604 & 55.642 & 0.053 & 0.179 & 0.278 \\
& E2  & 48.761 & 47.584 & 54.409 & 0.054 & 0.181 & 0.277 \\
& E3  & 49.078 & 47.612 & 54.679 & 0.050 & 0.173 & 0.271 \\
& Avg & 49.211 & 47.933 & 54.910 & 0.052 & 0.178 & 0.275 \\
\midrule
\multirow{4}{*}{\makecell[c]{Doubao\\Seedream 5.0}}
& E1  & 66.557 & 61.359 & 74.227 & 0.021 & 0.084 & 0.151 \\
& E2  & 67.539 & 62.417 & 74.138 & 0.013 & 0.059 & 0.118 \\
& E3  & 62.217 & 55.979 & 70.467 & 0.031 & 0.115 & 0.193 \\
& Avg & 65.438 & 59.918 & 72.944 & 0.021 & 0.086 & 0.154 \\
\midrule
Lotus
& -- & 14.661 & 9.423 & 22.690 & 0.726 & 0.803 & 0.827 \\
StableNormal
& -- & \textbf{12.850} & \textbf{8.591} & \textbf{20.325} & \textbf{0.770} & \textbf{0.826} & \textbf{0.846} \\
\bottomrule
\end{tabular}

\tabremark{
Scores are per-image averages within the source.
For proprietary systems, E1--E3 denote the three fixed RGB--normal exemplar-conditioned runs, and Avg denotes their arithmetic mean.
Specialized predictors do not use exemplars.
Acc@22.5 is the primary metric.
Boldface marks the best value for each metric.
}
\end{table*}

\paragraph{Normal source-specific interpretation.}
Tables~\ref{tab:normal_mainaxis_openrooms} and~\ref{tab:normal_mainaxis_interiorverse} provide source-specific breakdowns under the same multi-exemplar convention-stabilized normal protocol used in the pooled main table.
For proprietary systems, each source-specific table reports the three fixed exemplar-conditioned runs and their aggregate score.
The results confirm that the source-balanced conclusion is not driven by a single source or by a single exemplar pair: \textbf{Lotus} and \textbf{StableNormal} remain substantially stronger than proprietary generative systems on both OpenRooms-FF and InteriorVerse.

This conclusion holds across metric families.
Specialized normal predictors achieve much lower mean and median angular errors while also obtaining substantially higher Acc@11.25, Acc@22.5, and Acc@30.
Thus, the gap is not merely an artifact of using a strict threshold metric; rather, it reflects broadly stronger dense directional recovery under the shared evaluator.
The proprietary systems can follow the normal-map convention to a limited extent under exemplar conditioning, but they still fall far short of specialized predictors in both average angular accuracy and the fraction of locally correct normal vectors.

\begin{figure}[!htbp]
\centering
\includegraphics[width=0.72\textwidth]{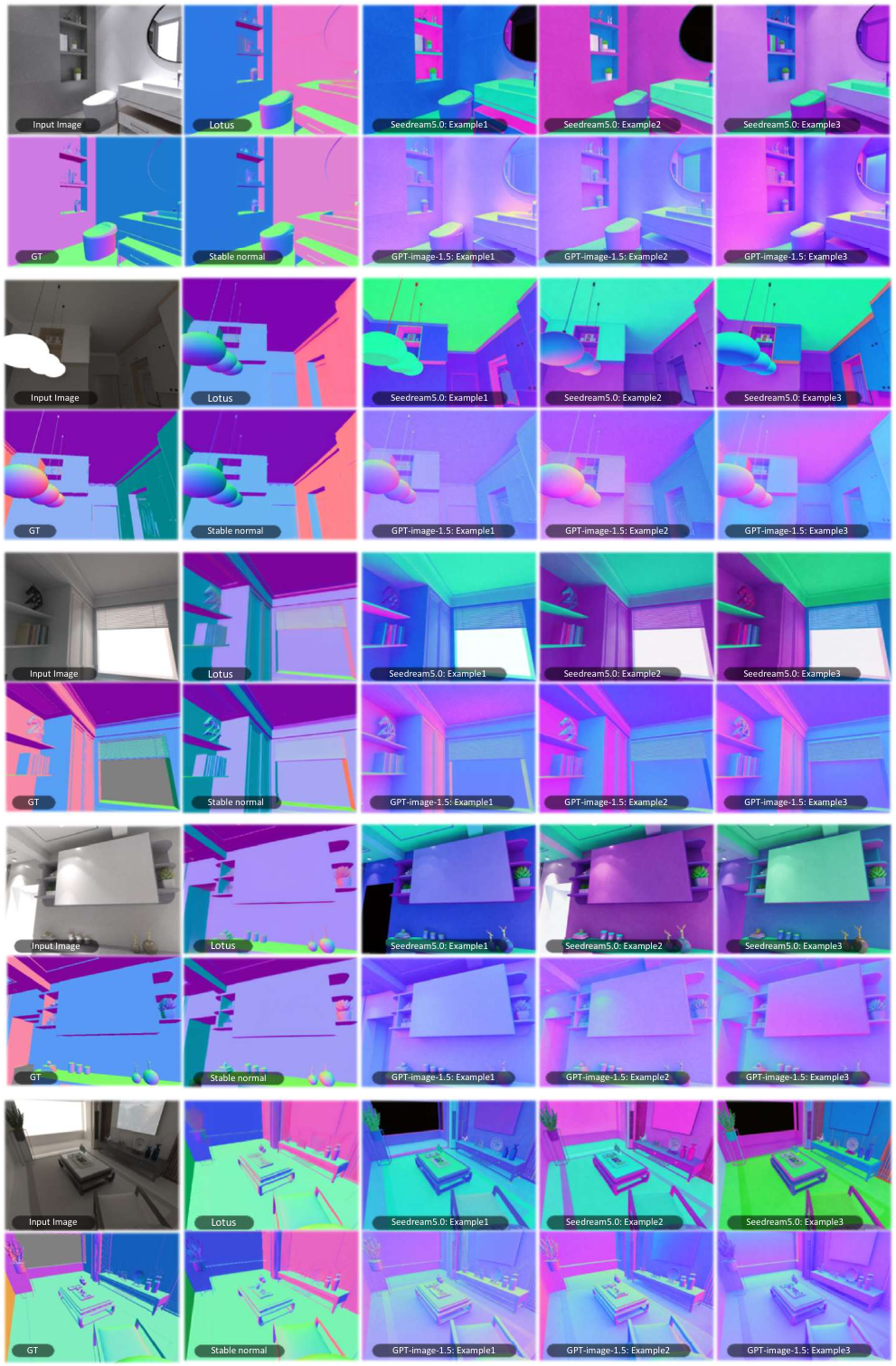}
\caption{Representative qualitative examples for normal. Each two rows show the input RGB
image, ground-truth map, and predictions from representative specialized predictors and proprietary image editors for
one target.}
\label{fig:normal_example}
\end{figure}

The cross-source results also show that InteriorVerse remains the more challenging source for proprietary generative systems.
We treat this as a source-specific robustness observation rather than as a causal claim about dataset properties or interface design. Figure~\ref{fig:normal_example} provides representative qualitative normal examples for this source-specific comparison.

\subsubsection{Albedo Source-Specific Results}
\label{app:albedo_source_results}

\begin{table}[!htbp]
\centering
\caption{Albedo estimation results on OpenRooms-FF under the disclosed structure-preserving albedo elicitation protocol. Scores are reported as per-image averages within the source. Lower MAE and LPIPS are better, while higher PSNR and SSIM are better. Boldface marks the numerical best for each metric; interpretation follows the paired scalar--structural profile rather than MAE alone.}
\label{tab:albedo_mainaxis_openrooms}
\footnotesize
\setlength{\tabcolsep}{5pt}
\renewcommand{\arraystretch}{0.95}
\begin{tabular}{lcccc}
\toprule
Method & MAE$\downarrow$ & PSNR$\uparrow$ & SSIM$\uparrow$ & LPIPS$\downarrow$ \\
\midrule
GPT-Image-2.0~\cite{openai2026gpt_image2}
& 0.158 & 14.302 & 0.485 & 0.408 \\
GPT-Image-1.5 & 0.212 & 12.089 & 0.457 & 0.425 \\
WAN2.7-Image & 0.175 & 13.480 & 0.560 & 0.362 \\
Doubao Seedream 5.0 & 0.164 & 14.061 & 0.586 & 0.320 \\
\midrule
Colorful Diffuse Intrinsic & \textbf{0.126} & \textbf{16.690} & \textbf{0.695} & 0.369 \\
DNF-Intrinsic & 0.132 & 16.389 & 0.654 & \textbf{0.225} \\
\bottomrule
\end{tabular}
\end{table}

\begin{table}[!htbp]
\centering
\caption{Albedo estimation results on InteriorVerse under the disclosed structure-preserving albedo elicitation protocol. Scores are reported as per-image averages within the source. Lower MAE and LPIPS are better, while higher PSNR and SSIM are better. Best results are highlighted in bold.}
\label{tab:albedo_mainaxis_interiorverse}
\footnotesize
\setlength{\tabcolsep}{5pt}
\renewcommand{\arraystretch}{0.95}
\begin{tabular}{lcccc}
\toprule
Method & MAE$\downarrow$ & PSNR$\uparrow$ & SSIM$\uparrow$ & LPIPS$\downarrow$ \\
\midrule
GPT-Image-2.0~\cite{openai2026gpt_image2}
& 0.191 & 12.736 & 0.624 & 0.419 \\
GPT-Image-1.5 & 0.196 & 11.877 & 0.585 & 0.461 \\
WAN2.7-Image & 0.264 & 10.254 & 0.606 & 0.461 \\
Doubao Seedream 5.0 & 0.202 & 12.191 & 0.625 & 0.382 \\
\midrule
Colorful Diffuse Intrinsic & 0.149 & 15.584 & 0.794 & 0.305 \\
DNF-Intrinsic & \textbf{0.056} & \textbf{22.044} & \textbf{0.863} & \textbf{0.126} \\
\bottomrule
\end{tabular}
\end{table}

\paragraph{Albedo source-specific interpretation.}
Tables~\ref{tab:albedo_mainaxis_openrooms} and~\ref{tab:albedo_mainaxis_interiorverse} provide source-specific breakdowns under the same disclosed structure-preserving albedo protocol used in the pooled main table.
The results show that specialized intrinsic methods remain stronger than proprietary generative systems on both sources under the paired scalar--structural metric profile.
This confirms that the source-balanced conclusion is not driven by a single dataset.

\begin{figure}[!htbp]
\centering
\includegraphics[width=1.0\textwidth]{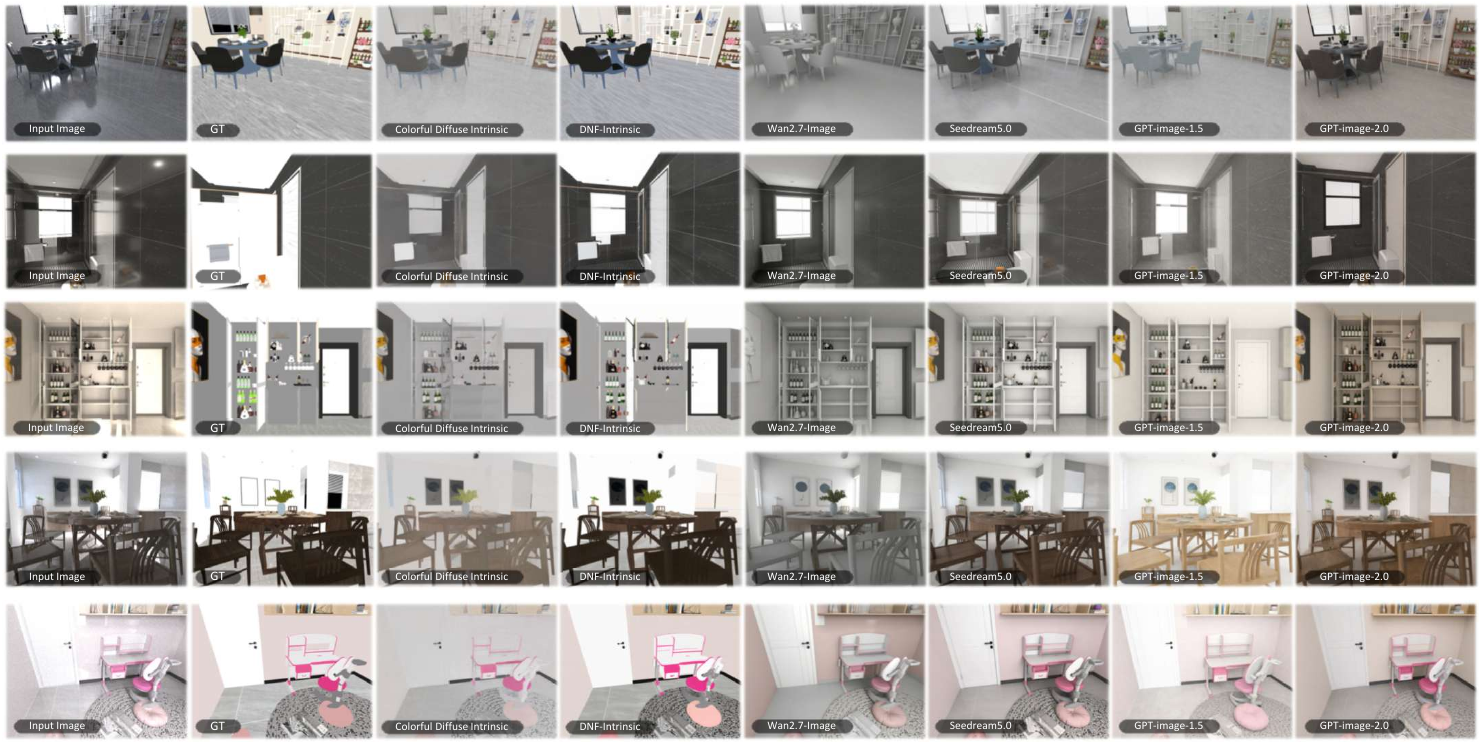}
\caption{Representative qualitative examples for albedo. Each row shows the input RGB
image, ground-truth map, and predictions from representative specialized predictors and proprietary image editors for
one target.}
\label{fig:albedo_example}
\end{figure}

The specialized-method ranking is source-dependent.
On OpenRooms-FF, \textbf{Colorful Diffuse Intrinsic} achieves the best MAE, PSNR, and SSIM, while \textbf{DNF-Intrinsic} is strongest on LPIPS.
On InteriorVerse, \textbf{DNF-Intrinsic} dominates all reported metrics.
This explains why the source-balanced main-axis summary favors \textbf{DNF-Intrinsic} overall while still preserving meaningful source-dependent variation among strong specialized baselines. Figure~\ref{fig:albedo_example} provides representative qualitative albedo examples corresponding to the source-specific tables.

\subsubsection{Roughness Source-Specific Results}
\label{app:roughness_source_results}

For roughness, we used only the OpenRooms-FF dataset. The results are shown in Table~\ref{tab:roughness_mainaxis_openrooms}. Figure~\ref{fig:roughness_example} provides representative qualitative roughness predictions that complement the main source-specific table and make the scalar--structural differences more visible.

\begingroup
\setlength{\floatsep}{8pt plus 2pt minus 2pt}
\begin{figure}[!tbp]
\centering
\includegraphics[width=0.82\textwidth,height=0.36\textheight,keepaspectratio]{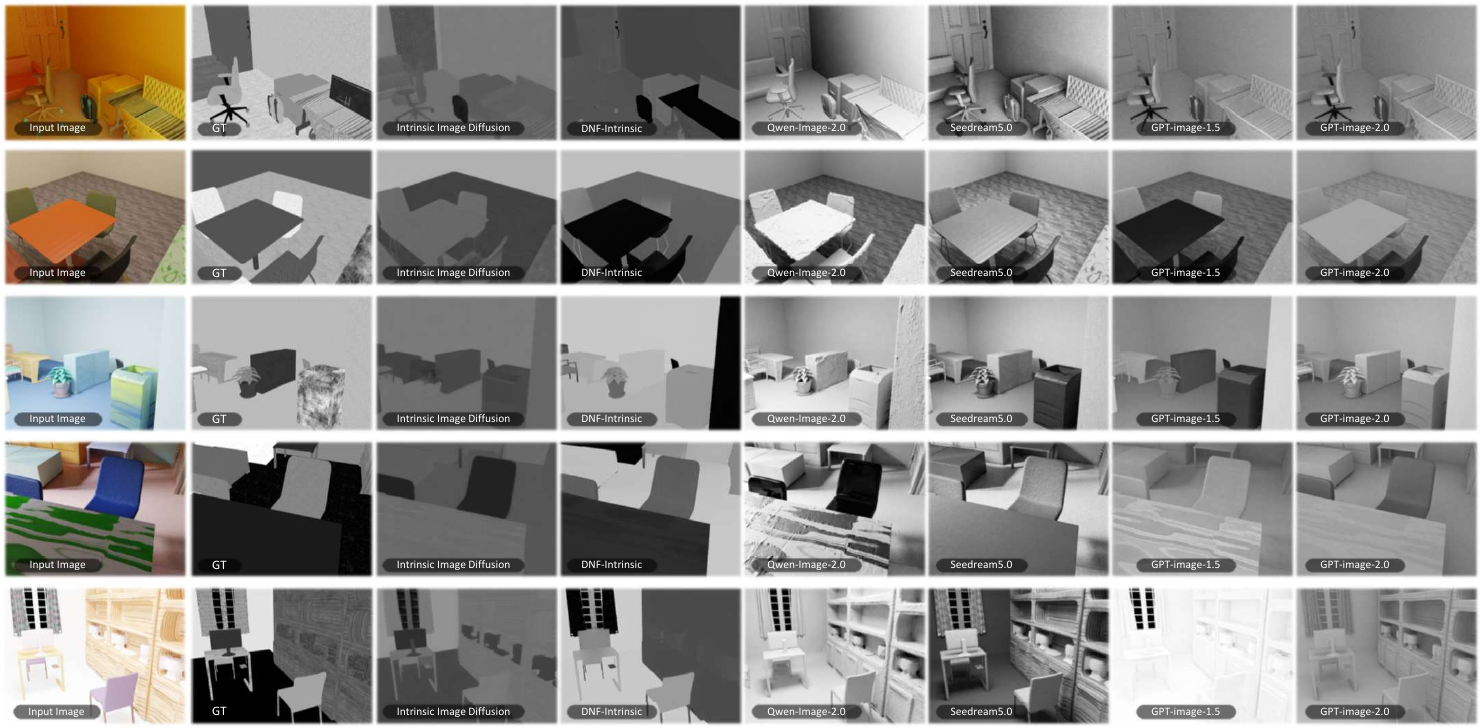}
\caption{Representative qualitative examples for roughness. Each row shows the input RGB
image, ground-truth map, and predictions from representative specialized predictors and proprietary image editors for
one target.}
\label{fig:roughness_example}
\end{figure}

\subsubsection{Metallic Source-Specific Results}
\label{app:metallic_source_results}

For the evaluation of metallic, we used our own custom dataset. Since the proportion of metallic pixels varies across different scene types and is generally sparse in most scenes, we performed uniform sampling across the different scene categories when constructing the dataset. The detailed results are shown in Table~\ref{tab:metallic_mainaxis_companion}. Figure~\ref{fig:metallic_example} provides representative qualitative metallic predictions that help interpret the sparsity-aware behavior discussed in the source-specific results.

\begin{figure}[!tbp]
\centering
\includegraphics[width=0.98\textwidth,height=0.53\textheight,keepaspectratio]{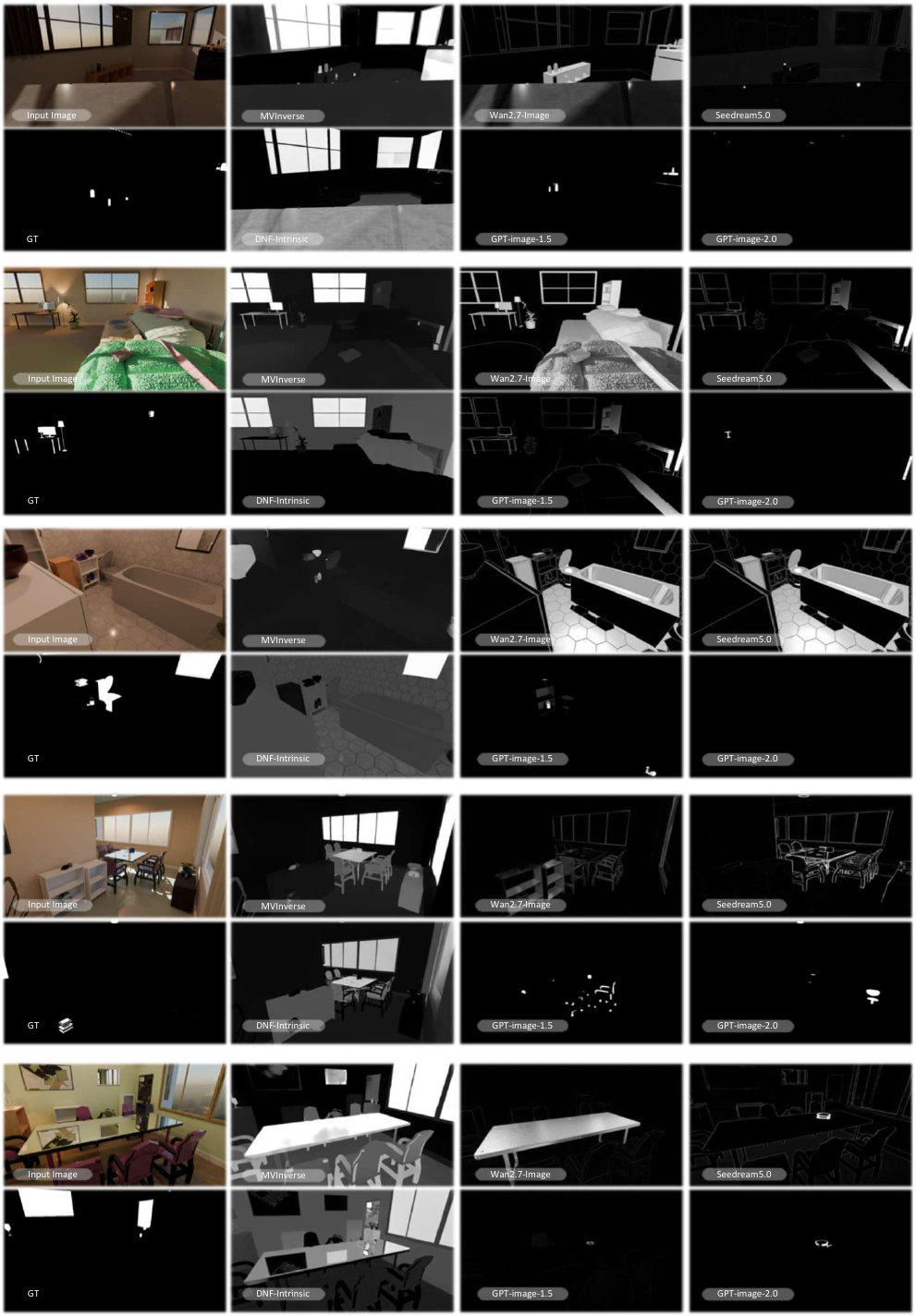}
\caption{Representative qualitative examples for metallic. Each two row shows the input RGB
image, ground-truth map, and predictions from representative specialized predictors and proprietary image editors for
one target.}
\label{fig:metallic_example}
\end{figure}
\endgroup

\FloatBarrier

\subsection{Stress-Subset Results}
\label{app:stress_results}

\subsubsection{Stress-Subset Counts}
\label{app:stress_subset_counts}

We first report the retained-image support for each programmatic stress slice so that the later stress-result tables can be interpreted with the corresponding sample coverage in mind.
These counts clarify which slices are source-balanced and which remain source-specific because of limited retained support in one source.

\begin{table}[!htbp]
\centering
\small
\caption{Retained-image counts for programmatic stress subsets. 
For shared slices, stress scores are computed by first evaluating each source separately and then macro-averaging source-level slice scores. 
The \textit{low\_light} slice is reported only for OpenRooms-FF because InteriorVerse has insufficient retained low-light coverage.}
\label{tab:stress_subset_counts}
\begin{tabular}{lccc}
\toprule
Slice & OR-FF count & IV count & Aggregation \\
\midrule
dark\_heavy & 176 & 472 & Source-balanced macro \\
hdr & 176 & 403 & Source-balanced macro \\
highlight\_heavy & 167 & 376 & Source-balanced macro \\
low\_light & 190 & -- & OR-FF only \\
\bottomrule
\end{tabular}
\end{table}

\subsubsection{Albedo Stress-Subset Results}
\label{app:albedo_stress_results}

Table~\ref{tab:albedo_stress_combined} summarizes how albedo performance changes across the predefined photometric challenge slices.
The main diagnostic question is whether the source-balanced main-axis ordering remains stable once illumination becomes more adverse or source support becomes asymmetric.

\begin{table}[!htbp]
\centering
\caption{Stress-test albedo estimation results under the disclosed structure-preserving albedo protocol.}
\label{tab:albedo_stress_combined}
\compactappendixtable
\begin{tabular}{llcccc}
\toprule
Method & Slice & MAE$\downarrow$ & PSNR$\uparrow$ & SSIM$\uparrow$ & LPIPS$\downarrow$ \\
\midrule
\multirow{4}{*}{\footnotesize GPT-Image-1.5}
& dark\_heavy & 0.197 & 12.433 & 0.495 & 0.440 \\
& hdr & 0.196 & 12.143 & 0.533 & 0.439 \\
& highlight\_heavy & 0.224 & 11.221 & 0.537 & 0.445 \\
& low\_light & 0.163 & 13.781 & 0.439 & 0.405 \\
\cmidrule(lr){1-6}

\multirow{4}{*}{\footnotesize GPT-Image-2.0}
& dark\_heavy & 0.163 & 14.060 & 0.533 & 0.422 \\
& hdr & 0.173 & 13.623 & 0.576 & 0.412 \\
& highlight\_heavy & 0.181 & 13.165 & 0.570 & 0.405 \\
& low\_light & 0.142 & 14.871 & 0.466 & 0.396 \\
\cmidrule(lr){1-6}

\multirow{4}{*}{\footnotesize WAN2.7-Image}
& dark\_heavy & 0.216 & 12.095 & 0.565 & 0.412 \\
& hdr & 0.232 & 11.708 & 0.589 & 0.425 \\
& highlight\_heavy & 0.221 & 11.763 & 0.600 & 0.399 \\
& low\_light & 0.182 & 13.139 & 0.524 & 0.373 \\
\cmidrule(lr){1-6}

\multirow{4}{*}{\makecell[l]{\footnotesize Doubao\\\footnotesize Seedream 5.0}}
& dark\_heavy & 0.179 & 13.446 & 0.583 & 0.361 \\
& hdr & 0.190 & 13.080 & 0.605 & 0.364 \\
& highlight\_heavy & 0.186 & 12.862 & 0.619 & 0.333 \\
& low\_light & 0.159 & 14.344 & 0.559 & 0.325 \\
\cmidrule(lr){1-6}

\multirow{4}{*}{\makecell[l]{\footnotesize Colorful\\\footnotesize Diffuse Intrinsic}}
& dark\_heavy & 0.145 & 15.825 & \textbf{0.718} & 0.364 \\
& hdr & 0.141 & 15.978 & \textbf{0.753} & 0.322 \\
& highlight\_heavy & 0.141 & 15.859 & 0.748 & 0.324 \\
& low\_light & \textbf{0.124} & \textbf{16.516} & \textbf{0.660} & 0.392 \\
\cmidrule(lr){1-6}

\multirow{4}{*}{\footnotesize DNF-Intrinsic}
& dark\_heavy & \textbf{0.106} & \textbf{18.214} & 0.701 & \textbf{0.200} \\
& hdr & \textbf{0.093} & \textbf{19.365} & 0.740 & \textbf{0.180} \\
& highlight\_heavy & \textbf{0.089} & \textbf{19.534} & \textbf{0.800} & \textbf{0.161} \\
& low\_light & 0.189 & 13.582 & 0.519 & \textbf{0.272} \\
\bottomrule
\end{tabular}

\tabremark{
For shared slice names, source-specific slice scores are first computed within each dataset and then macro-averaged across OpenRooms-FF and InteriorVerse.
The \textit{low\_light} slice is defined only for OpenRooms-FF and is reported as a source-specific stress diagnostic.
Masked MAE is the primary scalar distortion metric; SSIM and LPIPS are required diagnostics for structure preservation and perceptual consistency.
Lower MAE/LPIPS and higher PSNR/SSIM are better.
Best results within each slice are bolded per metric.
}
\end{table}

\FloatBarrier
\subsubsection{Normal Stress-Subset Results}
\label{app:normal_stress_results}

For normal estimation, the stress analysis focuses on whether exemplar-stabilized proprietary outputs remain far behind specialized predictors once illumination becomes more challenging.
Table~\ref{tab:normal_stress_combined} therefore complements the main normal comparison by showing how the directional-accuracy gap behaves under each stress slice.

\begin{table}[!htbp]
\centering
\caption{Combined stress-test normal estimation results under the disclosed multi-exemplar convention-stabilized normal protocol.}
\label{tab:normal_stress_combined}
\compactappendixtable
\begin{tabular}{llcccc}
\toprule
Method & Slice & Median$\downarrow$ & Acc@11.25$\uparrow$ & Acc@22.5$\uparrow$ & Acc@30$\uparrow$ \\
\midrule
\multirow{4}{*}{GPT-Image-1.5}
& dark\_heavy & 42.119 & 0.061 & 0.213 & 0.330 \\
& hdr & 42.338 & 0.058 & 0.205 & 0.328 \\
& highlight\_heavy & 42.967 & 0.055 & 0.201 & 0.321 \\
& low\_light & 37.101 & 0.071 & 0.251 & 0.386 \\
\cmidrule(lr){1-6}

\multirow{4}{*}{\makecell[l]{Doubao\\Seedream 5.0}}
& dark\_heavy & 49.642 & 0.031 & 0.121 & 0.215 \\
& hdr & 50.348 & 0.038 & 0.139 & 0.232 \\
& highlight\_heavy & 49.987 & 0.041 & 0.146 & 0.240 \\
& low\_light & 41.392 & 0.043 & 0.161 & 0.285 \\
\cmidrule(lr){1-6}

\multirow{4}{*}{Lotus}
& dark\_heavy & \textbf{7.904} & 0.742 & 0.837 & 0.865 \\
& hdr & \textbf{6.805} & \textbf{0.789} & 0.872 & 0.894 \\
& highlight\_heavy & \textbf{7.821} & 0.763 & 0.856 & 0.879 \\
& low\_light & \textbf{4.382} & \textbf{0.823} & 0.923 & 0.946 \\
\cmidrule(lr){1-6}

\multirow{4}{*}{StableNormal}
& dark\_heavy & 8.832 & \textbf{0.747} & \textbf{0.855} & \textbf{0.881} \\
& hdr & 7.774 & 0.785 & \textbf{0.885} & \textbf{0.906} \\
& highlight\_heavy & 8.250 & \textbf{0.771} & \textbf{0.868} & \textbf{0.889} \\
& low\_light & 6.938 & 0.791 & \textbf{0.928} & \textbf{0.952} \\
\bottomrule
\end{tabular}

\tabremark{
For proprietary generative systems, scores are averaged over the three fixed exemplar-conditioned runs.
For shared slice names, source-specific slice scores are first computed within each dataset and then macro-averaged across OpenRooms-FF and InteriorVerse.
The \textit{low\_light} slice is defined only for OpenRooms-FF and is not directly comparable to the cross-source stress slices.
Acc@22.5 is the primary directional-accuracy metric.
Lower Median is better; higher Acc@11.25, Acc@22.5, and Acc@30 are better.
Best results within each slice are bolded per metric.
}
\end{table}

\FloatBarrier
\subsubsection{Roughness Stress-Subset Results}
\label{app:roughness_stress_results}

For roughness, the stress subsets are useful mainly for checking whether the scalar--structural split observed on the main axis persists under harder lighting.
Table~\ref{tab:roughness_stress_openrooms} should therefore be read as a slice-wise diagnostic of RMSE versus SSIM behavior rather than as a second roughness leaderboard.

\begingroup
\compactappendixtable
\begin{longtable}{llccccc}
\caption{Stress-test roughness estimation results on OpenRooms-FF under the controlled direct RGB-to-map track.}
\label{tab:roughness_stress_openrooms}\\

\toprule
Method & Slice & RMSE$\downarrow$ & MAE$\downarrow$ & SSIM$\uparrow$ & LPIPS$\downarrow$ & PSNR$\uparrow$ \\
\midrule
\endfirsthead

\caption[]{Stress-test roughness estimation results on OpenRooms-FF under the controlled direct RGB-to-map track. Continued.}\\
\toprule
Method & Slice & RMSE$\downarrow$ & MAE$\downarrow$ & SSIM$\uparrow$ & LPIPS$\downarrow$ & PSNR$\uparrow$ \\
\midrule
\endhead

\midrule
\multicolumn{7}{r}{Continued on next page}\\
\endfoot

\bottomrule
\endlastfoot

\multirow{4}{*}{GPT-Image-2.0}
& dark\_heavy & 0.330 & 0.289 & 0.408 & 0.563 & 10.185 \\
& hdr & 0.289 & \textbf{0.248} & 0.449 & \textbf{0.502} & 11.324 \\
& highlight\_heavy & \textbf{0.289} & \textbf{0.248} & 0.482 & \textbf{0.492} & \textbf{11.180} \\
& low\_light & 0.309 & 0.269 & 0.430 & \textbf{0.511} & 10.793 \\
\midrule

\multirow{4}{*}{GPT-Image-1.5}
& dark\_heavy & 0.377 & 0.328 & 0.359 & 0.614 & 9.079 \\
& hdr & 0.354 & 0.304 & 0.392 & 0.574 & 9.534 \\
& highlight\_heavy & 0.364 & 0.314 & 0.432 & 0.569 & 9.360 \\
& low\_light & 0.367 & 0.316 & 0.366 & 0.574 & 9.273 \\
\midrule

\multirow{4}{*}{Qwen-Image-2.0}
& dark\_heavy & 0.394 & 0.333 & 0.363 & 0.627 & 8.366 \\
& hdr & 0.389 & 0.329 & 0.373 & 0.603 & 8.425 \\
& highlight\_heavy & 0.345 & 0.290 & 0.434 & 0.582 & 9.490 \\
& low\_light & 0.382 & 0.326 & 0.366 & 0.609 & 8.606 \\
\midrule

\multirow{4}{*}{\makecell[l]{Doubao\\Seedream 5.0}}
& dark\_heavy & 0.343 & 0.290 & 0.337 & \textbf{0.548} & 9.730 \\
& hdr & 0.336 & 0.281 & 0.357 & 0.507 & 9.879 \\
& highlight\_heavy & 0.354 & 0.296 & 0.365 & 0.502 & 9.443 \\
& low\_light & 0.334 & 0.278 & 0.355 & 0.513 & 9.882 \\
\midrule

\multirow{4}{*}{\makecell[l]{Intrinsic\\Image Diffusion}}
& dark\_heavy & \textbf{0.321} & \textbf{0.282} & \textbf{0.492} & 0.592 & \textbf{10.610} \\
& hdr & \textbf{0.286} & 0.251 & \textbf{0.517} & 0.534 & \textbf{11.546} \\
& highlight\_heavy & 0.333 & 0.292 & \textbf{0.536} & 0.563 & 10.181 \\
& low\_light & \textbf{0.294} & \textbf{0.253} & \textbf{0.509} & 0.547 & \textbf{11.132} \\
\midrule

\multirow{4}{*}{DNF-Intrinsic}
& dark\_heavy & 0.403 & 0.351 & 0.416 & 0.627 & 8.485 \\
& hdr & 0.360 & 0.314 & 0.445 & 0.573 & 9.479 \\
& highlight\_heavy & 0.340 & 0.285 & 0.514 & 0.558 & 9.971 \\
& low\_light & 0.420 & 0.368 & 0.391 & 0.619 & 8.082 \\

\end{longtable}
\vspace{-2em}
\tabremark{
Scores are reported as per-image averages within each OpenRooms-FF slice, with slice assignment determined by \texttt{openroomsff\_stresstest.csv}.
The evaluated slices are \textit{dark\_heavy}, \textit{hdr}, \textit{highlight\_heavy}, and \textit{low\_light}.
RMSE is the primary scalar error metric for roughness; MAE, SSIM, LPIPS, and PSNR are reported as auxiliary diagnostics.
Lower RMSE/MAE/LPIPS and higher SSIM/PSNR are better.
Best results within each slice are bolded per metric.
}
\endgroup

\subsubsection{Depth Stress-Subset Results}
\label{app:depth_stress_results}

Depth stress results are reported slice by slice to expose whether scalar relative-depth recovery and boundary recovery remain decoupled under adverse illumination.
Table~\ref{tab:depth_stress_slices} makes this comparison explicit by keeping the same affine-invariant evaluator and varying only the photometric slice.

\begingroup
\compactappendixtable
\begin{longtable}{llcccccc}
\caption{Stress-test relative-depth results across annotated challenging slices. For \textit{dark\_heavy}, \textit{hdr}, and \textit{highlight\_heavy}, scores are source-balanced macro-averages over OpenRooms-FF and InteriorVerse. The \textit{low\_light} slice is reported on OpenRooms-FF only because the InteriorVerse stress subset does not contain that slice.}
\label{tab:depth_stress_slices}\\

\toprule
Method & Slice & AbsRel-AI$\downarrow$ & RMSE-AI$\downarrow$ & MAE-AI$\downarrow$ & $\delta_1$-AI$\uparrow$ & $\delta_2$-AI$\uparrow$ & Boundary F1$\uparrow$ \\
\midrule
\endfirsthead

\caption[]{Stress-test relative-depth results across annotated challenging slices. Continued.}\\
\toprule
Method & Slice & AbsRel-AI$\downarrow$ & RMSE-AI$\downarrow$ & MAE-AI$\downarrow$ & $\delta_1$-AI$\uparrow$ & $\delta_2$-AI$\uparrow$ & Boundary F1$\uparrow$ \\
\midrule
\endhead

\midrule
\multicolumn{8}{r}{Continued on next page}\\
\endfoot

\bottomrule
\endlastfoot

\multirow{4}{*}{GPT-Image-2.0}
& dark\_heavy & 0.188 & 0.385 & 0.299 & 0.824 & 0.944 & 0.266 \\
& hdr & 0.154 & 0.379 & 0.293 & 0.824 & 0.947 & 0.280 \\
& highlight\_heavy & 0.158 & 0.372 & 0.286 & 0.834 & 0.947 & 0.256 \\
& low\_light (OR-FF) & 0.095 & 0.342 & 0.267 & 0.916 & 0.983 & 0.340 \\
\midrule

\multirow{4}{*}{GPT-Image-1.5}
& dark\_heavy & 0.251 & 0.524 & 0.403 & 0.733 & 0.919 & 0.030 \\
& hdr & 0.190 & 0.512 & 0.391 & 0.750 & 0.928 & 0.031 \\
& highlight\_heavy & 0.196 & 0.484 & 0.368 & 0.766 & 0.933 & 0.025 \\
& low\_light (OR-FF) & 0.131 & 0.507 & 0.387 & 0.840 & 0.967 & 0.039 \\
\midrule

\multirow{4}{*}{Qwen-Image-2.0}
& dark\_heavy & 0.163 & 0.363 & 0.281 & 0.855 & 0.956 & 0.193 \\
& hdr & 0.134 & 0.360 & 0.278 & 0.856 & 0.960 & 0.194 \\
& highlight\_heavy & 0.146 & 0.356 & 0.273 & 0.853 & 0.958 & 0.180 \\
& low\_light (OR-FF) & 0.097 & 0.366 & 0.286 & 0.913 & 0.979 & 0.220 \\
\midrule

\multirow{4}{*}{\makecell[l]{Doubao\\Seedream 5.0}}
& dark\_heavy & 0.181 & 0.462 & 0.353 & 0.782 & 0.946 & 0.131 \\
& hdr & 0.161 & 0.452 & 0.346 & 0.795 & 0.951 & 0.144 \\
& highlight\_heavy & 0.165 & 0.449 & 0.344 & 0.792 & 0.949 & 0.133 \\
& low\_light (OR-FF) & 0.123 & 0.474 & 0.365 & 0.864 & 0.976 & 0.163 \\
\midrule

\multirow{4}{*}{Depth Anything 3}
& dark\_heavy & \textbf{0.047} & \textbf{0.130} & \textbf{0.071} & 0.973 & 0.990 & 0.247 \\
& hdr & \textbf{0.034} & \textbf{0.121} & \textbf{0.066} & \textbf{0.981} & 0.993 & 0.257 \\
& highlight\_heavy & \textbf{0.038} & \textbf{0.122} & \textbf{0.066} & 0.978 & 0.992 & 0.231 \\
& low\_light (OR-FF) & \textbf{0.012} & \textbf{0.086} & \textbf{0.035} & \textbf{0.996} & \textbf{1.000} & 0.314 \\
\midrule

\multirow{4}{*}{Lotus Depth}
& dark\_heavy & 0.052 & 0.151 & 0.096 & \textbf{0.975} & \textbf{0.991} & \textbf{0.325} \\
& hdr & 0.045 & 0.147 & 0.094 & 0.978 & \textbf{0.995} & \textbf{0.340} \\
& highlight\_heavy & 0.048 & 0.143 & 0.091 & \textbf{0.978} & \textbf{0.994} & \textbf{0.311} \\
& low\_light (OR-FF) & 0.031 & 0.131 & 0.086 & 0.992 & 0.999 & \textbf{0.384} \\

\end{longtable}
\vspace{-2em}
\tabremark{
For \textit{dark\_heavy}, \textit{hdr}, and \textit{highlight\_heavy}, we first average per-image metrics within each source and slice, then macro-average the two source-wise means to obtain the reported source-balanced score.
The corresponding support sizes are 45 OpenRooms-FF scenes for each slice, and 38/36/35 InteriorVerse scenes for \textit{dark\_heavy}/\textit{hdr}/\textit{highlight\_heavy}, respectively.
The \textit{low\_light} slice is available only in OpenRooms-FF; its reported values are therefore OpenRooms-FF-only per-image means.
Lower AbsRel-AI, RMSE-AI, and MAE-AI are better; higher $\delta_1$-AI, $\delta_2$-AI, and Boundary F1 are better.
Boldface marks the best result within each slice and metric.
}
\endgroup

\subsubsection{Metallic Stress-Subset Results}
\label{app:metallic_stress_results}

For metallic, the stress analysis is primarily a robustness check on sparse continuous-map fidelity under lighting variation in the companion source.
Table~\ref{tab:metallic_stress_companion} therefore emphasizes metric behavior across lighting templates rather than calibrated metallic-region detection.

\begingroup
\compactappendixtable
\setlength{\tabcolsep}{2.8pt}
\renewcommand{\arraystretch}{0.88}
\begin{table}[!p]
\centering
\caption{Stress-test metallic continuous-map results on the companion dataset under the controlled direct RGB-to-map track.}
\label{tab:metallic_stress_companion}
\begin{tabular}{llcccc}
\toprule
Method & Slice & MAE$\downarrow$ & PSNR$\uparrow$ & SSIM$\uparrow$ & LPIPS$\downarrow$ \\
\midrule

\multirow{6}{*}{GPT-Image-2.0}
& normal\_light     & \textbf{0.036} & \textbf{20.272} & \textbf{0.960} & \textbf{0.111} \\
& low\_light        & \textbf{0.036} & \textbf{19.626} & \textbf{0.945} & \textbf{0.116} \\
& hdr               & \textbf{0.036} & \textbf{20.336} & \textbf{0.955} & \textbf{0.114} \\
& dark\_heavy       & \textbf{0.036} & \textbf{19.872} & \textbf{0.957} & \textbf{0.112} \\
& highlight\_heavy  & \textbf{0.036} & \textbf{20.351} & \textbf{0.958} & \textbf{0.116} \\
& mix\_temperature  & \textbf{0.035} & \textbf{20.785} & \textbf{0.954} & \textbf{0.112} \\
\midrule

\multirow{6}{*}{GPT-Image-1.5}
& normal\_light     & 0.041 & 18.234 & 0.890 & 0.174 \\
& low\_light        & 0.040 & 18.917 & 0.894 & 0.148 \\
& hdr               & 0.041 & 18.474 & 0.886 & 0.159 \\
& dark\_heavy       & 0.040 & 18.948 & 0.884 & 0.157 \\
& highlight\_heavy  & 0.040 & 18.189 & 0.890 & 0.163 \\
& mix\_temperature  & 0.041 & 18.548 & 0.872 & 0.169 \\
\midrule

\multirow{6}{*}{WAN2.7-Image}
& normal\_light     & 0.109 & 13.614 & 0.510 & 0.482 \\
& low\_light        & 0.103 & 14.943 & 0.511 & 0.466 \\
& hdr               & 0.127 & 13.116 & 0.505 & 0.489 \\
& dark\_heavy       & 0.111 & 14.197 & 0.507 & 0.473 \\
& highlight\_heavy  & 0.109 & 15.295 & 0.519 & 0.468 \\
& mix\_temperature  & 0.108 & 13.771 & 0.501 & 0.509 \\
\midrule

\multirow{6}{*}{\makecell[l]{Doubao\\Seedream 5.0}}
& normal\_light     & 0.057 & 18.600 & 0.523 & 0.339 \\
& low\_light        & 0.056 & 19.232 & 0.424 & 0.282 \\
& hdr               & 0.058 & 18.393 & 0.457 & 0.298 \\
& dark\_heavy       & 0.056 & 18.717 & 0.490 & 0.310 \\
& highlight\_heavy  & 0.057 & 18.083 & 0.599 & 0.364 \\
& mix\_temperature  & 0.055 & 18.493 & 0.541 & 0.342 \\
\midrule

\multirow{6}{*}{MVInverse}
& normal\_light     & 0.178 & 12.507 & 0.120 & 0.460 \\
& low\_light        & 0.191 & 11.948 & 0.144 & 0.476 \\
& hdr               & 0.200 & 11.700 & 0.125 & 0.488 \\
& dark\_heavy       & 0.186 & 12.213 & 0.140 & 0.467 \\
& highlight\_heavy  & 0.167 & 12.904 & 0.132 & 0.439 \\
& mix\_temperature  & 0.181 & 12.419 & 0.123 & 0.465 \\
\midrule

\multirow{6}{*}{DNF-Intrinsic}
& normal\_light     & 0.150 & 14.034 & 0.210 & 0.475 \\
& low\_light        & 0.121 & 15.193 & 0.277 & 0.470 \\
& hdr               & 0.132 & 14.507 & 0.264 & 0.460 \\
& dark\_heavy       & 0.123 & 15.351 & 0.281 & 0.453 \\
& highlight\_heavy  & 0.177 & 13.247 & 0.222 & 0.515 \\
& mix\_temperature  & 0.170 & 13.233 & 0.240 & 0.513 \\
\bottomrule
\end{tabular}
\tabremark{
Scores are reported separately for the six released stress-test lighting slices:
\textit{normal\_light}, \textit{low\_light}, \textit{hdr}, \textit{dark\_heavy}, \textit{highlight\_heavy}, and \textit{mix\_temperature}.
Following the metallic continuous-map evaluation protocol, predictions are clipped to \([0,1]\), evaluated within \texttt{MetallicEvalMask}, and are not normalized per image.
MAE is the primary continuous scalar distortion metric, while PSNR, SSIM, and LPIPS are complementary full-map diagnostics.
Lower MAE/LPIPS and higher PSNR/SSIM are better.
Best results within each slice are bolded per metric.
}
\end{table}

\endgroup

\subsection{Main-Axis Uncertainty Analysis}
\label{app:main_axis_uncertainty}

This section reports scene-cluster bootstrap uncertainty analyses for primary main-axis metrics where complete scene-level grouping is available.
For multi-source targets, bootstrap resampling is performed separately within each source and then macro-averaged across sources.
For single-source targets, scenes are resampled within the retained source subset.
These intervals quantify scene-level variation in the reported point estimates and are not used to redefine the primary metric ranking.

\subsubsection{Depth Primary-Metric Uncertainty}
\label{app:depth_uncertainty}

We begin the uncertainty analysis with the primary depth metric because the main depth comparison is source-balanced and scene-cluster variability is non-negligible across the two retained sources.
Table~\ref{tab:depth_absrel_bootstrap_ci} quantifies how stable the main AbsRel-AI ordering remains under scene-level resampling.

\begin{table}[!htbp]
\centering
\caption{Source-balanced scene-cluster bootstrap confidence intervals for the primary depth metric on the main-axis subset.}
\label{tab:depth_absrel_bootstrap_ci}
\footnotesize
\setlength{\tabcolsep}{7pt}
\renewcommand{\arraystretch}{0.95}
\begin{tabular}{lcc}
\toprule
Method & AbsRel-AI point estimate$\downarrow$ & 95\% CI\\
\midrule
GPT-Image-1.5 & 0.2339 & [0.1867, 0.3156] \\
GPT-Image-2.0 & 0.1682 & [0.1441, 0.2076] \\
WAN2.7-Image & 0.1657 & [0.1309, 0.2316] \\
Doubao Seedream 5.0 & 0.1713 & [0.1586, 0.1870] \\
Depth Anything 3 & \textbf{0.0350} & [0.0310, 0.0399] \\
Lotus Depth & 0.0479 & [0.0431, 0.0541] \\
\bottomrule
\end{tabular}

\tabremark{
We report 95\% confidence intervals for the primary depth metric using source-balanced scene-cluster bootstrap with 1,000 resamples.
Within each bootstrap sample, scenes are resampled with replacement separately inside OpenRooms-FF and InteriorVerse, all images from sampled scenes are retained, source-wise AbsRel-AI means are computed, and the final score is the macro-average across the two sources.
The evaluation support contains 1,125 OpenRooms-FF images from 48 scenes and 2,109 InteriorVerse images from 167 scenes.
Lower AbsRel-AI is better.
Boldface marks the best point estimate only.
}
\end{table}

\subsubsection{Normal Primary-Metric Uncertainty}
\label{app:normal_uncertainty}

For normal estimation, uncertainty is reported on Acc@22.5 because it is the official primary metric and because the proprietary runs additionally involve exemplar averaging.
Table~\ref{tab:normal_acc225_bootstrap_ci} therefore complements the main normal table by showing scene-level uncertainty for both per-exemplar rows and the averaged proprietary aggregate.

\begin{table}[!htbp]
\centering
\caption{Source-balanced scene-cluster bootstrap confidence intervals for the primary normal metric on the main-axis subset.}
\label{tab:normal_acc225_bootstrap_ci}
\footnotesize
\setlength{\tabcolsep}{6pt}
\renewcommand{\arraystretch}{0.95}
\begin{tabular}{lcc}
\toprule
Method / aggregate row & Acc@22.5 point estimate$\uparrow$ & 95\% CI \\
\midrule
GPT-Image-1.5 E1 & 0.1885 & [0.1820, 0.1948] \\
GPT-Image-1.5 E2 & 0.2189 & [0.2118, 0.2257] \\
GPT-Image-1.5 E3 & 0.2049 & [0.1986, 0.2109] \\
GPT-Image-1.5 Avg & 0.2041 & [0.2002, 0.2079] \\
\midrule
Doubao Seedream 5.0 E1 & 0.1395 & [0.1317, 0.1484] \\
Doubao Seedream 5.0 E2 & 0.1177 & [0.1098, 0.1258] \\
Doubao Seedream 5.0 E3 & 0.1403 & [0.1336, 0.1485] \\
Doubao Seedream 5.0 Avg & 0.1325 & [0.1283, 0.1371] \\
\midrule
Lotus & 0.8626 & [0.8517, 0.8728] \\
StableNormal & \textbf{0.8758} & [0.8660, 0.8851] \\
\bottomrule
\end{tabular}

\tabremark{
We report percentile 95\% confidence intervals for the primary normal metric using source-balanced scene-cluster bootstrap with 1,000 resamples.
Within each bootstrap sample, scenes are resampled with replacement separately inside OpenRooms-FF and InteriorVerse, all images from sampled scenes are retained, source-wise Acc@22.5 means are computed, and the final score is the macro-average across the two sources.
For proprietary multi-exemplar entries, the Avg row is computed within each bootstrap replicate as the arithmetic mean of the three exemplar-conditioned source-balanced scores.
The evaluation support contains 1,125 OpenRooms-FF images from 48 scenes and 2,109 InteriorVerse images from 167 scenes.
Higher Acc@22.5 is better.
Boldface marks the best point estimate only.
}
\end{table}

\subsubsection{Albedo Primary-Metric Uncertainty}
\label{app:albedo_uncertainty}

For albedo, we focus on MAE because it is the primary scalar distortion axis used in the main benchmark comparison.
Table~\ref{tab:albedo_mae_bootstrap_ci} shows that the source-balanced MAE ordering is not driven by a small number of retained scenes.

\begin{table}[!htbp]
\centering
\caption{Source-balanced scene-cluster bootstrap confidence intervals for the primary albedo metric on the main-axis subset.}
\label{tab:albedo_mae_bootstrap_ci}
\footnotesize
\setlength{\tabcolsep}{7pt}
\renewcommand{\arraystretch}{0.95}
\begin{tabular}{lcc}
\toprule
Method & MAE point estimate$\downarrow$ & 95\% CI \\
\midrule
GPT-Image-1.5 & 0.2038 & [0.1964, 0.2114] \\
GPT-Image-2.0 & 0.1743 & [0.1705, 0.1778] \\
WAN2.7-Image & 0.2197 & [0.2149, 0.2249] \\
Doubao Seedream 5.0 & 0.1832 & [0.1785, 0.1879] \\
Colorful Diffuse Intrinsic & 0.1379 & [0.1341, 0.1415] \\
DNF-Intrinsic & \textbf{0.0942} & [0.0905, 0.0978] \\
\bottomrule
\end{tabular}

\tabremark{
We report percentile 95\% confidence intervals for the primary albedo metric using source-balanced scene-cluster bootstrap with 1,000 resamples.
Within each bootstrap sample, scenes are resampled with replacement separately inside OpenRooms-FF and InteriorVerse, all images from sampled scenes are retained, source-wise MAE means are computed, and the final score is the macro-average across the two sources.
The evaluation support contains 1,125 OpenRooms-FF images from 48 scenes and 2,109 InteriorVerse images from 167 scenes.
Lower MAE is better.
Boldface marks the best point estimate only.
}
\end{table}

\subsubsection{Roughness Main-Axis Uncertainty Analysis}
\label{app:roughness_uncertainty}

Roughness uncertainty is reported in a paired form because the main interpretive point is the scalar--structural split between the leading proprietary editor and the strongest structural baseline.
Table~\ref{tab:roughness_mainaxis_ci} therefore reports uncertainty not only for each model separately, but also for their scene-level paired differences.

\begin{table}[!htbp]
\centering
\caption{Scene-cluster paired-bootstrap confidence intervals for the roughness scalar--structural split between GPT-Image-2.0 and Intrinsic Image Diffusion on OpenRooms-FF.}
\label{tab:roughness_mainaxis_ci}
\footnotesize
\setlength{\tabcolsep}{6pt}
\renewcommand{\arraystretch}{0.95}
\begin{tabular}{lccc}
\toprule
Entry & Metric & Point estimate & 95\% CI \\
\midrule
GPT-Image-2.0 & RMSE$\downarrow$ & 0.323 & [0.312, 0.336] \\
Intrinsic Image Diffusion & RMSE$\downarrow$ & 0.346 & [0.332, 0.361] \\
\midrule
GPT-Image-2.0 & SSIM$\uparrow$ & 0.444 & [0.423, 0.464] \\
Intrinsic Image Diffusion & SSIM$\uparrow$ & 0.513 & [0.491, 0.533] \\
\midrule
$\Delta$(GPT-Image-2.0 $-$ Intrinsic Image Diffusion) & RMSE$\downarrow$ & $-0.023$ & [$-0.038$, $-0.008$] \\
$\Delta$(GPT-Image-2.0 $-$ Intrinsic Image Diffusion) & SSIM$\uparrow$ & $-0.069$ & [$-0.080$, $-0.055$] \\
\bottomrule
\end{tabular}

\tabremark{
We report percentile 95\% confidence intervals from scene-cluster paired bootstrap with 1,000 resamples on the OpenRooms-FF main-axis roughness set.
Each bootstrap replicate resamples scenes with replacement and retains all evaluated images from each sampled scene.
The paired differences are computed as GPT-Image-2.0 minus Intrinsic Image Diffusion.
For RMSE, lower is better, so a negative $\Delta$ favors GPT-Image-2.0.
For SSIM, higher is better, so a negative $\Delta$ favors Intrinsic Image Diffusion.
The paired intervals exclude zero for both metrics, indicating that the scalar--structural split is stable at the scene level under this evaluation protocol.
}
\end{table}

\subsubsection{Metallic Primary-Metric Uncertainty}
\label{app:metallic_uncertainty}

For metallic, the uncertainty table is included mainly to temper over-interpretation of small MAE gaps among the top proprietary systems.
Table~\ref{tab:metallic_mae_bootstrap_ci} should therefore be read together with the main metallic table as a stability check on the sparse-map MAE ranking.

\begin{table}[!htbp]
\centering
\caption{Scene-cluster bootstrap confidence intervals for the primary metallic metric on the companion main-axis subset.}
\label{tab:metallic_mae_bootstrap_ci}
\footnotesize
\setlength{\tabcolsep}{7pt}
\renewcommand{\arraystretch}{0.95}
\begin{tabular}{lcc}
\toprule
Method & MAE point estimate$\downarrow$ & 95\% CI \\
\midrule
GPT-Image-1.5 & 0.0554 & [0.0371, 0.0772] \\
GPT-Image-2.0 & \textbf{0.0514} & [0.0324, 0.0738] \\
WAN2.7-Image & 0.1204 & [0.1054, 0.1368] \\
Doubao Seedream 5.0 & 0.0728 & [0.0555, 0.0935] \\
MVInverse & 0.2110 & [0.1951, 0.2267] \\
DNF-Intrinsic & 0.1500 & [0.1262, 0.1758] \\
\bottomrule
\end{tabular}

\tabremark{
We report percentile 95\% confidence intervals for the primary metallic metric using scene-cluster bootstrap with 1,000 resamples on the companion main-axis subset.
Within each bootstrap sample, scenes are resampled with replacement and all images from sampled scenes are retained.
Scene clusters are defined at the room-seed level, yielding 960 evaluated images across 40 scene clusters.
Lower MAE is better.
Boldface marks the best point estimate only.
}
\end{table}

\FloatBarrier

\subsection{Diagnostic and Ablation Results}
\label{app:diagnostic_results}

\subsubsection{Normal Per-Exemplar Main-Axis Results}
\label{app:normal_exemplar_diagnostics}

Table~\ref{tab:normal_mainaxis_per_exemplar} reports the per-exemplar normal results corresponding to the averaged proprietary-editor entries in the main normal table.
The three RGB--normal exemplar pairs are fixed before evaluation and are shared across all test images and proprietary systems.
The Avg row is the arithmetic mean over the three exemplar-conditioned runs and is the value reported in the main text.
These per-exemplar rows are included to expose residual sensitivity to exemplar choice, not to select a best exemplar after observing test performance.

\begin{table*}[!htbp]
\centering
\caption{Per-exemplar source-balanced normal results under the fixed three-exemplar normal setting.}
\label{tab:normal_mainaxis_per_exemplar}
\compactappendixtable
\begin{tabular}{llccccc}
\toprule
Method & Exemplar & Mean$\downarrow$ & Median$\downarrow$ & Acc@11.25$\uparrow$ & Acc@22.5$\uparrow$ & Acc@30$\uparrow$ \\
\midrule

\multirow{4}{*}{\makecell[c]{GPT-Image-1.5~\cite{openai2025gpt_image15}}}
& E1  & 45.700 & 43.658 & 0.049 & 0.188 & 0.310 \\
& E2  & 43.921 & 41.581 & 0.063 & 0.219 & 0.342 \\
& E3  & 44.787 & 42.312 & 0.058 & 0.205 & 0.324 \\
& Avg & 44.803 & 42.517 & 0.057 & 0.204 & 0.325 \\
\midrule
\addlinespace[0.15em]

\multirow{4}{*}{\makecell[c]{Doubao\\Seedream 5.0~\cite{seedream5_lite}}}
& E1  & 56.492 & 50.394 & 0.039 & 0.139 & 0.234 \\
& E2  & 57.213 & 51.257 & 0.029 & 0.118 & 0.211 \\
& E3  & 55.025 & 48.820 & 0.037 & 0.140 & 0.237 \\
& Avg & 56.243 & 50.157 & 0.035 & 0.132 & 0.227 \\
\midrule
\addlinespace[0.15em]

WAN2.7-Image~\cite{wan27_image}
& -- & \multicolumn{5}{c}{\footnotesize Non-scoreable normal-representation failure; see Figure~\ref{fig:wan_normal_failure_cases}} \\

\midrule

Lotus~\cite{Lotus}
& -- & 11.679 & \textbf{7.305} & 0.770 & 0.863 & 0.888 \\

StableNormal~\cite{stablenormal}
& -- & \textbf{11.216} & 7.699 & \textbf{0.777} & \textbf{0.876} & \textbf{0.900} \\

\bottomrule
\end{tabular}

\tabremark{
\editormark\ denotes proprietary image editors evaluated under the fixed three-exemplar normal setting;  denotes specialized RGB-to-normal predictors evaluated in native inference mode.
Scores are source-balanced macro-averages over OpenRooms-FF and InteriorVerse.
E1--E3 denote the three fixed RGB--normal exemplar-conditioned runs.
Avg is their arithmetic mean and is the value reported in the main normal table.
Acc@22.5 is the primary metric.
Per-exemplar rows expose residual sensitivity to exemplar choice and are not used to select a best exemplar.
Non-scoreable representation failures are not assigned artificial angular scores because angular metrics assume a valid dense normal-map representation.
Boldface marks the best score among scoreable entries for each metric.
}
\end{table*}

\subsection{Diagnostic-Subset Sampling Protocol}
\label{app:diagnostic_subset_sampling}

Protocol diagnostics and ablation studies are evaluated on target-specific diagnostic subsets rather than on the full retained benchmark split.
This design keeps diagnostic runs computationally feasible while preserving coverage over source, illumination, stress condition, and scene category.
These subsets are used only for protocol-sensitivity analysis and are not used to compute the official main-axis rankings.

For depth, normal, and albedo, we sample 96 images from the corresponding diagnostic support.
The subset contains four parts.
For InteriorVerse main-axis images, we stratify by \texttt{illumination\_level} into \textit{bright} and \textit{normal} bins and sample 12 images from each bin.
For InteriorVerse stress images, we stratify by \textit{hdr}, \textit{highlight\_heavy}, and \textit{dark\_heavy}, and sample 8 images from each bin.
For OpenRooms-FF main-axis images, we stratify by \texttt{illumination\_level} into \textit{bright}, \textit{normal}, and \textit{dim} bins and sample 8 images from each bin.
For OpenRooms-FF stress images, we stratify by \textit{hdr}, \textit{highlight\_heavy}, \textit{dark\_heavy}, and \textit{low\_light}, and sample 6 images from each bin.
Whenever possible, samples are balanced across scene categories within each stratum.

For roughness, diagnostics use OpenRooms-FF only, because OpenRooms-FF is the official roughness source.
We sample 96 images.
The main-axis portion is stratified by \texttt{illumination\_level} into \textit{bright}, \textit{normal}, and \textit{dim} bins, with 16 images sampled from each bin.
The stress portion is stratified by \textit{hdr}, \textit{highlight\_heavy}, \textit{dark\_heavy}, and \textit{low\_light}, with 12 images sampled from each bin.
Scene-category balance is enforced where the available samples allow it.

For metallic, diagnostics use only the companion source.
For the metallic diagnostic subset, we first select two scenes from each of the five scene categories.
For the main-axis portion, we select six primary views from each selected scene.
This yields \(6 \times 2 \times 5 = 60\) main-axis diagnostic images.
For the stress portion, we select one primary view from each selected scene and render it under six lighting templates.
This yields \(1 \times 6 \times 2 \times 5 = 60\) stress diagnostic images.
The metallic diagnostic subset therefore contains 120 images in total.
These metallic diagnostic samples are separate from the official companion main-axis and stress-axis evaluation splits.

\subsubsection{Albedo Protocol Diagnostics}

The official albedo protocol was motivated by a pilot failure mode rather than by diagnostic-set metric selection. 
In preliminary trials, proprietary image-editing systems frequently treated strong highlights, saturated regions, cast shadows, and other illumination-dependent appearances as intrinsic surface color, leading to albedo maps that were visually plausible but illumination-confounded. 
We therefore define A0 as an illumination-aware structure-preserving protocol: a query-image-only auxiliary analysis first identifies illumination-confounded regions from the input RGB image, and a fixed transformation prompt then asks the system to suppress illumination-dependent appearance while preserving scene geometry, object boundaries, and material-region organization. 
The auxiliary analysis uses no ground-truth maps, benchmark labels, external references, segmentation masks, or multi-view information; it only reformulates evidence already present in the query image. This official A0 setting was fixed before full retained-split scoring and was not chosen by selecting the best diagnostic-set or test-set scalar metric.

Table~\ref{tab:albedo_ablation_main} reports protocol-sensitivity diagnostics. 
A0 is the official illumination-aware protocol used in the main benchmark. 
A1 removes the auxiliary analysis while keeping the full structure-preserving prompt. 
A2 weakens the prompt constraints, and A3 uses a minimal prompt. 
B0 adds a fixed external exemplar pair and is therefore outside the official query-image-only albedo access setting.

The results show that weaker prompting can improve some aggregate scalar scores on this diagnostic subset. 
In particular, A2 slightly improves MAE, PSNR, and SSIM over A0, and A3 achieves the lowest MAE and highest PSNR. 
However, these variants do not instantiate the illumination-aware protocol that motivates the official albedo setting. 
A0 is designed to test whether a proprietary image editor can use only query-image-derived analysis to identify illumination-confounded regions and produce a structure-preserving intrinsic albedo map. 
A2 and A3 instead diagnose how performance changes when these constraints are weakened or removed.

These diagnostics are intended to expose protocol sensitivity, not to perform protocol selection.
In particular, lower MAE under a weakened prompt does not by itself redefine the official albedo task, because the official setting also requires illumination-aware decomposition and structure-preserving material-region organization.
MAE remains the scalar distortion axis within a fixed access setting, but it is not used to choose the access setting itself.
The diagnostic results therefore expose a scalar--structural and protocol-validity tradeoff: weaker prompts may reduce average pixel error, while the official setting retains the illumination-aware constraints required by the benchmark question.

\begin{table}[!htbp]
  \centering
  \small
  \setlength{\tabcolsep}{6pt}
  \renewcommand{\arraystretch}{0.98}
  \caption{Illumination-aware albedo protocol diagnostics on the merged evaluation set.}
  \label{tab:albedo_ablation_main}
  \begin{tabular}{lcccc}
    \toprule
    Setting & MAE$\downarrow$ & PSNR$\uparrow$ & SSIM$\uparrow$ & LPIPS$\downarrow$ \\
    \midrule
    A0: analysis + structure-preserving prompt & 0.1665 & 14.0757 & 0.6407 & 0.3243 \\
    A1: structure-preserving prompt only       & 0.1680 & 14.0558 & 0.6330 & 0.3365 \\
    A2: weakened prompt only                   & 0.1656 & 14.2763 & 0.6448 & 0.3350 \\
    A3: minimal prompt only                    & 0.1561 & 14.7119 & 0.6110 & 0.3600 \\
    \midrule
    B0: minimal prompt + fixed exemplar        & 0.1565 & 14.6472 & 0.6462 & 0.3404 \\
    \bottomrule
  \end{tabular}

    \tabremark{
    A0 is the official illumination-aware structure-preserving albedo protocol and is fixed before benchmark scoring.
    A1--A3 weaken or remove this elicitation design, while B0 adds an external exemplar and is outside the query-image-only setting.
    The table diagnoses protocol sensitivity over 96 images pooled from \texttt{mainaxis} and \texttt{stresstest}; it is not used for post-hoc prompt selection, and diagnostic optima are not bolded.
    }
    \end{table}

\subsubsection{Normal Exemplar Diagnostics}

We next examine the role of exemplar conditioning and prompt constraint strength under the fixed three-exemplar normal protocol.

\begin{table}[!htbp]
  \centering
  \small
  
  \setlength{\tabcolsep}{8pt} 
  \renewcommand{\arraystretch}{1.1}
  
  \caption{Normal elicitation diagnostics on the merged evaluation set.}
  \label{tab:normal_ablation_overall}
  
  \begin{tabular}{@{}lccc@{}}
    \toprule
    Setting & Mean Angular $\downarrow$ & Acc@22.5 $\uparrow$ & Acc@30 $\uparrow$ \\
    \midrule
    A0: full prompt + three-exemplar aggregate & 57.00 & 0.1181 & 0.2128 \\
    A1: full prompt only & 71.34 & 0.0672 & 0.1105 \\
    A2: minimal prompt + three-exemplar aggregate & 61.40 & 0.0806 & 0.1469 \\
    A3: minimal prompt only & 71.77 & 0.0777 & 0.1329 \\
    \bottomrule
  \end{tabular}

\tabremark{
This diagnostic is computed on the merged protocol-sensitivity evaluation set.
A0 and A2 use the fixed three-exemplar aggregate, while A1 and A3 remove exemplar conditioning.
Mean Angular is reported as an error diagnostic; Acc@22.5 is the primary threshold metric for local directional correctness.
This table diagnoses access-setting sensitivity and should not be read as a main benchmark ranking.
}
\end{table}

Table~\ref{tab:normal_ablation_overall} isolates the effect of exemplar conditioning and prompt constraint strength under the revised normal protocol. 
Comparing A0 with A1 shows whether the fixed exemplar set improves convention-consistent normal-map elicitation over prompt-only generation. 
Comparing A2 with A3 tests the same effect under a weaker prompt. 
The key diagnostic question is therefore no longer which exemplar pair should be selected, but whether the pre-specified exemplar set provides a stable convention cue for proprietary generative systems.

Because E1--E3 are all included in the main normal protocol, per-exemplar variability is reported in the main normal results rather than treated as a separate post-hoc ablation. 
This avoids retroactively choosing a single best exemplar pair and makes the exemplar sensitivity visible in the official normal comparison itself. 
Accordingly, the diagnostic table is restricted to access-setting changes such as removing exemplars or weakening the prompt.

The results show that exemplar conditioning substantially improves the full-prompt setting, reducing Mean Angular from 71.34 to 57.00 and increasing Acc@22.5 from 0.0672 to 0.1181. 
Under the minimal-prompt setting, exemplar conditioning also reduces Mean Angular, although its gain on threshold accuracy is smaller, suggesting that exemplar cues and prompt constraints are complementary.

\FloatBarrier
\subsubsection{Qualitative Normal-Representation Failure Cases}
\label{app:normal_failure_cases}

In addition to the officially scored systems reported in the main normal table, we document representative WAN2.7-Image outputs under the evaluated normal-map access setting.
These examples are not used as additional quantitative results; rather, they illustrate a representation-level output-validity failure mode for normal-map elicitation.

\begin{figure}[!htbp]
    \centering
    \setlength{\tabcolsep}{2pt}
    \renewcommand{\arraystretch}{0.6}
    \begin{tabular}{@{}ccc@{}}
        \textbf{Input RGB} & \textbf{WAN2.7-Image Output} & \textbf{Ground-Truth Normal} \\

        \includegraphics[width=0.20\linewidth]{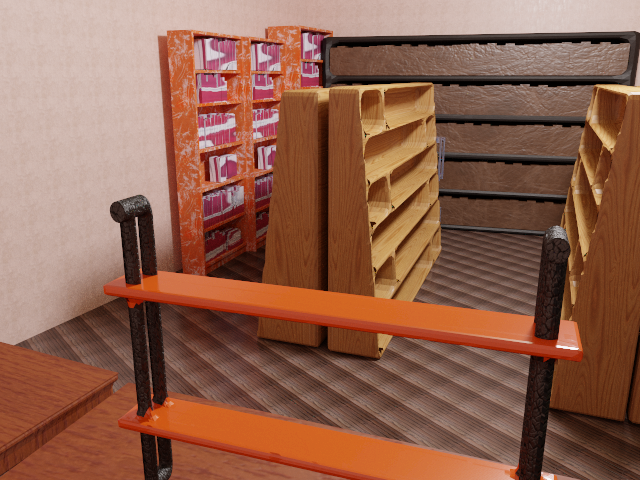} &
        \includegraphics[width=0.20\linewidth]{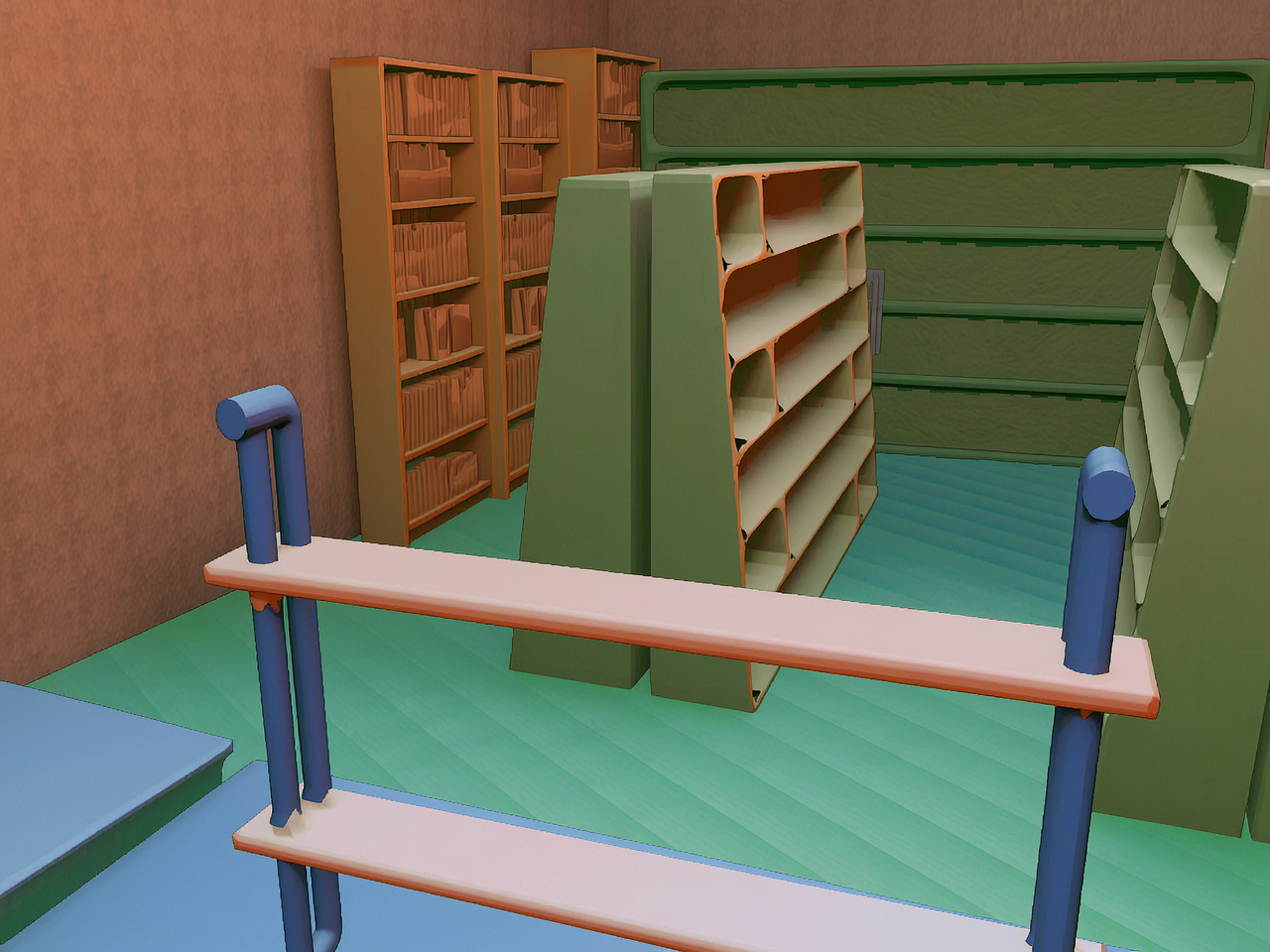} &
        \includegraphics[width=0.20\linewidth]{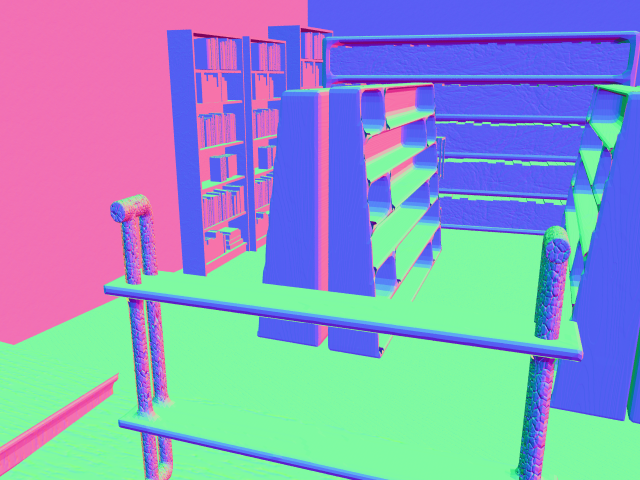} \\

        \includegraphics[width=0.20\linewidth]{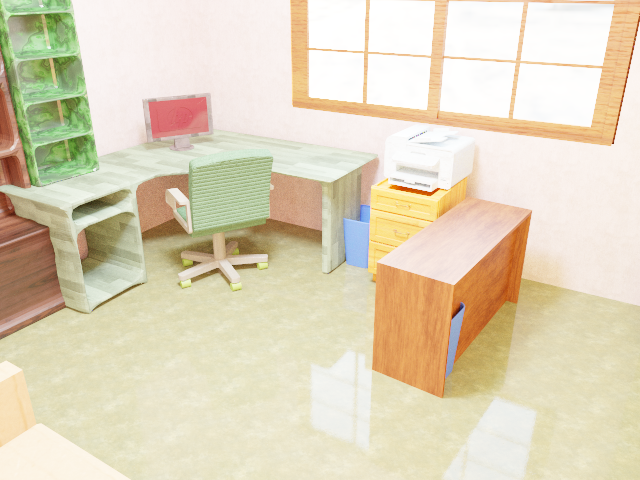} &
        \includegraphics[width=0.20\linewidth]{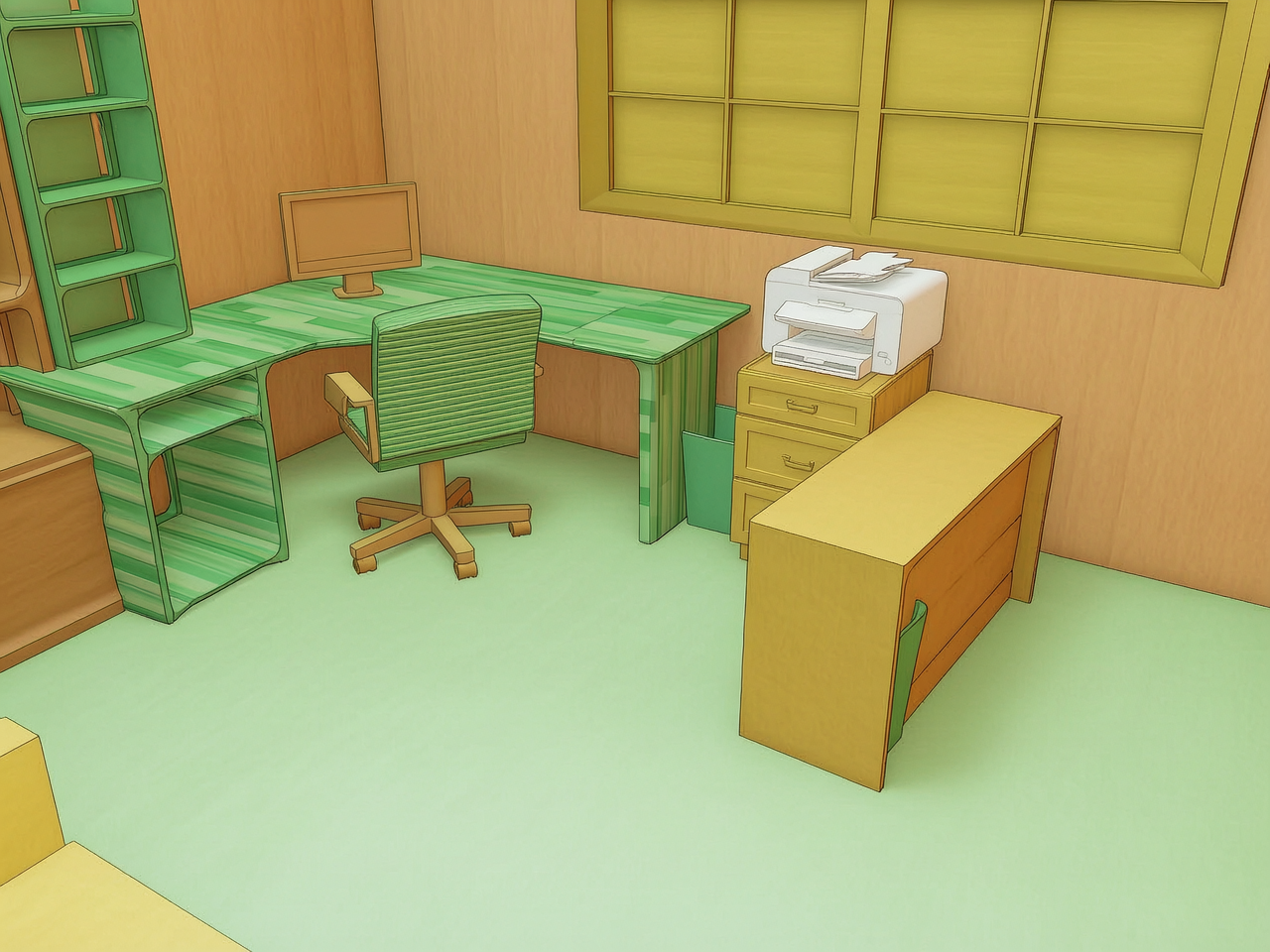} &
        \includegraphics[width=0.20\linewidth]{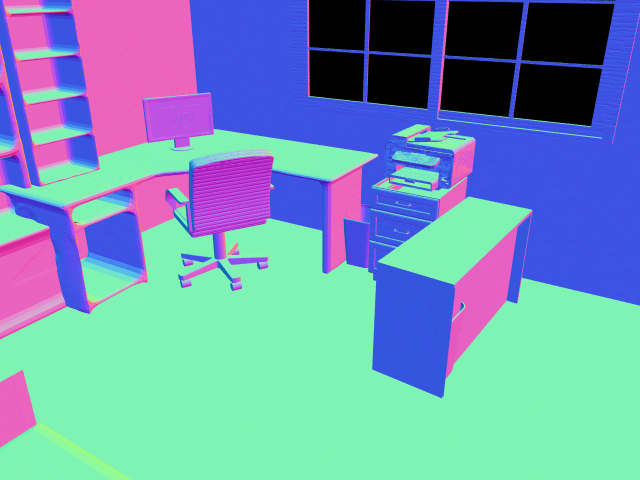} \\

        \includegraphics[width=0.20\linewidth]{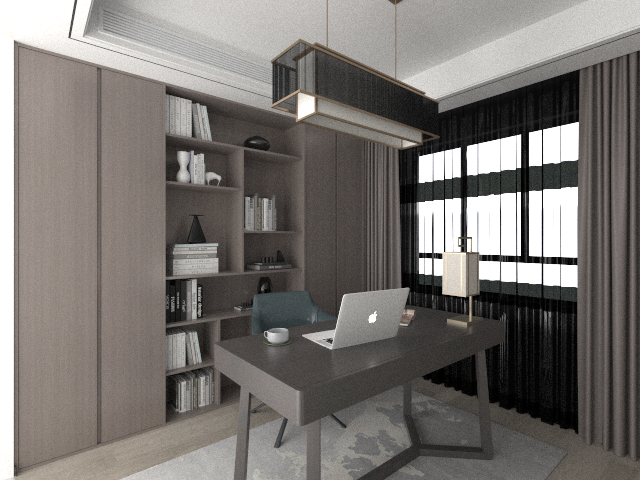} &
        \includegraphics[width=0.20\linewidth]{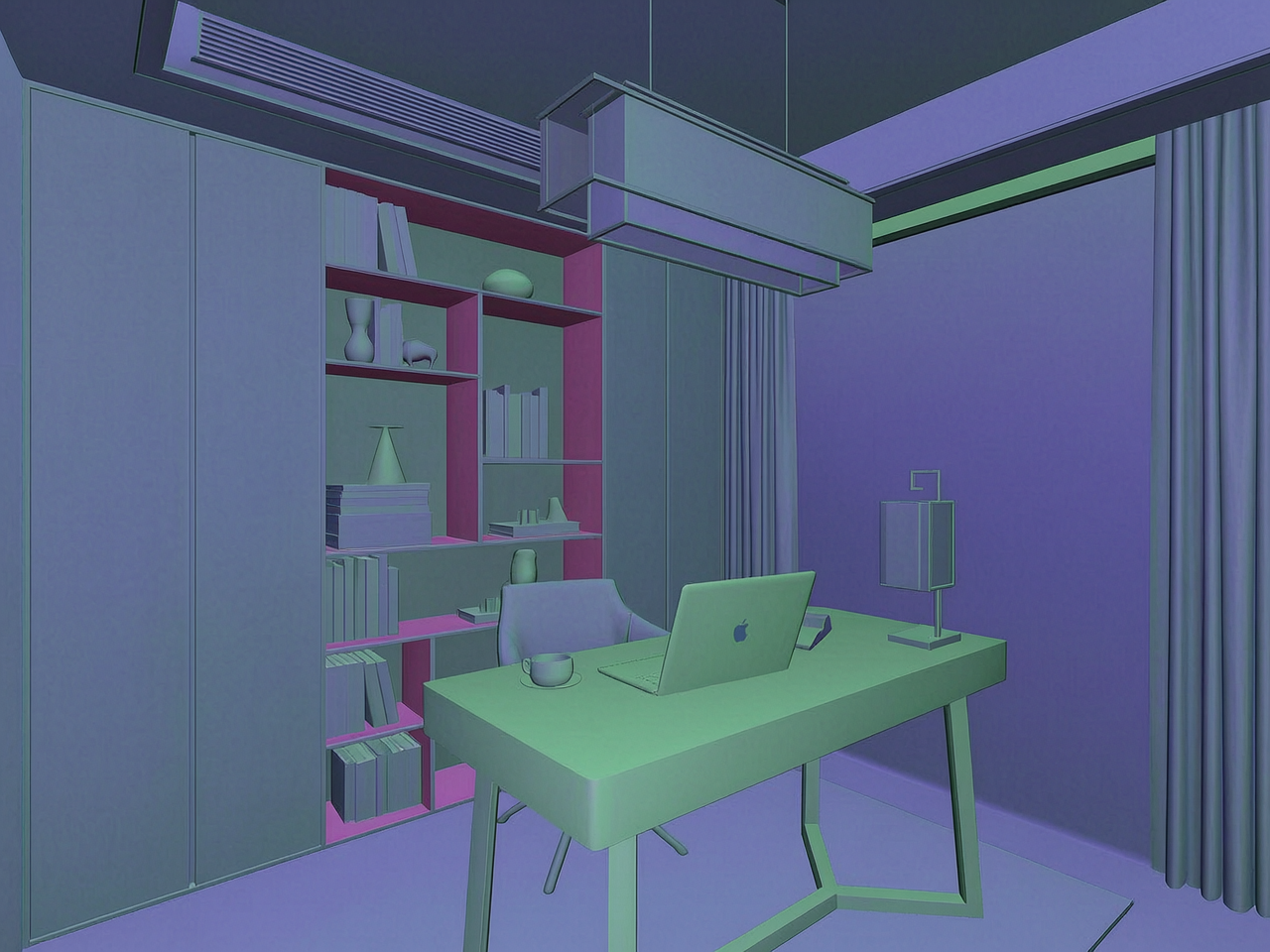} &
        \includegraphics[width=0.20\linewidth]{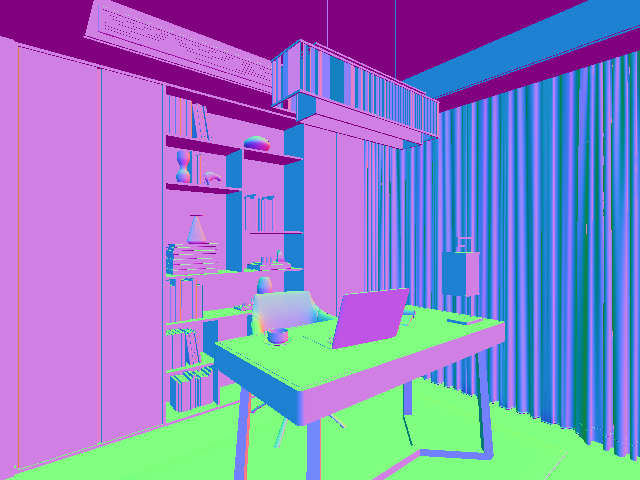} \\

        \includegraphics[width=0.20\linewidth]{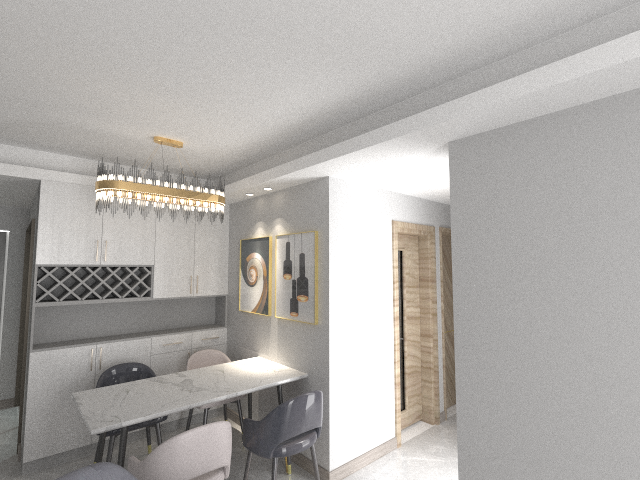} &
        \includegraphics[width=0.20\linewidth]{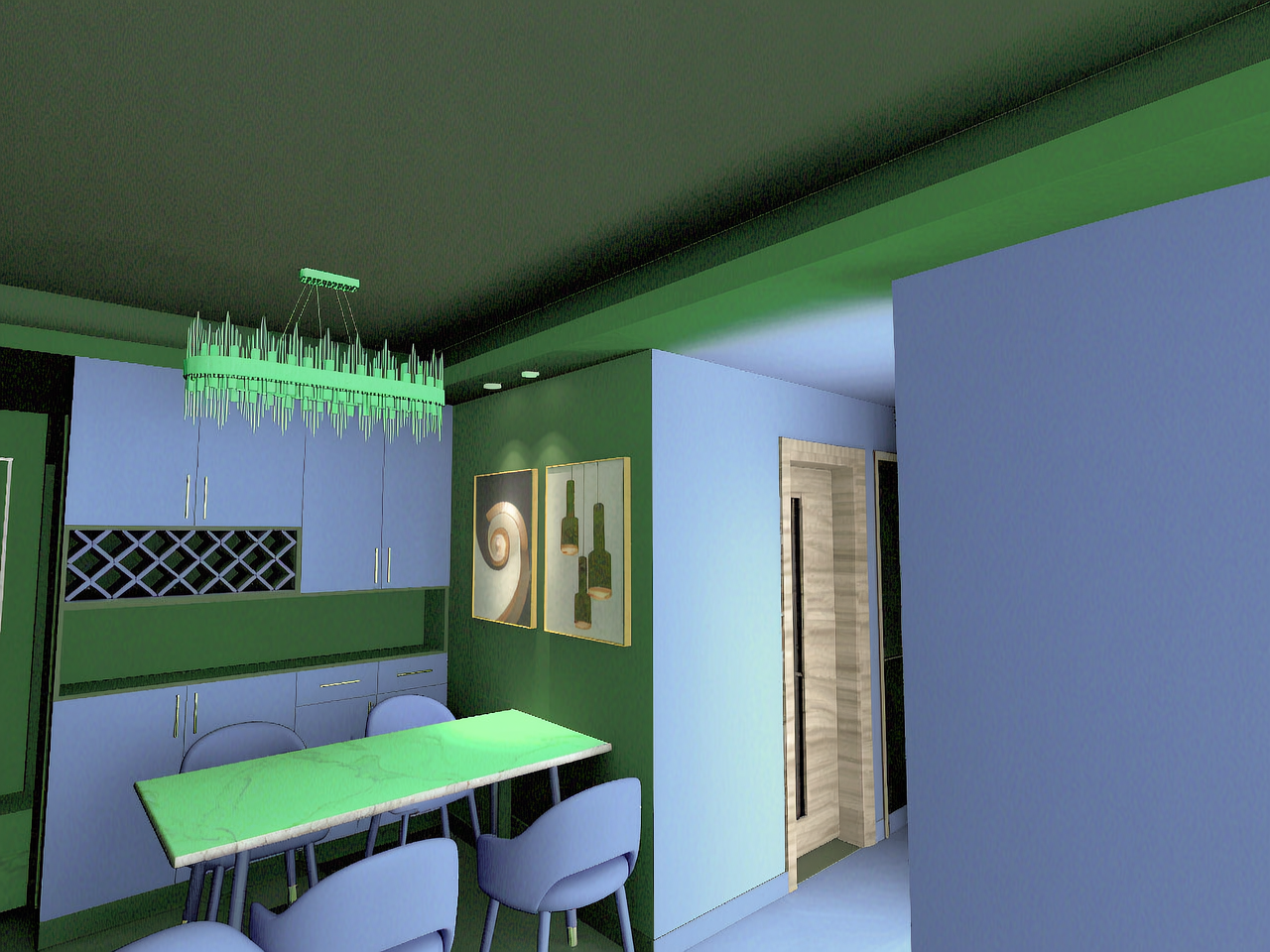} &
        \includegraphics[width=0.20\linewidth]{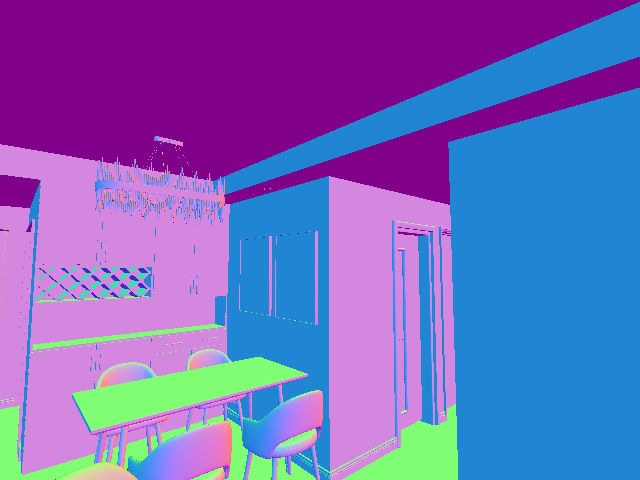} \\
    \end{tabular}
    \caption{
    Representative WAN2.7-Image normal-representation failures under the evaluated normal access setting.
    Each row shows the input RGB image, the WAN2.7-Image output, and the corresponding ground-truth normal map.
    The outputs are documented as representation failures rather than assigned Mean Angular or Acc@22.5 scores because they do not instantiate the spatially aligned dense normal-map representation required by the angular evaluator.
    }
    \label{fig:wan_normal_failure_cases}
\end{figure}

Before applying angular metrics, an output must instantiate the requested dense normal-map representation.
A valid output should be a spatially aligned RGB map over the image domain and should follow the specified normal-map color convention at least as an attempted directional encoding.
The WAN2.7-Image outputs in Figure~\ref{fig:wan_normal_failure_cases} instead show stylized scene renderings, normal-map-like colorizations, or irregular red--green--blue channel patterns that do not correspond to coherent dense surface-orientation fields.
They lack stable directional encoding, surface-wise spatial coherence, and reliable alignment with the scene geometry.

We therefore document these outputs as representation-failure cases rather than as additional angular-error entries.
This distinction is important because angular metrics assume that the prediction is at least an attempted normal map in the evaluator's color convention.
When the output is not a valid dense normal-map representation, a numerical angular error is less informative than documenting the output-validity failure directly.

These examples illustrate a representation-level failure mode of image-editor access for normal prediction: the model can produce colored map-like or stylized outputs while failing to adopt the dense physical-map representation required for angular evaluation.

\subsubsection{Roughness Prior-Utilization Diagnostics}
\label{app:roughness_diagnostics}

\begin{table}[!htbp]
  \centering
  \small
  \setlength{\tabcolsep}{5pt}
  \renewcommand{\arraystretch}{0.98}
  \caption{Roughness prior-utilization diagnostics for \textbf{Doubao Seedream 5.0} on the merged OpenRooms-FF evaluation set.}
  \label{tab:roughness_ablation_overall}
  \begin{tabular}{lccccc}
    \toprule
    Setting & RMSE $\downarrow$ & MAE $\downarrow$ & SSIM $\uparrow$ & LPIPS $\downarrow$ & PSNR $\uparrow$ \\
    \midrule
    A0: RGB + prompt & 0.348 & 0.293 & 0.367 & 0.517 & 9.562 \\
    A1: RGB + prompt + segmentation soft & 0.352 & 0.299 & 0.372 & \textbf{0.513} & 9.483 \\
    A2: RGB + prompt + segmentation region-fill & \textbf{0.324} & \textbf{0.283} & \textbf{0.534} & 0.559 & \textbf{10.324} \\
    A3: RGB + prompt + fixed RGB--roughness exemplar & 0.367 & 0.317 & 0.356 & 0.524 & 9.114 \\
    \bottomrule
  \end{tabular}

  \tabremark{
    A0 is the controlled RGB-to-map setting used in the main roughness comparison.
    A1--A3 add soft segmentation, segmentation-based region filling, or a fixed RGB--roughness exemplar.
    These variants change the access setting and evaluate prior-assisted pipelines rather than the controlled direct RGB-to-map setting.
    RMSE is the primary roughness metric; MAE, SSIM, LPIPS, and PSNR are auxiliary diagnostics.
    LPIPS is auxiliary because roughness maps are single-channel maps replicated to three channels before LPIPS evaluation.
    Boldface marks within-diagnostic numerical optima only and does not indicate official benchmark ranking.
  }
\end{table}

For \textbf{roughness}, the official comparison remains the controlled direct RGB-to-map track because it evaluates whether a system can infer a spatially aligned roughness map from the query RGB image alone, without an additional spatial prior module. We additionally study prior-assisted variants for \textbf{Doubao Seedream 5.0}. These variants are informative but change the evaluated system: even when the prior is derived from the query image, the prediction becomes a segmentation- or exemplar-assisted pipeline rather than a direct RGB-to-map elicitation setting. We therefore report them as diagnostics rather than as part of the controlled-track ranking.

Table~\ref{tab:roughness_ablation_overall} shows that the utility of auxiliary priors depends on how they are operationalized. 
Soft segmentation changes visual organization but does not improve scalar error, whereas region filling improves MAE, PSNR, and SSIM, indicating that explicit region-level structure is useful for roughness. 
The fixed exemplar does not improve the primary or structural metrics, suggesting limited transferability from a single RGB--roughness reference pair.

\subsubsection{Metallic Prior-Utilization Diagnostics}
\label{app:metallic_prior_diagnostics}

For \textbf{metallic}, the official comparison remains the controlled RGB-to-metallic setting because it evaluates whether a system can infer a spatially aligned metallic map from the query RGB image alone.
We additionally study prior-assisted variants for \textbf{Doubao Seedream 5.0}.
These variants add segmentation-derived spatial cues or a fixed RGB--metallic exemplar, and therefore change the access setting.
They are reported as diagnostics rather than as part of the official metallic ranking.

\begin{table}[!htbp]
  \centering
  \small
  \setlength{\tabcolsep}{5pt}
  \renewcommand{\arraystretch}{0.98}
  \caption{Metallic prior-utilization diagnostics for \textbf{Doubao Seedream 5.0} on the 120-image companion diagnostic subset.}
  \label{tab:metallic_ablation_overall}
  \begin{tabular}{lccccc}
    \toprule
    Setting & RMSE $\downarrow$ & MAE $\downarrow$ & SSIM $\uparrow$ & LPIPS $\downarrow$ & PSNR $\uparrow$ \\
    \midrule
    A0: RGB + prompt & 0.205 & 0.083 & 0.496 & 0.358 & 15.880 \\
    A1: RGB + prompt + segmentation soft & 0.205 & \textbf{0.080} & \textbf{0.622} & 0.339 & 15.849 \\
    A2: RGB + prompt + segmentation region-fill & \textbf{0.200} & 0.086 & 0.481 & \textbf{0.261} & \textbf{16.347} \\
    A3: RGB + prompt + fixed RGB--metallic exemplar & 0.218 & 0.096 & 0.429 & 0.368 & 14.941 \\
    \bottomrule
  \end{tabular}

  \tabremark{
    A0 is the controlled RGB-to-metallic setting used in the main metallic comparison.
    A1--A3 add soft segmentation, segmentation-based region filling, or a fixed RGB--metallic exemplar.
    These variants change the access setting and evaluate prior-assisted pipelines rather than the controlled direct RGB-to-map setting.
    Metrics are computed on the 120-image companion diagnostic subset.
    MAE is the primary metallic metric; RMSE, SSIM, LPIPS, and PSNR are auxiliary diagnostics.
    SSIM and LPIPS are auxiliary because metallic maps are sparse continuous material-property maps rather than natural RGB images.
    Boldface marks within-diagnostic numerical optima only and does not indicate official benchmark ranking.
  }
\end{table}

Table~\ref{tab:metallic_ablation_overall} shows that metallic prior utilization is metric-dependent.
The soft segmentation variant improves the primary MAE and SSIM relative to the controlled RGB-to-metallic setting.
The segmentation region-fill variant gives the best RMSE, LPIPS, and PSNR.
The fixed RGB--metallic exemplar does not improve over the controlled setting.
These results suggest that spatial priors can affect metallic-map structure, but they also change the access assumptions and therefore remain diagnostic rather than official ranking results.

\subsection{Depth Protocol Diagnostics}
\label{app:depth_protocol_diagnostics}

For \textbf{depth}, we additionally report protocol diagnostics for exemplar- and segmentation-assisted variants.
These variants change the access setting and are therefore not used for the official controlled RGB-to-depth ranking.

The fixed RGB--depth exemplar improves scalar and ordinal depth agreement on this diagnostic subset, while Boundary F1 decreases, suggesting that global relative-depth alignment and local discontinuity recovery respond differently to auxiliary cues.
The segmentation-reference variant does not improve over the controlled RGB-to-depth prompt setting on these diagnostics.

\subsubsection{Dataset-Wise Diagnostic Breakdowns}
\label{app:dataset_wise_diagnostic_breakdowns}

Tables~\ref{tab:albedo_prompt_dataset_breakdown} and~\ref{tab:normal_prompt_dataset_breakdown}
provide dataset-wise diagnostic breakdowns for the albedo prompt variants and the normal prompt/exemplar-conditioning variants.
For each source, the corresponding \textit{mainaxis} and \textit{stresstest} diagnostic subsets are merged before scoring.
These breakdowns are included only to expose protocol sensitivity and are not used to redefine the official access settings.

\begin{table}[!htbp]
\centering
\footnotesize
\setlength{\tabcolsep}{5pt}
\renewcommand{\arraystretch}{0.95}
\caption{Dataset-wise albedo prompt-ablation diagnostics.}
\label{tab:albedo_prompt_dataset_breakdown}
\begin{tabular}{llcccc}
\toprule
Dataset & Setting & MAE$\downarrow$ & PSNR$\uparrow$ & SSIM$\uparrow$ & LPIPS$\downarrow$ \\
\midrule
\multirow{4}{*}{InteriorVerse}
& A0 & 0.1803 & 13.3495 & \textbf{0.6686} & \textbf{0.3364} \\
& A1 Full prompt & 0.1854 & 13.2065 & 0.6540 & 0.3583 \\
& A2 Weakened prompt & 0.1884 & 13.1744 & 0.6650 & 0.3561 \\
& A3 Minimal prompt & \textbf{0.1641} & \textbf{14.3143} & 0.6479 & 0.3586 \\
\midrule
\multirow{4}{*}{OpenRooms-FF}
& A0 & 0.1535 & 14.8019 & 0.6129 & \textbf{0.3121} \\
& A1 Full prompt & 0.1516 & 14.9051 & 0.6120 & 0.3147 \\
& A2 Weakened prompt & \textbf{0.1442} & \textbf{15.3783} & \textbf{0.6245} & 0.3138 \\
& A3 Minimal prompt & 0.1485 & 15.1095 & 0.5740 & 0.3614 \\
\bottomrule
\end{tabular}

\tabremark{
For each dataset, \textit{mainaxis} and \textit{stresstest} diagnostic subsets are merged into a single score.
A0 denotes the official illumination-aware structure-preserving protocol; A1--A3 are reduced prompt variants.
The table is diagnostic-only and is not used for post-hoc prompt selection.
}
\end{table}

\begin{table}[!htbp]
\centering
\scriptsize
\setlength{\tabcolsep}{4pt}
\renewcommand{\arraystretch}{1.03}
\caption{Dataset-wise normal prompt/exemplar-conditioning diagnostics.}
\label{tab:normal_prompt_dataset_breakdown}
\begin{tabular}{lccccc}
\toprule
Setting & Mean Angular$\downarrow$ & RMSE$\downarrow$
& Acc@11.25$\uparrow$ & Acc@22.5$\uparrow$ & Acc@30$\uparrow$ \\
\midrule
\multicolumn{6}{l}{\textbf{InteriorVerse}} \\
\midrule
A0 & 62.31 & 69.69 & 0.0262 & 0.0980 & 0.1724 \\
A1 & 77.11 & 84.48 & 0.0242 & 0.0747 & 0.1140 \\
A2 & 63.30 & 69.93 & 0.0164 & 0.0694 & 0.1337 \\
A3 & 70.51 & 77.65 & 0.0179 & 0.0739 & 0.1362 \\
\midrule
\multicolumn{6}{l}{\textbf{OpenRooms-FF}} \\
\midrule
A0 & 54.53 & 61.74 & 0.0351 & 0.1327 & 0.2355 \\
A1 & 65.73 & 73.76 & 0.0297 & 0.1080 & 0.1658 \\
A2 & 60.64 & 67.79 & 0.0310 & 0.1089 & 0.1823 \\
A3 & 76.56 & 83.65 & 0.0171 & 0.0660 & 0.1074 \\
\bottomrule
\end{tabular}

\tabremark{
For each dataset, \textit{mainaxis} and \textit{stresstest} diagnostic subsets are merged into a single score.
A0 denotes full prompt + three-exemplar aggregate; A1 removes exemplar conditioning while keeping the full prompt.
A2 denotes minimal prompt + three-exemplar aggregate; A3 uses a minimal prompt without exemplars.
A0 and A2 are computed as per-dataset averages over the three fixed exemplar-conditioned runs.
The table is diagnostic-only and is not used as a main benchmark ranking.
}
\end{table}

Together, the dataset-wise breakdowns show that protocol sensitivity is not confined to a single source.
They are reported to expose access-setting dependence under controlled diagnostic subsets, rather than to select prompts, exemplars, or auxiliary priors after observing benchmark performance.

\section{Qualitative Examples and Failure Cases}
\label{app:qualitative_examples}

This appendix provides qualitative evidence for source-reliability screening. 
The examples are intended to support the benchmark-design decisions and to make typical source-side failure patterns visible beyond aggregate metrics.

\subsection{InteriorVerse Metallic-Channel Reliability Examples}
\label{app:iv_metallic_reliability_examples}

\begin{figure}[!htbp]
\centering
\setlength{\tabcolsep}{2pt}
\renewcommand{\arraystretch}{0.95}

\begin{tabular}{cccc}
\includegraphics[width=0.19\linewidth]{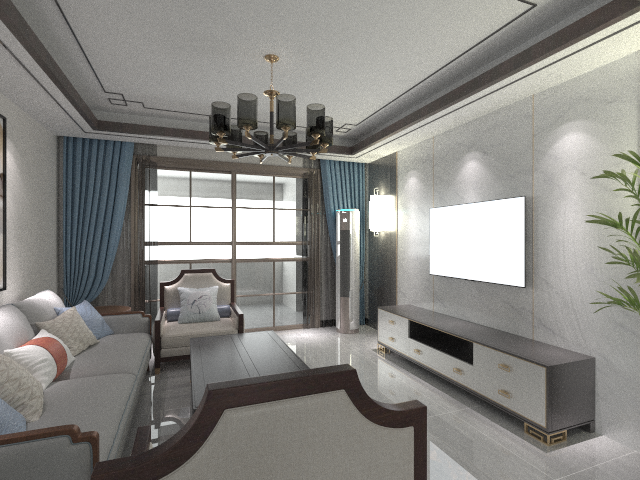} &
\includegraphics[width=0.19\linewidth]{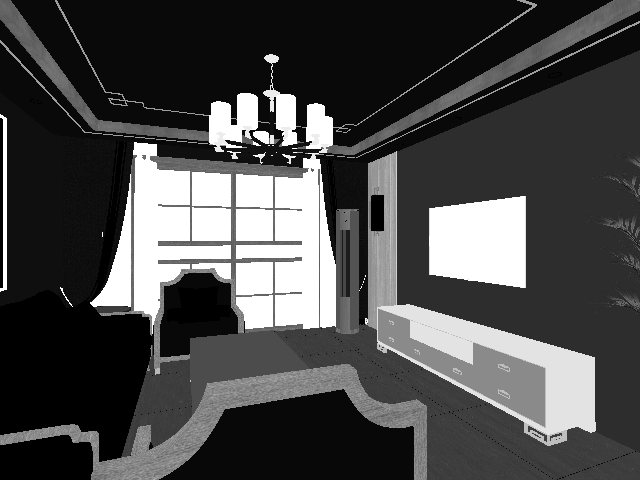} &
\includegraphics[width=0.19\linewidth]{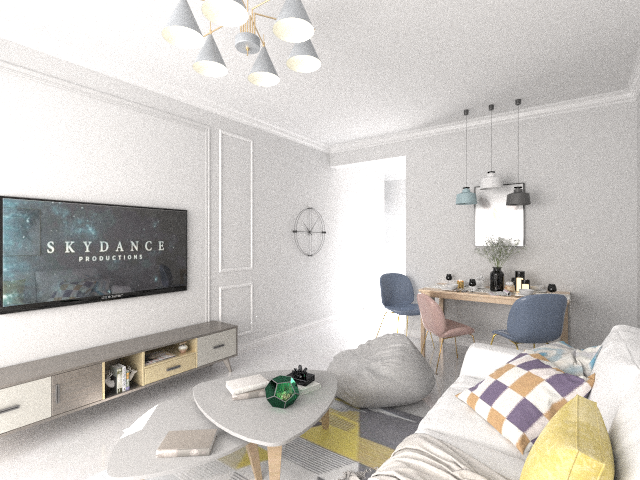} &
\includegraphics[width=0.19\linewidth]{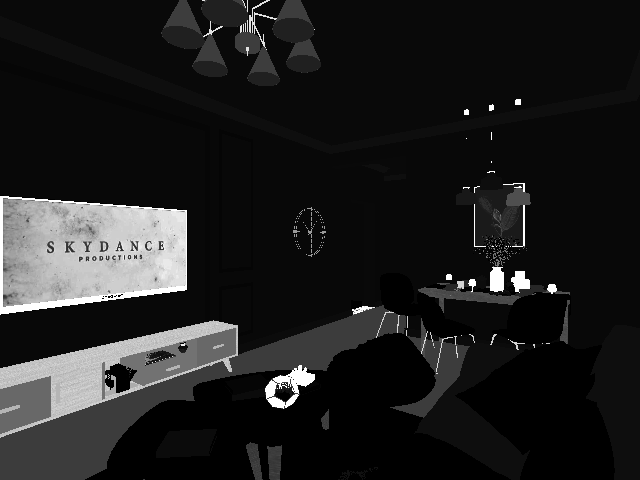} \\[2pt]

\includegraphics[width=0.19\linewidth]{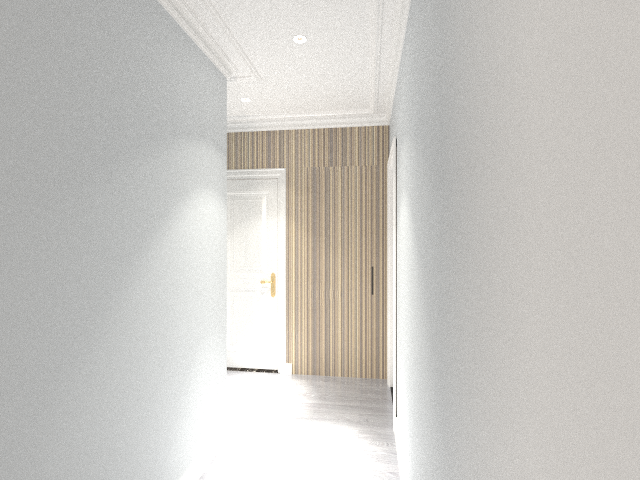} &
\includegraphics[width=0.19\linewidth]{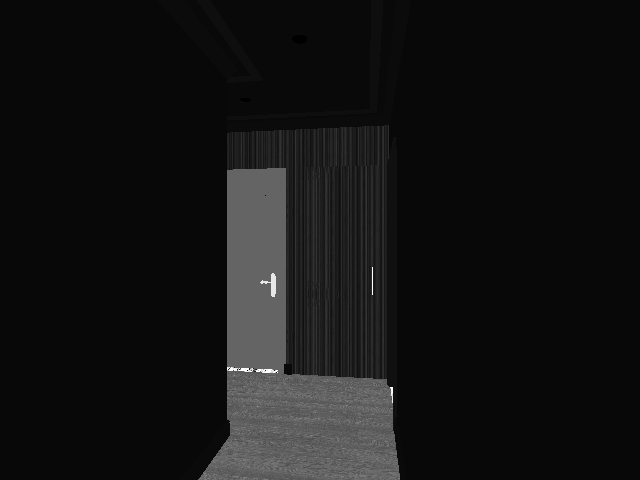} &
\includegraphics[width=0.19\linewidth]{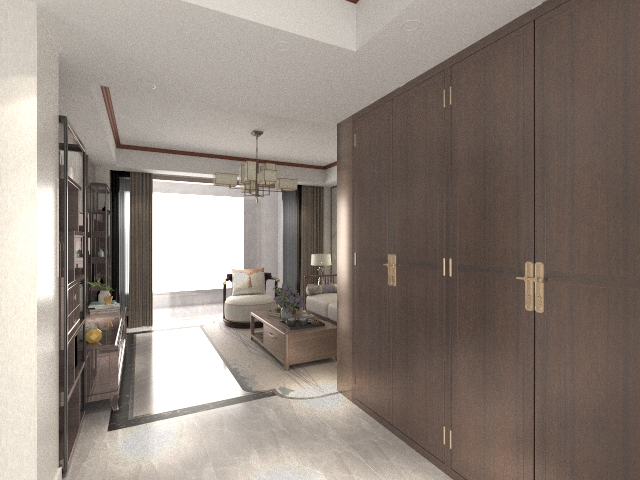} &
\includegraphics[width=0.19\linewidth]{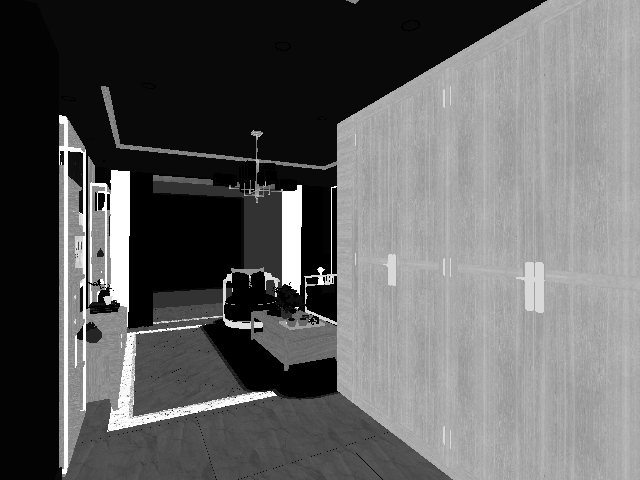} \\[2pt]

\includegraphics[width=0.19\linewidth]{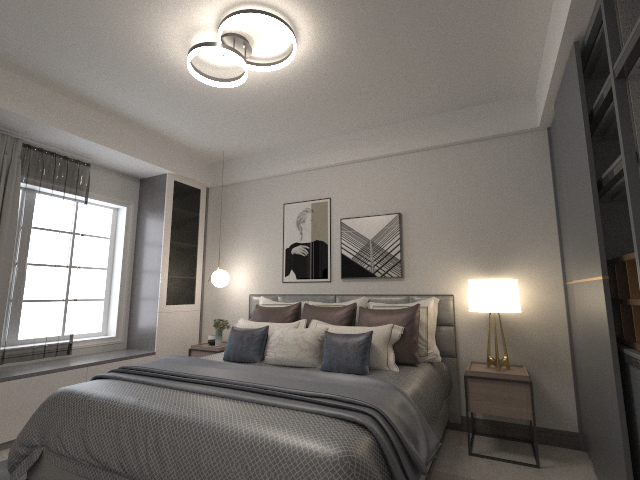} &
\includegraphics[width=0.19\linewidth]{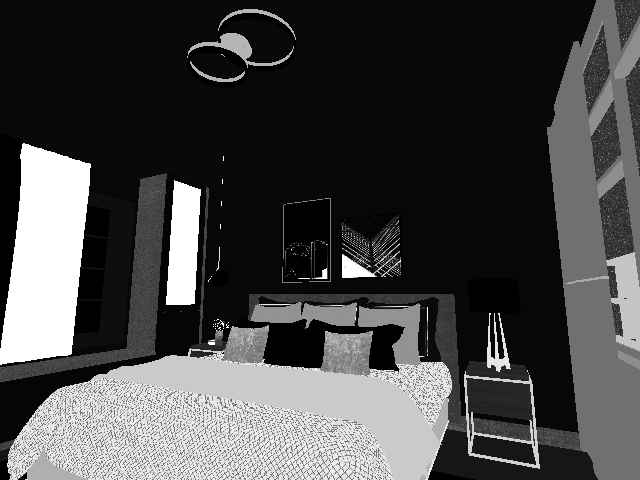} &
\includegraphics[width=0.19\linewidth]{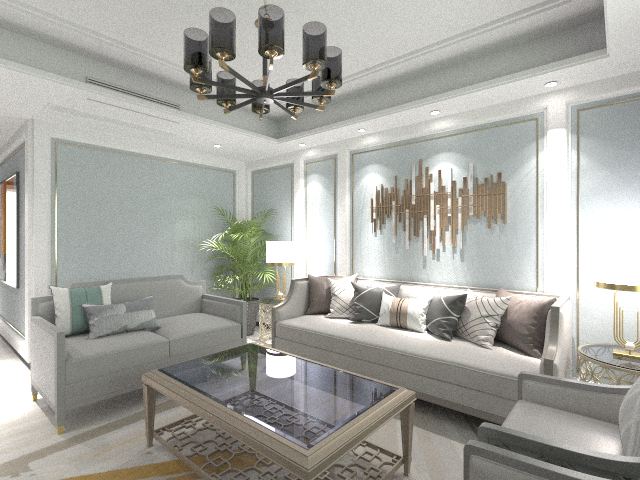} &
\includegraphics[width=0.19\linewidth]{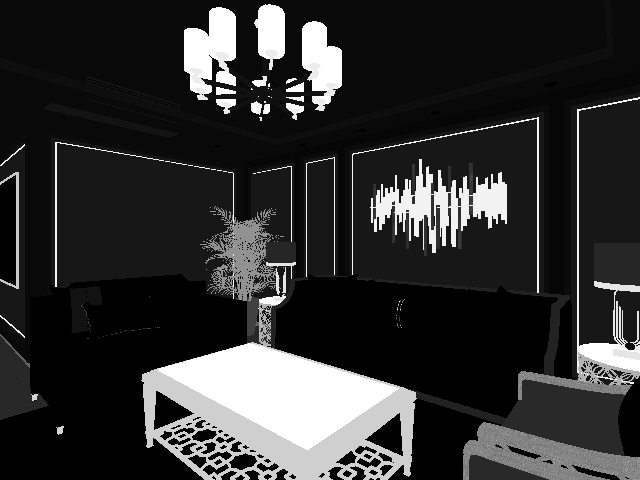}
\end{tabular}

\caption{
Representative InteriorVerse metallic-channel reliability examples.
Within each pair, the left image is the RGB rendering and the right image is the corresponding released metallic channel.
The examples show recurring entanglement with visible texture, screen or artwork content, emissive or over-exposed regions, specular highlights, decorative structures, and object appearance.
These artifacts motivate our conservative exclusion of InteriorVerse metallic from official benchmark scoring.
}
\label{fig:iv_metallic_reliability_examples}
\end{figure}

Figure~\ref{fig:iv_metallic_reliability_examples} shows representative InteriorVerse cases used to support the source-eligibility decision for metallic evaluation.
These examples are not intended as a complete dataset-wide audit.
Instead, they provide qualitative evidence of recurring reliability concerns in the released metallic channel.
Across the examples, the released metallic channel preserves visible RGB texture, screen or artwork content, decorative structures, specular highlights, emissive or over-exposed regions, and object appearance.
Such behavior is problematic for benchmark-grade metallic scoring, because the evaluator may reward agreement with annotation artifacts rather than physically meaningful metallic recovery.

\subsection{InteriorVerse Roughness-Channel Reliability Examples}
\label{app:iv_roughness_reliability_examples}

\begin{figure}[!htbp]
\centering
\setlength{\tabcolsep}{2pt}
\renewcommand{\arraystretch}{0.95}

\begin{tabular}{cccc}
\includegraphics[width=0.19\linewidth]{Metallic_counterexample_im1.png} &
\includegraphics[width=0.19\linewidth]{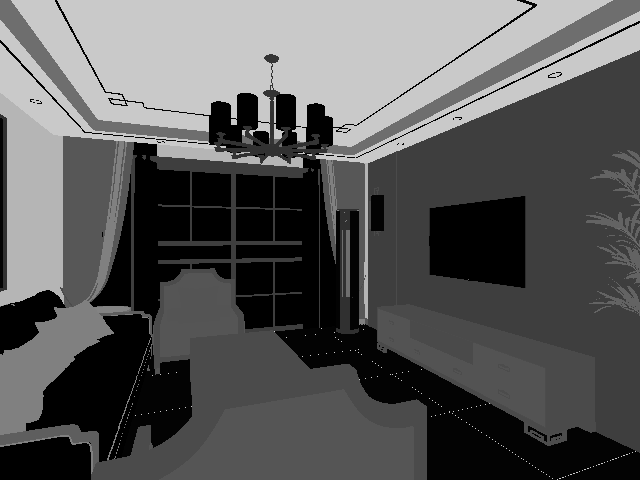} &
\includegraphics[width=0.19\linewidth]{Metallic_counterexample_im2.png} &
\includegraphics[width=0.19\linewidth]{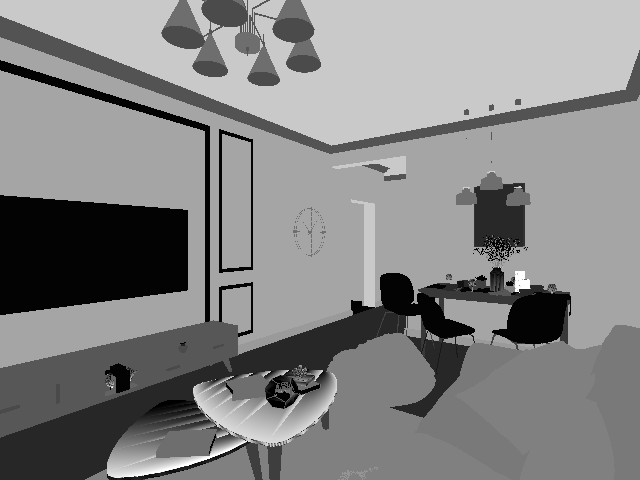} \\[2pt]

\includegraphics[width=0.19\linewidth]{Metallic_counterexample_im4.png} &
\includegraphics[width=0.19\linewidth]{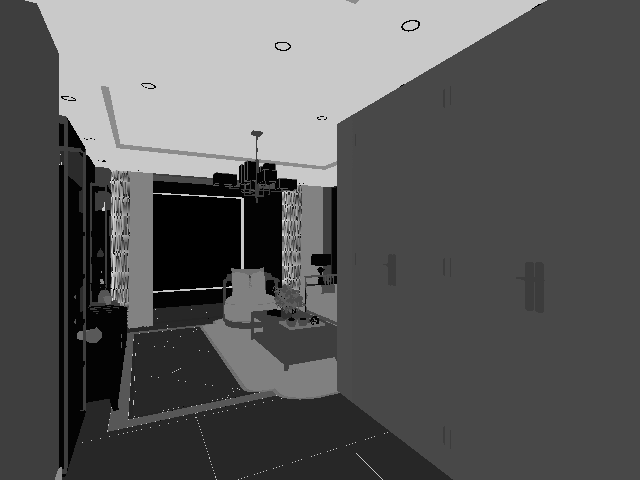} &
\includegraphics[width=0.19\linewidth]{Metallic_counterexample_im5.png} &
\includegraphics[width=0.19\linewidth]{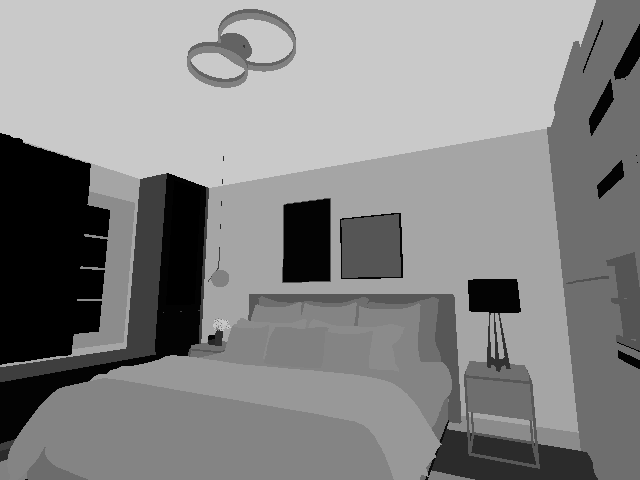} \\[2pt]

\includegraphics[width=0.19\linewidth]{Metallic_counterexample_im6.png} &
\includegraphics[width=0.19\linewidth]{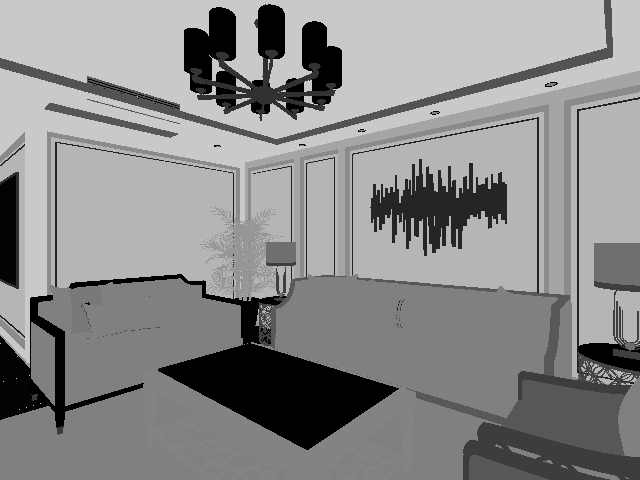} &
\includegraphics[width=0.19\linewidth]{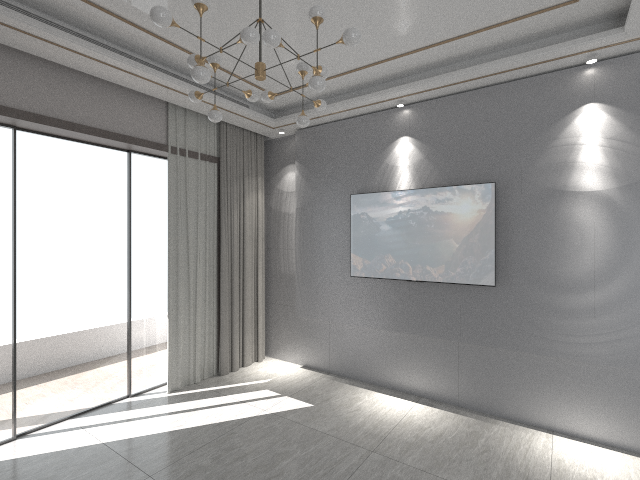} &
\includegraphics[width=0.19\linewidth]{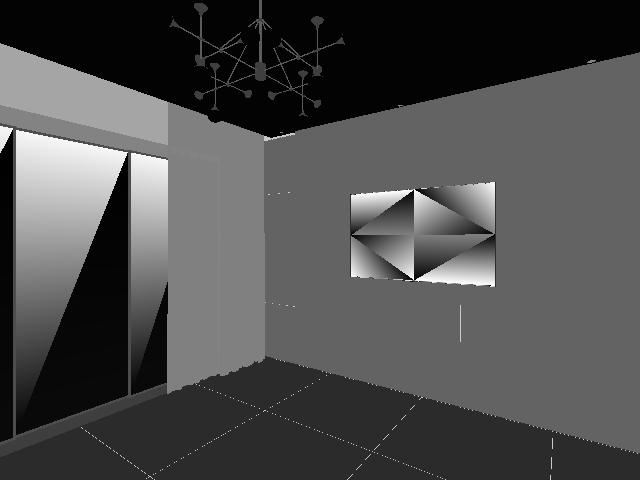}
\end{tabular}

\caption{
Representative InteriorVerse roughness-channel reliability examples.
Within each pair, the left image is the RGB rendering and the right image is the corresponding released roughness channel.
The examples show coarse region-wise responses, weak differentiation across visibly distinct regions such as textile, wood, tile, wall, furniture, screen, glass, and lighting-fixture surfaces, and incomplete alignment with visible material boundaries.
The final example further shows shading-like geometric patterns on screen and glass regions, suggesting that the released roughness channel may be contaminated by view-dependent appearance or rendering artifacts.
These qualitative observations support our conservative exclusion of InteriorVerse roughness from official benchmark scoring.
}
\label{fig:iv_roughness_reliability_examples}
\end{figure}

Figure~\ref{fig:iv_roughness_reliability_examples} shows representative InteriorVerse cases used to support the source-eligibility decision for roughness evaluation.
These examples are not intended as a complete dataset-wide audit.
Instead, they provide qualitative evidence of recurring reliability concerns in the released roughness channel.
Across the examples, the released roughness maps often appear overly piecewise-constant, weakly differentiated across visibly distinct materials, and only coarsely aligned with object or material boundaries.
In several cases, the channel also contains appearance- or shading-like patterns on screen, glass, or strongly illuminated regions, suggesting that the released map may encode view-dependent rendering artifacts rather than only stable material roughness.

These failure patterns are problematic for benchmark-grade roughness scoring.
Roughness is a bounded material parameter, and its evaluation should reflect material-region consistency rather than agreement with source-side artifacts.
If such channels are used as ground truth, a method may be rewarded for matching coarse segmentation-like regions, illumination-correlated responses, or rendering artifacts instead of physically meaningful roughness.
We therefore conservatively exclude InteriorVerse roughness from official scoring and use OpenRooms-FF as the official roughness source.

\FloatBarrier

\section{Reproducibility Details}
\label{app:reproducibility}

The review-time release is organized as a benchmark code and release scaffold rather than a complete bundle of all large benchmark artifacts.
It provides lightweight example manifests, split descriptors, evaluation and aggregation utilities, prompt templates, access-setting documentation, experiment organization, and scene-audit utilities.
The official reported results are tied to frozen evaluation manifests; the manifest schema records, for each evaluated target, the source dataset, retained image identifier, target-map reference, valid-mask reference when applicable, stress-slice metadata, and aggregation group.

For proprietary-system evaluations, full generated outputs may not be redistributable because of provider policies.
We therefore organize run records around prompts or prompt-template identifiers, access-setting identifiers, model/version identifiers, output references when available, post-processing records, evaluator logs, and failure records.
Main-setting runs and diagnostic variants are kept separate, so that outputs generated under different access assumptions are not mixed in the same official comparison.
For specialized predictors, we record the released model or checkpoint identifier, preprocessing configuration, prediction reference, and evaluator version.

The evaluator applies the same resizing, masking, output-validity checking, post-processing, and metric computation rules to all methods evaluated for the same target.
Invalid, unreadable, unmatched, or non-scoreable outputs are handled according to the target-specific failure rules described in Appendix~\ref{app:extended_eval}.
For targets with multiple retained sources, source-balanced summaries are computed by first evaluating each source separately and then macro-averaging source-level scores.

The companion metallic source is released separately from the code repository.
Although it provides multiple aligned inverse-rendering channels for inspection and quality control, its official role in this benchmark is restricted to metallic evaluation.
The public benchmark release will extend the current scaffold with the complete frozen manifests, valid-region definitions, stress-subset metadata, and aggregated scoring artifacts needed to audit the reported results.

\end{document}